\documentclass[10pt]{article}
\usepackage[hidelinks]{hyperref}
\usepackage[preprint]{icml2026}
\usepackage{booktabs}
\usepackage[utf8]{inputenc} % allow utf-8 input
\usepackage[T1]{fontenc}    % use 8-bit T1 fonts
\usepackage{url}            % simple URL typesetting
\usepackage{booktabs}       % professional-quality tables
\usepackage{amsfonts}       % blackboard math symbols
\usepackage{nicefrac}       % compact symbols for 1/2, etc.
\usepackage{microtype}      % microtypography
\usepackage{soul}
\usepackage{graphicx}
\usepackage{amsmath}
\usepackage{outlines}
\usepackage{xurl}

\usepackage{amsmath,amsfonts,bm}

\def\eqref#1{equation~\ref{#1}}
\def\1{\bm{1}}

\DeclareMathAlphabet{\mathsfit}{\encodingdefault}{\sfdefault}{m}{sl}
\SetMathAlphabet{\mathsfit}{bold}{\encodingdefault}{\sfdefault}{bx}{n}

\usepackage{multirow}
\usepackage{paralist}

\usepackage{multicol}
\usepackage{diagbox}

\usepackage[ruled,noend,algo2e]{algorithm2e}

\newtheorem{theorem}{Theorem}
\SetCommentSty{mycommfont}

\usepackage{here}

\usepackage{amsmath,amssymb,amsfonts,amsbsy,amsfonts,latexsym}
\usepackage{makecell}
\usepackage{xcolor}
\usepackage{colortbl}

\usepackage{tabularx,colortbl,xcolor}
\usepackage[normalem]{ulem}
\useunder{\uline}{\ul}{}

\usepackage{enumitem}

\usepackage{xparse}% http://ctan.org/pkg/xparse

\SetKwInput{KwInput}{Input}
\SetKwInput{KwRequire}{Require}

\usepackage{longtable}

\NewDocumentCommand{\var}{O{s} m O{}}{%
  \ensuremath{#1_{#2}^{#3}}% add \vphantom{<bizarre sup>}
}
\usepackage{siunitx}

\newcommand{\commentout}[1]{}

\definecolor{light-gray}{gray}{0.80}

\usepackage{amsthm}

\usepackage{amsthm}

\newtheorem{mytheorem}{Theorem}[section]
\newtheorem{theorem}[mytheorem]{Theorem}
\newtheorem{corollary}{Corollary}[section]
\newtheorem{assumption}{Assumption}
\newtheorem{lemma}{Lemma}[section]
\theoremstyle{definition}

\newtheorem{proposition}{Proposition}[section]
\newtheorem{remark}{Remark}[section]
\theoremstyle{plain}
\newtheorem{remk}{Remark}[section]

\definecolor{myblue}{rgb}{0,0.2,0.8}
\usepackage{xcolor}
\usepackage{listings}

\definecolor{upforestgreen}{rgb}{0.6, 0.8, 0.2}

\usepackage{chngcntr}
\usepackage{adjustbox}
\usepackage{wrapfig}
\usepackage{arydshln}
\usepackage{tcolorbox}
\usepackage{amsthm,amsmath,amssymb}
\usepackage{amsfonts,dsfont}
\usepackage{mathtools}
\usepackage{float}

\definecolor{mygreen}{HTML}{009901}
\definecolor{myred}{HTML}{A52A2A}

\icmltitlerunning{Metric-Dependent Scaling Laws for Sketched Multiclass Softmax Regression}

\begin{document}

\twocolumn[
  \icmltitle{Metric-Dependent Scaling Laws for Sketched Multiclass Softmax Regression}
  \begin{icmlauthorlist}
    \icmlauthor{Ziyan Chen}{sydney}
    \icmlauthor{Linhe Zheng}{unsw}
    \icmlauthor{Peilin Liu}{pennstate}
    \icmlauthor{Zhongzhu Zhou}{futuremls}
    \icmlauthor{Dingxuan Zhou}{sydney}
  \end{icmlauthorlist}
  \icmlaffiliation{sydney}{The University of Sydney}
  \icmlaffiliation{unsw}{The University of New South Wales}
  \icmlaffiliation{pennstate}{The Pennsylvania State University}
  \icmlaffiliation{futuremls}{FutureMLS}
  \icmlcorrespondingauthor{Ziyan Chen}{ziyan.chen@sydney.edu.au}
  \icmlkeywords{Scaling Laws, Softmax Regression, Gradient Descent}
  \vskip 0.3in
]
\printAffiliationsAndNotice{}

\begin{abstract}
We study scaling laws for full-batch gradient descent with general stable
deterministic learning rates in sketched multiclass softmax regression. In a Gaussian teacher--student model with power-law
feature and teacher spectra, the student observes a random feature sketch and
reuses a finite labeled sample throughout optimization.  Sketching generally
makes the student misspecified, so the analysis must jointly account for
approximation, nonlinear finite-time learning, and the fixed empirical
perturbation induced by data reuse.  On an exponentially likely sketch event,
we show that cross-entropy excess and excess zero--one risk share a common
spectral scale, consisting of a resolution term and an effective learned
dimension divided by sample size; teacher--student decision disagreement
follows the square root of that scale.  The resulting balance gives
rate-optimal sketch and cumulative-horizon choices, and synthetic experiments
support the predicted regimes and metric-dependent exponents.
\end{abstract}

\section{Introduction}
\label{sec:introduction}

Scaling laws describe how predictive performance changes with model size,
data, and optimization time. Empirical studies have reported regular
power-law trends across deep learning, language modeling, and autoregressive
generation~\cite{hestness2017deep,kaplan2020scaling,henighan2020scaling,hoffmann2022training},
providing a basis for forecasting performance and allocating training
resources. Theoretical work has sought to explain such behavior through data
geometry, learning dynamics, and spectral structure~\cite{bahri2021explaining,sharma2022manifold,bordelon2024dynamical}.
In particular, sketched linear regression connects power-law feature and
teacher spectra to the interaction among model size, data, and optimization,
including under data reuse~\cite{lin2024scaling,lin2025reuse}.

Softmax classifiers extend logistic regression to multiple classes by
mapping logits to a probability distribution and training with cross-entropy.
They provide output layers for image classifiers and neural language
models~\cite{krizhevsky2012imagenet,he2016deep,bengio2003neural}; the
softmax function is also central to Transformer attention and next-token
prediction~\cite{vaswani2017attention}. Their predicted probabilities and
decision rules capture different aspects of performance, a distinction also
highlighted by work on neural-network calibration~\cite{guo2017calibration}.

Theory has studied multiclass consistency and calibration
\citep{tewari2007consistency,wang2024unified}, as well as the implicit bias of
gradient descent on separable classification data
\citep{soudry2018implicit,ravi2024implicit} and the statistical benefits of
early stopping in logistic regression~\citep{wu2025earlystopping}. Recent
scaling analyses consider frequency heterogeneity in softmax and language
models~\citep{kunstner2024heavytailed,kunstner2025bigram,tsiolis2026sgd},
and power laws arising from softmax concentration or online boundary
dynamics~\citep{liu2026onethird,liu2026emergent,kuhn2026boundary}. These
works study mechanisms complementary to our fixed-class, spectrum- and
sketch-driven setting; see Section~\ref{sec:related-work}. We ask how
probability estimation, excess misclassification risk, and disagreement
with the Bayes classifier scale jointly with model size, data, and training
time.

We study a Gaussian teacher--student model in which a multiclass softmax
student learns from randomly sketched features using full-batch gradient
descent with data reuse. Under power-law spectral assumptions, our results
characterize how approximation, optimization, and finite-sample effects
jointly determine performance. Cross-entropy excess and, under an additional
decision-nondegeneracy condition, excess classification risk share a common
scaling rate, while teacher--student disagreement follows its square root.
The model and formal results are given in Sections~\ref{sec:preliminaries}
and~\ref{sec:main-results}, respectively. Our main contributions are as
follows.

\paragraph{Contributions.}
\begin{itemize}
    \item \textbf{All-horizon metric scaling laws.}
    We establish matching rates for cross-entropy excess, excess zero--one
    risk, and decision disagreement in sketched multiclass softmax regression
    trained with general stable deterministic learning rates.

    \item \textbf{Spectral learning under data reuse.}
    We analyze sketch approximation, nonlinear finite-time optimization, and
    filtered empirical-score fluctuations from reusing a fixed training
    sample. The resulting rates capture both optimization-limited and
    sketch-limited regimes.

    \item \textbf{Decision geometry and rate-optimal tuning.}
    We explain the different metric exponents through decision-boundary
    geometry, derive rate-optimal choices of sketch dimension and cumulative
    horizon within the theorem's regime, and illustrate the predicted trends
    with synthetic experiments.
\end{itemize}

\paragraph{Notation.}
For nonnegative quantities $f$ and $g$, we write $f\lesssim g$ or
$g\gtrsim f$ if $f\le Cg$ for a constant independent of $N$, $M$, and the
optimization horizon, but possibly dependent on fixed problem parameters;
$f\asymp g$ means that both $f\lesssim g$ and $g\lesssim f$ hold. For a vector
$v$, $\|v\|_2$ is its Euclidean norm; for a matrix $A$, $\|A\|_F$ and
$\|A\|_{\mathrm{op}}$ denote the Frobenius and operator norms. For a
positive-semidefinite matrix $W$, set
$\|A\|_W^2:=\operatorname{tr}(AWA^\top)$. We write
$A\preceq B$ when $B-A$ is positive semidefinite. Expectations over the
training sample in the main results are conditional on the sketch.

\section{Preliminaries}
\label{sec:preliminaries}

We consider a multiclass teacher--student classification problem with a Gaussian
random sketch. Let $C \ge 2$ denote the number of classes, which is considered as a fixed value, and we define the effective class dimension $r:=C-1$ to remove the translation invariance of the softmax function. Furthermore, we define $U \in \mathbb{R}^{C\times r}$ be any orthonormal basis mapping to $C$ class probabilities, satisfying
\[
    U^\top U = I_r,
    \qquad
    U^\top \mathbf{1}_C = 0 .
\]

\subsection{Teacher model and sketched observations}
\label{subsec:teacher}

Let the input feature in $d$-dimensional space be sampled from \(x \sim \mathcal{N}(0,\Sigma),\) where $\Sigma$ is a positive semidefinite covariance operator on the input space. The ground-truth teacher is parameterized by  \(A_* \in \mathbb{R}^{r\times d},\) and generates the conditional class probabilities
\[
    p_*(x)
    :=
    \operatorname{softmax}(U A_* x)
    \in \Delta^{C-1}.
\]
Then conditional on $x$, the corresponding one-hot label $y\in\{e_1,\ldots,e_C\}$ is drawn given its conditional class probabilities
\[
    y \mid x
    \sim
    \operatorname{Categorical}\!\left(p_*(x)\right),
\]
where $e_c$ denotes the $c$-th standard basis vector of $\mathbb{R}^C$.

The learner does not observe the full input $x \in \mathbb{R}^d$ to simulate the imperfect data observation. Instead, it observes a Gaussian-sketched input from a lower dimension $M$ as
\[
    z := Sx \in \mathbb{R}^{M},
\]
where the Gaussian sketch matrix $S$ satisfies
\[
    S \in \mathbb{R}^{M\times d},
    \qquad
    S_{ij} \overset{\mathrm{i.i.d.}}{\sim}
    \mathcal{N}\!\left(0,\frac{1}{M}\right).
\]
Here the ambient dimension $d$ is finite and we assume $d\ge C_{\mathrm{amb}}M$ for a fixed $C_{\mathrm{amb}}\ge2$.  It can be verified that conditional on the sketch matrix $S$, the observed input is also normal distributed with sketched covariance, namely
\[
    z \sim \mathcal{N}(0,\Sigma_S),
    \qquad
    \Sigma_S := S\Sigma S^\top .
\]
The training data consist of $N$ i.i.d.\ pairs
\[
    \mathcal{D}_N
    :=
    \{(z_i,y_i)\}_{i=1}^N,
\]
generated by the following teacher sampling model:
\[
\begin{gathered}
    x_i \overset{\mathrm{i.i.d.}}{\sim}\mathcal{N}(0,\Sigma),
    \qquad z_i=Sx_i,\\
    y_i\mid x_i\sim\operatorname{Categorical}\!\left(p_*(x_i)\right).
\end{gathered}
\]
Note again that the learner only has access to the pair $(z_i,y_i)$ with sketched input.

\subsection{Sketched softmax model}
\label{subsec:model}

The student is parameterized by \(V \in \mathbb{R}^{r\times M}\) from a lower dimension. Given a sketched input $z$, its predicted class probabilities are
\[
    p_V(z)
    :=
    \operatorname{softmax}(UVz),
\]
where \(UVz \in \mathbb{R}^{C}\) represents the logits for each class. Equivalently, the predicted probability of class \(c\) is given by
\[
    [p_V(z)]_c
    =
    \frac{
        \exp\!\left((UVz)_c\right)
    }{
        \sum_{\ell=1}^{C}
        \exp\!\left((UVz)_\ell\right)
    },
    \qquad
    c=1,\ldots,C .
\]

For an observation $(z,y)$, we use the standard multiclass cross-entropy
loss as the training target,
\begin{align*}
    \ell(V;z,y)
    :&=
    -y^\top \log p_V(z) \\
    &= \log\!\left(
        \sum_{c=1}^{C}
        \exp\!\left((UVz)_c\right)
    \right)
    -
    y^\top UVz .
\end{align*}
Its gradient with respect to $V$ is
\[
    \nabla_V \ell(V;z,y)
    =
    U^\top
    \bigl(p_V(z)-y\bigr)z^\top .
\]

\subsection{Optimization procedures}
\label{subsec:optimization}

We study population and empirical full-batch gradient descent initialized at
zero and using a general deterministic positive learning rates $\eta_1,\ldots,\eta_T$. Define the cumulative learning horizon and the maximum step size by
\[
    H_t:=\sum_{s=1}^t\eta_s,
    \qquad H_0:=0,
    \qquad \eta_{\max}:=\max_{1\le t\le T}\eta_t.
\]
The sequence of learning rate is prefixed and independent of the sketch matrix and training samples. We work with stable steps
\[
    0<\eta_t\le\eta_{\max}\le\eta_0\qquad(1\le t\le T),
\]
where $\eta_0>0$ is sufficiently small and depends only on the fixed problem parameters or constants $(a,b,r,c_\lambda,C_\lambda,c_*,C_*,R)$, introduced in Section~\ref{sec:assumptions}. 

It is considered to be small enough so that all population and empirical curvature factors on the high-probability events (introduced at the end of Section~\ref{sec:assumptions}) obey $\eta_t\mu\le c_0<1$, and the corresponding GD steps are descending. The empirical Hessian event in Lemma~\ref{lem:uniform-empirical-hessian} further specifies the reason why the smallness of $\eta_0$, which ensures descent along the localized trajectory under the sample-size condition in Theorem~\ref{thm:metric-scaling}.

We first consider the population gradient descent starts from $V_0^{\mathrm{pop}}=0$, which updates according to the population risk
\[
\begin{gathered}
    V_t^{\mathrm{pop}}
    =V_{t-1}^{\mathrm{pop}}-\eta_t\nabla\mathcal R_S(V_{t-1}^{\mathrm{pop}}),
    \quad t=1,\ldots,T.
\end{gathered}
\]
Then for the empirical full-batch gradient descent, define
\[
    \widehat R_N(V):=\frac1N\sum_{i=1}^N\ell(V;z_i,y_i)
\]
and similarly, starting from $V_0=0$,
\[
\begin{gathered}
    V_t=V_{t-1}-\eta_t\nabla\widehat R_N(V_{t-1}),
    \quad t=1,\ldots,T.
\end{gathered}
\]
Note that full-batch means the same $N$ independent training observations are reused at every iteration. Here $N$ is the sample size, $T$ is the number of full-batch
iterations, and we use the last iterate $V_T$ as the empirical estimator.

\subsection{Population risk and evaluation}
\label{subsec:evaluation}

For any student parameter $V$, define its population cross-entropy risk
\[
    R_{\mathrm{CE}}(V)
    :=
    \mathbb E_{x,y}\bigl[\ell(V;Sx,y)\bigr],
    \qquad
    \mathcal R_S(V):=R_{\mathrm{CE}}(V),
\]
where $x\sim\mathcal N(0,\Sigma)$ and $y\mid x\sim\operatorname{Categorical}(p_*(x))$. The teacher Bayes
cross-entropy risk is
\[
    R_{\mathrm{CE}}^\star
    :=
    \mathbb E_{x,y}\bigl[-y^\top\log p_*(x)\bigr],
    \qquad
    \mathcal R_*:=R_{\mathrm{CE}}^\star.
\]
The cross-entropy excess risk is
\[
    \boxed{
    \mathcal E_{\mathrm{CE}}(V)
    :=R_{\mathrm{CE}}(V)-R_{\mathrm{CE}}^\star.
    }
\]

Define the teacher and student top-1 classifiers by
\[
\begin{aligned}
    g_V(x)&:=\arg\max_{1\le c\le C}(UVSx)_c,\\
    g_\star(x)&:=\arg\max_{1\le c\le C}(UA_\star x)_c.
\end{aligned}
\]
The second metric is excess misclassification risk,
\[
    R_{01}(g):=\Pr(g(x)\ne y),
\]
and the corresponding excess risk is denoted as
\[
    \boxed{
    \begin{aligned}
    \mathcal E_{01}(V)
    &:=R_{01}(g_V)-R_{01}(g_\star)\\
    &=\mathbb E_x\!
    \left[p_{\star,g_\star(x)}(x)-p_{\star,g_V(x)}(x)\right].
    \end{aligned}
    }
\]
The equality follows since $g_\star$ is the conditional Bayes classifier
under $p_*(x)$. 

The third metric is the unweighted teacher--student decision disagreement
\[
    \boxed{
    D(V):=\Pr_x\bigl(g_V(x)\ne g_\star(x)\bigr).
    }
\]
Unlike $\mathcal E_{01}$, $D$ weights every disagreement equally; $\mathcal E_{01}$ weights it by the teacher's conditional class-probability gap.

For the cross-entropy analysis, let
\[
    V_M^*\in\arg\min_{V\in\mathbb R^{r\times M}}R_{\mathrm{CE}}(V)
\]
be the population-optimal sketched predictor. A supplementary risk-level
approximation--bias--variance identity is given in
Appendix~\ref{app:preliminary}.

\subsection{Oracle-Aware Error Decomposition}

To understand how sketching and training affect the learned classifier, we
first compare the student and teacher logits. This comparison separates the
information lost through sketching from the errors within the observed
feature space, including finite-time optimization and sample fluctuations.
It also makes explicit the shift between the conditional-mean predictor and
the population softmax optimum caused by misspecification. The resulting
decomposition is an exact identity for logit error; its connection to the
evaluation risks is established through softmax and decision geometry.

\begin{lemma}[Oracle-aware error decomposition]
\label{lem:main-error-decomposition}
Fix a sketch $S$ and suppose that a population minimizer $V_M^*$ exists.
Let $V_M^{\mathrm{LS}}$ be a linear conditional-mean predictor satisfying
$V_M^{\mathrm{LS}}Sx=\mathbb E[A_*x\mid Sx]$, and define
\[
\begin{gathered}
    H_M:=V_M^*-V_M^{\mathrm{LS}},\qquad
    A_*x=V_M^{\mathrm{LS}}Sx+\xi_M,
\end{gathered}
\]
and recall the exact empirical-GD decomposition
$V_T-V_M^*=B_T^{\mathrm{fr}}+Z_T+D_T$ from
Appendix~\ref{app:fullbatch-empirical}, using its finite-sum definitions
when the frozen Hessian is singular. For an independent test input, let
$\mathbb E_x$ denote expectation conditional on $S$ and the training data.
Then
\[
\boxed{
\begin{aligned}
&\underbrace{V_TSx-A_*x}_{\text{total logit error}} 
={}-\underbrace{\xi_M}_{\text{sketch residual}}
+\underbrace{H_MSx}_{\text{softmax oracle shift}} \\
& +\underbrace{B_T^{\mathrm{fr}}Sx}_{\text{frozen optimization bias}}
+\underbrace{Z_TSx}_{\text{filtered sample fluctuation}}
 +\underbrace{D_TSx}_{\text{nonlinear remainder}} .
\end{aligned}
}
\]
Moreover, with $W=\Sigma_S$ and
$\Delta_M=\mathbb E_x\|\xi_M\|_2^2$,
\[
\boxed{
\begin{aligned}
\mathbb E_x\|V_TSx-A_*x\|_2^2
={}&\underbrace{\Delta_M}_{\text{sketch approximation}}\\
&+\underbrace{\bigl\|H_M+B_T^{\mathrm{fr}}+Z_T+D_T\bigr\|_W^2}
_{\text{student-observable error}} .
\end{aligned}
}
\]
The four terms inside the second norm are generally coupled; in particular,
no independence between $Z_T$ and $D_T$ is assumed.
\end{lemma}

\begin{proof}
Gaussian conditional regression makes $\xi_M$ orthogonal to every function
linear in $Sx$.  Substituting the empirical-GD parameter decomposition and
then taking the test-input second moment proves both identities.  The fully
expanded cross terms and the corresponding CE interpretation are given in
Appendix~\ref{app:preliminary}.
\end{proof}

Lemma~\ref{lem:main-error-decomposition} separates information lost before
training from errors accumulated inside the observed feature space. It does
not require the power-law assumptions or the sketch event introduced below.

\section{Main Results}
\label{sec:main-results}

\subsection{Assumptions and Interpretations}
\label{sec:assumptions}

In this section, we state the assumptions used in our analysis and explain their
roles and reasonability. 

Let the eigendecomposition of the input covariance operator
\[
    \Sigma v_j = \lambda_j v_j,
    \qquad
    \lambda_1 \ge \lambda_2 \ge \cdots >0.
\]

\begin{assumption}[Power-law covariance]
\label{ass:covariance}
There exist constants $0<c_\lambda\le C_\lambda<\infty$ and an exponent
$a>1$ such that
\[
    c_\lambda j^{-a}
    \le
    \lambda_j
    \le
    C_\lambda j^{-a},
    \qquad
    1\le j\le d.
\]
\end{assumption}

The power law models inputs whose leading directions carry more variation,
while retaining a spectrum of progressively weaker directions. Such spectral
assumptions are standard in sketched linear-regression scaling analyses
\citep{lin2024scaling,lin2025reuse}. The restriction $a>1$ makes
$\operatorname{tr}(\Sigma)$ uniformly bounded as $d$ grows. Working in the
eigenbasis of $\Sigma$ is a change of coordinates: the Gaussian sketch is
invariant in distribution under a common orthogonal rotation of the input,
teacher, and sketch.

\begin{assumption}[Power-law teacher source]
\label{ass:source}
There exist constants $0<c_*\le C_*<\infty$ and an exponent $b$ satisfying
\[
    1<b<a+1
\]
such that
\[
    c_* j^{-b}
    \le
    \lambda_j\|A_*v_j\|_2^2
    \le
    C_* j^{-b},
    \qquad
    1\le j\le d.
\]
\end{assumption}

The quantity $\lambda_j\|A_*v_j\|_2^2$ is the teacher's logit energy in
input direction $v_j$. Its power-law decay is a stylized source-regularity
condition, analogous to spectral target assumptions used in linear-regression
scaling laws~\citep{lin2024scaling,lin2025reuse}; it constrains this energy,
rather than the teacher's eigenvectors. The lower bound ensures a sufficiently rich
tail for matching lower rates. The condition $b>1$ bounds the total teacher
prediction energy, while $b<a+1$ places the source regime used in our main theorem. We write $q:=\frac{b-1}{a}\in(0,1)$ for the resulting source exponent.

\begin{remark}[Bounded teacher prediction norm]
\label{rem:teacher-bound}
Under Assumption~\ref{ass:source}, there exists a constant $R<\infty$ such that
\[
    \|A_*\Sigma^{1/2}\|_{\mathrm{op}}\le R.
\]
This follows because
$\|A_*\Sigma^{1/2}\|_{\mathrm{op}}^2\le
\sum_{j=1}^d\lambda_j\|A_*v_j\|_2^2\le C_*\sum_{j\ge1}j^{-b}<\infty$.
The bound prevents the variance of teacher logits from growing without
bound and is used in the uniform softmax-curvature and localization
arguments.
\end{remark}

\begin{assumption}[Nondegenerate teacher contrasts]
\label{ass:decision-nondegeneracy}
There is a fixed $\kappa_{\mathrm{dec}}>0$ such that
\[
    A_*\Sigma A_*^\top\succeq\kappa_{\mathrm{dec}}I_r.
\]
\end{assumption}

This condition keeps the teacher's contrast logits nondegenerate in every
class-space direction; together with the bound in Remark~\ref{rem:teacher-bound}, it
gives uniform lower and upper bounds on pairwise margin variances. It is a
sufficient condition for the decision-boundary geometry behind
our classification and disagreement results, but is not needed for the
cross-entropy theorems. Margin and low-noise conditions also play a central
role in classification-rate analyses~\citep{audibert2007fast}, although
they are not identical to this covariance condition.

\paragraph{Spectral interpretation.}

Here we explicitly interpret the effect brings by the spectral power law. Recall we study the joint scaling law in the large-sample and large-model regime
\[
    N\to\infty,
    \qquad M\to\infty,
\]
with fixed class dimension $r$ and other structural constants. Under
power law Assumption~\ref{ass:covariance}, modes satisfying $\lambda_j H_T\asymp1$
have index $j\asymp H_T^{1/a}$. Thus the effective number of learned input
directions is
\[
    d_{\mathrm{eff}}(H_T,M)\asymp\min\{M,H_T^{1/a}\}=:k_T,
\]
and the total effective dimension is $r\,d_{\mathrm{eff}}(H_T,M)\asymp r k_T$.

We distinguish the sketch-limited regime $M\lesssim H_T^{1/a}$ from the
optimization-limited regime $M\gtrsim H_T^{1/a}$. In the latter, the bias caused by insufficient optimization is the bottleneck and it is with scale $H_T^{-(b-1)/a}$; in the former, by contrast, the sketch approximation scale $M^{1-b}$ limits the attainable resolution.

\paragraph{High-probability sketch event.}
To derive the main results, we consider the joint event
\[
    \mathcal G_M^{\mathrm{all}}
    :=
    \mathcal G_M^{\mathrm{spec}}
    \cap\mathcal G_M^{\mathrm{trace}}
    \cap\mathcal G_M^{\mathrm{ridge}}
    \cap\mathcal G_M^{\mathrm{approx}},
\]
whose components have the following meanings.

\begin{itemize}[leftmargin=*]
    \item \textbf{Spectrum:} $\mathcal G_M^{\mathrm{spec}}$ is the event that the eigenvalues of $\Sigma_S$ satisfy
    $\mu_j(\Sigma_S)\asymp j^{-a}$ for every $1\le j\le M$. Thus the sketch preserves the input's power-law spectrum up to its dimension.

    \item \textbf{Trace:} $\mathcal G_M^{\mathrm{trace}}$ is the event $\operatorname{tr}(\Sigma_S)\le C_\Sigma$, namely the trace of the sketched covariance is bounded. It bounds the total sketched feature energy uniformly in dimension, which is useful for controlling moments and softmax curvature.

    \item \textbf{Ridge source:} Recall the conditional-mean predictor
    $V_M^{\mathrm{LS}}$ from Lemma~\ref{lem:main-error-decomposition}.
    Define the ridge and low-frequency source energies by
    \[
    \begin{aligned}
    K_{\Sigma_S}(u)&:=\operatorname{tr}\!\left[V_M^{\mathrm{LS}}
      u\Sigma_S(\Sigma_S+uI)^{-1}(V_M^{\mathrm{LS}})^\top\right],\\
    F_{\Sigma_S}(u)&:=\|V_M^{\mathrm{LS}}\mathbf1_{(0,u]}(\Sigma_S)\|_{\Sigma_S}^2.
    \end{aligned}
    \]
    The event $\mathcal G_M^{\mathrm{ridge}}$ requires the following, where
    $L>0$ is a sufficiently large fixed constant independent of $M$:
    \begin{itemize}
        \item $K_{\Sigma_S}(u)\asymp u^q$ and $F_{\Sigma_S}(u)\lesssim u^q$ uniformly for $u\in[D_0M^{-a},u_0]$,
    \item $F_{\Sigma_S}(u)\gtrsim u^q$ for $u\in[LD_0M^{-a},u_0]$.
    \end{itemize}

Intuitively, it preserves the teacher's source energy at every relevant spectral resolution, enabling finite-time bias bounds.

    \item \textbf{Approximation:} $\mathcal G_M^{\mathrm{approx}}$
    requires that (i) the sketch is well conditioned on a fixed fraction of
    the leading input directions, (ii) its tail covariance is suitably
    bounded, and (iii) a fixed fraction of the teacher energy in a block of
    order $M$ lies outside the corresponding random sketch row space. The
    first two properties control approximation from above, while the third
    supplies its matching lower bound; in particular, the latter is not
    implied by the spectrum event alone.

\end{itemize}

The precise constants, block definitions, and component probability bounds
are given in Appendix~\ref{app:sketch-events}. Under the stated assumptions,
\[
    \mathbb P_S(\mathcal G_M^{\mathrm{all}})
    \ge 1-Ce^{-cM}.
\]
The intersection therefore holds with exponentially high probability in
the sketch dimension $M$.

Throughout this section, condition on a sketch
$S\in\mathcal G_M^{\mathrm{all}}$. Write
\[
    Q_{M,N,T}=k_T^{1-b}+\frac{k_T}{N}.
\]

\subsection{Population and Full-Batch GD Scaling}

We first remove sampling effects and ask how much of the teacher can be
resolved by population GD at a finite optimization horizon.

\begin{theorem}[All-horizon population cross-entropy scaling]
\label{thm:population-gd-scaling}
Under the setup of Section~\ref{sec:preliminaries}, suppose
Assumptions~\ref{ass:covariance}--\ref{ass:source} hold, with
fixed $C$, $a>1$, and $1<b<a+1$. Let $M,T\to\infty$ and assume
$H_T\to\infty$. Then, on
$\mathcal G_M^{\mathrm{all}}$,
\[
\boxed{
\mathcal E_{\mathrm{CE}}(V_T^{\mathrm{pop}})
\asymp k_T^{1-b}.
}
\]
The sketch event has probability at least $1-Ce^{-cM}$.
\end{theorem}

Theorem~\ref{thm:population-gd-scaling} shows that population optimization
resolves only $k_T$ spectral directions. Equivalently, its risk is of order
$M^{1-b}+H_T^{-q}$: before the sketch cutoff the rate is
set by finite training time, and after it the sketch approximation floor
dominates.  We next restore the reused sample and quantify the additional
filtered empirical-score fluctuation.

\begin{theorem}[Finite-iteration full-batch GD cross-entropy scaling]
\label{thm:metric-scaling}
Under the setup of Section~\ref{sec:preliminaries}, suppose
Assumptions~\ref{ass:covariance}--\ref{ass:source} hold, with
fixed $C$ and $1<b<a+1$. Consider
sequences $N,M,T\to\infty$ with $H_T\to\infty$ such that
\[
    \frac{M^2\log^3(N+T)}{Nk_T}\longrightarrow0.
\]
Then, on the stated sketch event,
\[
\mathbb E\mathcal E_{\mathrm{CE}}(V_T)
\asymp
\underbracket{M^{1-b}}_{\text{Approx.}}
+\underbracket{H_T^{-q}}_{\text{Bias}}
+\underbracket{\frac{k_T}{N}}_{\text{Variance}}.
\]
\end{theorem}

The approximation term measures information lost to sketching, the bias
term reflects finite optimization time, and the variance term is the
filtered finite-sample fluctuation scale. These are scales in the total-risk
law, not an exact additive decomposition: after the sketch cutoff, the bias
term is not asserted to match the parameter-optimization bias, and the
variance term is not identified with the exact centered-iterate variance.

\paragraph{On the sample-size condition.}
The condition $M^2\log^3(N+T)/(Nk_T)\to0$ is sufficient for the uniform
empirical Hessian and localization estimates and the nonlinear-remainder
control along the reused-data GD trajectory. It is a proof condition; we do
not claim that the factor $M^2$ or the logarithmic factors are
information-theoretically necessary.

\paragraph{Proof sketch.}
The oracle-aware decomposition organizes the proof, while the exact
risk-level identity in Appendix~\ref{app:preliminary} separates the CE
approximation term from the parameter excess.  Gaussian-sketch approximation
gives
$\operatorname{Approx}_M\asymp\Delta_M\asymp M^{1-b}$, and oracle
compatibility gives $\|H_M\|_W\lesssim\Delta_M$, so the squared shift is
lower order.  The frozen bias is of order $H_T^{-q}$ before the cutoff and
is at most $M^{1-b}$ after it, while the filtered score satisfies
$\mathbb E\|Z_T\|_G^2\asymp k_T/N$.  Before the cutoff the nonlinear
remainder is lower order under the displayed sample-size condition.  After
the cutoff it is controlled at the total scale and is lower order whenever
the sampling fluctuation supplies the lower bound; when approximation
dominates, $\Delta_M$ itself supplies that bound.  All cross terms are
handled by moment bounds and reverse-triangle estimates, without assuming
that $D_T$ and $Z_T$ are independent.  Local Fisher geometry then transfers
the parameter bounds to CE excess, and
$M^{1-b}+H_T^{-q}\asymp k_T^{1-b}$ gives the stated all-horizon law.
In the optimization-limited regime before the sketch cutoff, the
source-filter lower bound supplies the
finite-time scale $H_T^{-(b-1)/a}\asymp k_T^{1-b}$, while the filtered
score supplies the $k_T/N$ contribution when it is visible. After the cutoff,
the approximation floor $M^{1-b}\asymp k_T^{1-b}$ supplies the first term
in the total-risk lower bound, without a matching parameter-bias lower bound.

\paragraph{From nonlinear dynamics to a spectral filter.}
The key step is to expand empirical GD around the sketched population
optimum and freeze the population Hessian there. The resulting exact
decomposition separates a linear spectral filter and a fixed empirical-score
term from a nonlinear remainder. The Fisher sandwich
(Lemma~\ref{lem:fisher-sandwich}) makes the frozen Hessian comparable to
the sketched covariance, while quasi-Fisher equivalence
(Lemma~\ref{lem:quasi-fisher}) controls the score covariance even under
misspecification. Uniform empirical localization
(Lemma~\ref{lem:uniform-empirical-hessian}) and the quadratic softmax
remainder bound (Lemma~\ref{lem:local-softmax-remainder}) feed into the
nonlinear-response bounds (Lemmas~\ref{lem:fullbatch-nonlinear-response} and
\ref{lem:post-cutoff-response}), which control the error made by freezing the
dynamics in their respective regimes. Thus spectral-filter estimates can be
transferred back to the original nonlinear problem without treating the
linearization as an exact equality.

\paragraph{From population to finite samples.}
Theorem~\ref{thm:population-gd-scaling} identifies
$k_T^{1-b}$ as the population spectral-resolution scale, while
Theorem~\ref{thm:metric-scaling} shows that reusing a sample of size $N$
adds the effective-dimension term $k_T/N$.  Equivalently,
$k_T^{1-b}\asymp M^{1-b}+H_T^{-q}$, with the larger of the sketch
resolution and finite-time resolution determining the population rate.
This equivalence concerns total teacher-relative risk: after the sketch
cutoff, $M^{1-b}$ is the approximation floor rather than a parameter
bias around $V_M^*$.  The latter nevertheless satisfies the uniform
upper bound
\[
\mathbb E[\mathcal R_S(V_T)-\mathcal R_S(V_M^*)\mid S]
\lesssim k_T^{1-b}+\frac{k_T}{N}.
\]
Before the sketch cutoff, the approximation, optimization-bias, and
frozen-score scales are respectively $M^{1-b}$, $H_T^{-q}$, and
$k_T/N$. There $k_T\asymp H_T^{1/a}$, so
\[
    \frac{k_T/N}{H_T^{-q}}\asymp\frac{k_T^b}{N}.
\]
Thus the large-sample condition does not force the frozen-score term to be
negligible: the theorem includes bias-dominated, balanced, and
fluctuation-dominated subregimes. When $N\gg k_T^b$, the frozen-score
fluctuation is lower order; the total-risk equivalence then does not identify
a separate leading statistical-variance contribution. The frozen-score
variance remains a distinct statement about $Z_T$, not an identification
of the exact centered-iterate variance.

After the cutoff, $k_T\asymp M$ and the irreducible approximation term
$M^{1-b}$ replaces the smaller formal continuation $H_T^{-q}$ in the
total-risk lower bound. Thus the unified law is controlled by the learned
dimension $k_T$.

\subsection{Further Implications}

Cross-entropy alone does not determine how often the learned decision rule
changes.  The next theorem transfers the same angular prediction scale to
excess classification risk and to unweighted teacher--student disagreement.

\begin{theorem}[Full-batch GD classification scaling]
\label{thm:classification-scaling}
Under the assumptions and asymptotic regime of
Theorem~\ref{thm:metric-scaling} and Assumption~\ref{ass:decision-nondegeneracy}, conditional on
$S\in\mathcal G_M^{\mathrm{all}}$,
\[
\begin{aligned}
\mathbb E\mathcal E_{01}(V_T)
&\asymp M^{1-b}+H_T^{-q}+\frac{k_T}{N},\\
\mathbb E D(V_T)
&\asymp\left(M^{1-b}+H_T^{-q}+\frac{k_T}{N}\right)^{1/2}.
\end{aligned}
\]
The expectation is over the training sample, and the constants are
uniform on the stated sketch event.
\end{theorem}

Thus cross-entropy and excess zero--one risk are quadratic in the underlying
angular error, whereas decision disagreement is first order and therefore
has the square-root rate.  Having identified all three metric laws, we can
balance spectral resolution against finite-sample fluctuation.

\begin{corollary}[Rate-optimal tuning]
\label{cor:rate-optimal-tuning}
Under the assumptions of Theorem~\ref{thm:classification-scaling}, choose
\[
    M\asymp N^{1/b},\qquad H_T\asymp N^{a/b},
\]
with any fixed positive transition-scale constants, and assume additionally
that the schedule length $T$ satisfies the large-sample condition in
Theorem~\ref{thm:metric-scaling}. This choice gives
$Q_{M,N,T}\asymp N^{-(b-1)/b}$. Consequently,
\[
\boxed{
\begin{aligned}
\mathbb E\mathcal E_{\mathrm{CE}}(V_T)&\asymp N^{-(b-1)/b},\\
\mathbb E\mathcal E_{01}(V_T)&\asymp N^{-(b-1)/b},\\
\mathbb E D(V_T)&\asymp N^{-(b-1)/(2b)}.
\end{aligned}}
\]
This is the optimal rate obtained by balancing the two terms of the
all-horizon scaling law.
\end{corollary}

These are asymptotic exponent-level tuning laws, with hidden constants
depending on the fixed problem and stability parameters, rather than
universal finite-sample hyperparameter prescriptions.

The population theorem is proved in Appendix~\ref{app:scaling-proofs};
the empirical cross-entropy proof is assembled there from
Appendices~\ref{app:fullbatch-bias}--\ref{app:fullbatch-empirical}. The
classification proof additionally uses the low-frequency angular-bias
bound and the fixed-codimension filtered-score estimate in
Appendix~\ref{app:decision-geometry}.

\section{Synthetic Experiments}
\label{sec:experiments}

\begin{figure*}[!t]
    \centering
    \includegraphics[width=0.995\linewidth]{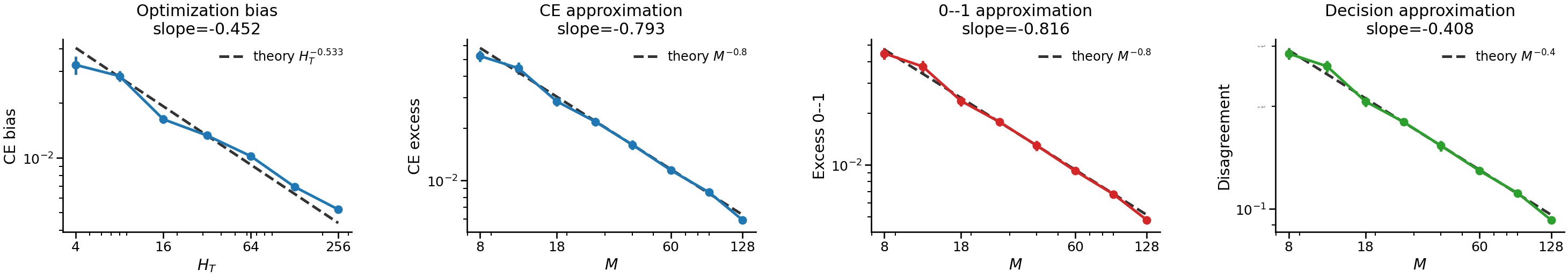}
    \caption{Population component checks.  From left to right:
    optimization-only cross-entropy bias along a path with fixed
    $M/H_T^{1/a}$, followed by the three converged Gaussian-sketch
    approximation metrics as $M$ varies.  Dashed lines are
    one-point-normalized theoretical power-law references.}
    \label{fig:population-checks}
\end{figure*}

\begin{figure*}[!t]
    \centering
    \includegraphics[width=0.995\linewidth]{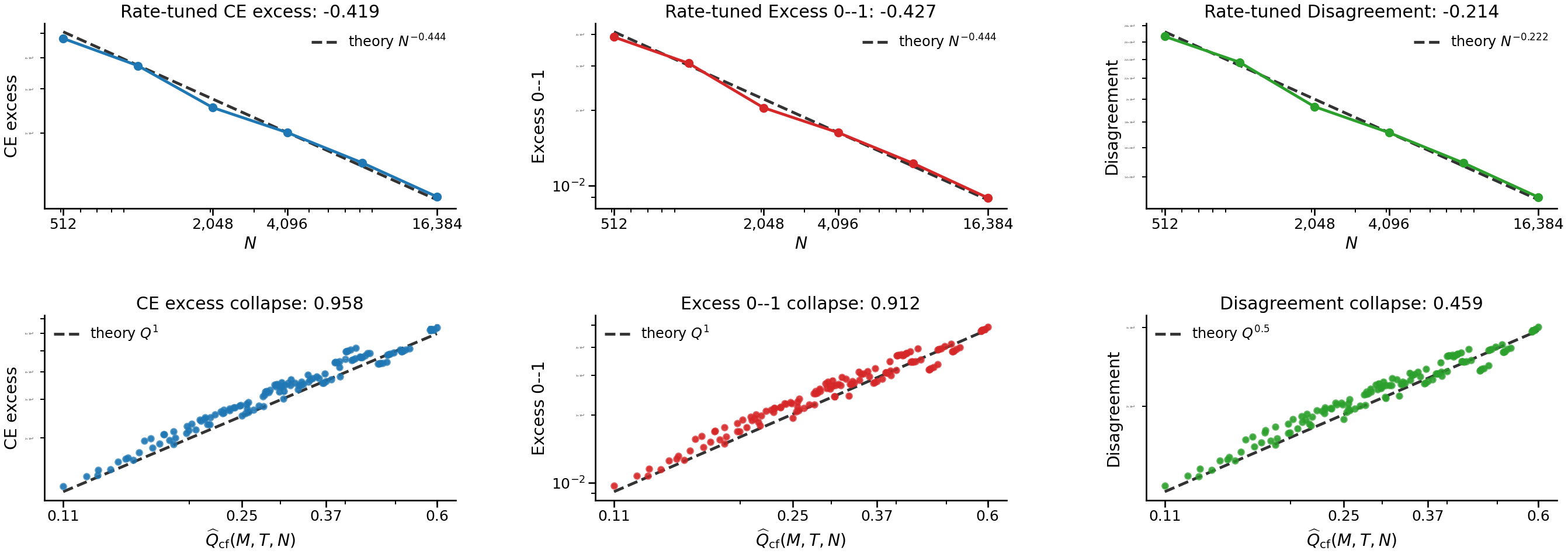}
    \caption{Empirical full-batch GD checks. Top row: rate-optimal joint
    tuning of $M$ and $H_T$ as $N$ varies. Bottom row: the three metrics plotted
    against the leave-one-$N$-out, prefactor-calibrated common scale
    $\widehat Q_{\mathrm{cf}}(M,T,N)$.  The calibration changes only relative
    finite-grid constants; the displayed exponents are fixed by theory.}
    \label{fig:empirical-checks}
\end{figure*}

\paragraph{Setup.}
We simulate the Gaussian teacher--student model in Section~\ref{sec:main-results}
with three classes, $a=1.5$, $b=1.8$, and an ambient truncation at least
eight times the relevant sketch dimension.
Thus $q=(b-1)/a=0.533$, while the rate-optimal prediction and
disagreement exponents are respectively $(b-1)/b=0.444$ and $0.222$.
The teacher source is constructed so that
$\lambda_j\|A_*v_j\|_2^2\propto j^{-b}$, the sketch is Gaussian, and
full-batch GD starts at zero and reuses one training sample. We use the
constant learning rate $\eta_t=0.5$, so $H_T=0.5T$. Population
metrics are evaluated on an independent Gaussian test set.  Error bars in
Figures~\ref{fig:population-checks}--\ref{fig:empirical-checks} are standard
errors over independent sketches and, where applicable, independent training
samples.

\paragraph{Population checks.}
The first experiment varies $H_T$ while keeping the spectral-resolution ratio
$M/H_T^{1/a}$ fixed. The cross-entropy, excess $0$--$1$, and
disagreement slopes are $-0.499$, $-0.477$, and $-0.241$, compared with
$-0.533$, $-0.533$, and $-0.267$; subtracting the corresponding converged
population optimum gives an optimization-only cross-entropy slope of
$-0.452$.
The second experiment numerically converges the population student for each
sketch dimension $M$.  The resulting cross-entropy, excess $0$--$1$ risk,
and disagreement slopes are $-0.793$, $-0.816$, and $-0.408$, respectively,
matching the predicted sketch exponents $-0.8$, $-0.8$, and $-0.4$.

\paragraph{Rate-tuned empirical GD.}
For the third experiment we tune
$M\asymp N^{1/b}$ and $H_T\asymp N^{a/b}$ and train empirical full-batch GD
with reused sampled labels.  This directly tests Corollary~\ref{cor:rate-optimal-tuning}:
the fitted cross-entropy, excess $0$--$1$, and disagreement slopes are
$-0.419$, $-0.427$, and $-0.214$, versus $-0.444$, $-0.444$, and $-0.222$.
The agreement supports both the common prediction-risk rate and the
square-root metric transfer for disagreement.

\paragraph{Common-scale collapse.}
Finally, we vary $(M,T,N)$ jointly over $130$ design points.  The relation
$Q_{M,N,T}\asymp M^{1-b}+H_T^{-q}+k_T/N$ specifies the three contributions
only up to constants, so assigning every finite-grid prefactor the value one
is not an invariant test.  We therefore estimate two positive relative
prefactors jointly across the three metrics, normalizing the optimization
prefactor to one, while keeping the metric exponents $1,1,1/2$ fixed.  For
each sample size $N$, the prefactors are estimated using only the other four
sample sizes and are then evaluated on the held-out $N$ slice.  Against this
cross-fitted common scale, the fitted slopes are $0.958$, $0.912$, and
$0.459$, close to $1$, $1$, and $1/2$. Thus the collapse tests the predicted
common spectral scale without using any point to calibrate its own horizontal
coordinate.

\section{Related Work}
\label{sec:related-work}

\paragraph{Sketched spectral scaling in linear regression.}
Our methodological backbone is the spectral theory of scaling laws in linear
regression developed by \citet{lin2024scaling,lin2025reuse}, which combines
power-law feature and source spectra with random sketching and optimization
filtering.  We adopt this framework but study multiclass softmax
cross-entropy: sketching generally induces misspecification, requiring
control of nonlinear curvature, Fisher geometry, and the fixed empirical
perturbation created by label reuse.

\paragraph{Early stopping in logistic regression.}
\citet{wu2025earlystopping} show that early-stopped gradient descent
can outperform interpolating or asymptotic estimators in overparameterized,
well-specified binary logistic regression, and relate early stopping to
explicit $\ell_2$ regularization.  We also view stopping as a spectral filter,
but analyze a fixed-class, sketched and misspecified multiclass model and
jointly track sketch dimension, sample size, and optimization time.

\paragraph{Calibration, statistical rates, and effective dimension.}
Classical work relates surrogate and classification risks through calibration
\citep{bartlett2006convexity} and establishes minimax rates under margin
conditions~\citep{audibert2007fast}; effective-rank analyses characterize
covariance concentration~\citep{koltchinskii2017concentration}. These
results are complementary to our joint study of sketch dimension, sample
size, and finite optimization horizon under spectral decay and
sketch-induced misspecification.

\paragraph{Frequency-driven softmax scaling.}
Power laws can also arise from heavy-tailed class or token frequencies.
\citet{tsiolis2026sgd} show that SGD learns frequent classes before rare ones
in high-dimensional multiclass Gaussian mixtures, while
\citet{kunstner2024heavytailed,kunstner2025bigram} study how
class imbalance or Zipf frequencies produce optimizer-dependent scaling.
Like us, they study capacity and optimization effects, but their mechanism is
frequency heterogeneity.  We instead
derive scaling from input and teacher spectra, sketching, and sample size.

\paragraph{Intrinsic softmax and boundary dynamics.}
\citet{liu2026onethird,liu2026emergent} derive a one-third time law
from softmax concentration on peaked targets, including in attention, without
requiring a power-law input spectrum.  \citet{kuhn2026boundary}
obtain a related law in online teacher--student classification from examples
near the decision boundary and persistent online noise.  These mechanisms are
complementary: theirs comes from softmax saturation or boundary dynamics,
whereas ours is set by spectral decay and source regularity and is coupled to
sketch and sample-size cutoffs.

\section{Conclusion}
\label{sec:conclusion}

We studied full-batch gradient descent with stable deterministic learning rates for a sketched multiclass softmax
student trained on samples from a Gaussian teacher.  Under power-law
feature and teacher spectra, our analysis gives an all-horizon scaling law
that combines feature compression, finite optimization time, and statistical
error.  It also shows how decision geometry separates the evaluation metrics:
cross-entropy and excess classification risk have the same rate, while
teacher--student disagreement has a different, square-root rate.  The theory
further yields rate-optimal choices of sketch dimension and cumulative learning horizon,
and the experiments reproduce the predicted regimes.

Natural next steps are to treat non-Gaussian inputs, data-dependent sketches,
and stochastic optimization, and to investigate whether the same spectral
mechanism persists in learned representations and growing-class problems.

% \newpage
% \input{app/08_statements}

\newpage
\bibliography{ref}
\bibliographystyle{icml2026}

\clearpage
\appendix
\onecolumn
\counterwithin{mytheorem}{section}
\counterwithin{equation}{section}
\begin{center}
	\textbf{Metric-Dependent Softmax Scaling Law\\ \vspace{8pt} \textit{Supplementary Material}}
\end{center}

\vspace*{0.8em}
\noindent{\LARGE\bfseries Table of Contents}\par
\vspace{0.4em}
\hrule
\vspace{0.6em}

\begingroup
\footnotesize
\setlength{\parindent}{0pt}
\setlength{\parskip}{2pt}
\newcommand{\apptocsection}[2]{%
  \noindent\hspace*{1.2em}%
  \makebox[2.1em][l]{\bfseries\ref{#2}}%
  {\bfseries #1}\nobreak\hfill\makebox[2em][r]{\bfseries\pageref{#2}}\par%
}
\newcommand{\apptocsubsection}[2]{%
  \noindent\hspace*{3.3em}%
  \makebox[3em][l]{\ref{#2}}%
  #1\nobreak\leaders\hbox to 0.55em{\hss.\hss}\hfill\makebox[2em][r]{\pageref{#2}}\par%
}

\apptocsection{Preliminary Geometry and Risk Decomposition}{app:preliminary}
\apptocsubsection{Deterministic learning-rate filters}{app:toc-filters}
\apptocsubsection{High-probability Gaussian-sketch events}{app:sketch-events}
\apptocsubsection{Oracle-aware logit-error decomposition}{app:toc-oracle}
\apptocsubsection{Pairwise contrast frame and angular energy}{app:toc-frame}
\apptocsection{Proofs for the Approximation Error}{app:approximation}
\apptocsubsection{General Upper Bound}{app:approx-upper}
\apptocsubsection{General Lower Bound}{app:approx-lower}
\apptocsubsection{Power-Law Scaling}{app:approx-scaling}
\apptocsubsection{Compatibility of the conditional-mean oracle}{app:toc-oracle-compatibility}
\apptocsection{Full-Batch GD: Population and Frozen Bias}{app:fullbatch-bias}
\apptocsubsection{Genuine nonlinear population gradient descent}{app:toc-population-bias}
\apptocsubsection{Frozen-linear spectral bias}{app:toc-frozen-bias}
\apptocsection{Empirical Full-Batch GD: Score Fluctuation and Nonlinear Response}{app:fullbatch-empirical}
\apptocsubsection{Exact decomposition}{app:toc-empirical-decomposition}
\apptocsubsection{Uniform empirical localization}{app:toc-localization}
\apptocsubsection{An actual empirical-GD trajectory bound}{app:toc-trajectory}
\apptocsubsection{Frozen score variance and nonlinear response}{app:toc-score-response}
\apptocsubsection{Unconditional estimation-risk consequence}{app:toc-estimation-risk}
\apptocsection{Multiclass Decision Geometry and Metric Transfer}{app:decision-geometry}
\apptocsubsection{Gaussian pairwise boundary estimates}{app:toc-boundary}
\apptocsubsection{Angular energy of the frozen empirical score}{app:toc-angular-score}
\apptocsubsection{Transfer to random iterates}{app:toc-random-transfer}
\apptocsection{Proofs of the Full-Batch Scaling Theorems}{app:scaling-proofs}
\apptocsection{Auxiliary Lemmas}{app:auxiliary}
\apptocsubsection{Softmax Geometry}{app:toc-softmax}
\apptocsubsection{Gaussian Moment Bounds}{app:toc-gaussian-moments}
\apptocsubsection{Bounded Population Sublevel Sets}{app:toc-sublevels}
\apptocsubsection{Gaussian Stein Identity}{app:toc-stein}
\apptocsubsection{Fisher Geometry}{app:toc-fisher}
\apptocsubsection{Finite-Rank Fisher Correction}{app:toc-fisher-correction}
\apptocsubsection{Quasi-Fisher Noise Covariance}{app:toc-quasi-fisher}
\apptocsubsection{Local Risk Geometry}{app:toc-risk-geometry}
\apptocsubsection{Global Softmax Remainder}{app:toc-global-remainder}
\apptocsubsection{Stein-Structured Softmax Remainder}{app:toc-stein-remainder}
\apptocsubsection{Uniform Ridge Source Filters}{app:toc-ridge}
\apptocsubsection{Gaussian-Sketch Spectrum}{app:toc-spectrum}
\apptocsubsection{Gaussian-Sketch Effective Dimension}{app:toc-effective-dimension}
\apptocsection{Synthetic-Experiment Details}{app:additional-experiments}
\endgroup

\vspace{0.6em}
\hrule

\clearpage

%%%%%%%%%%%%%%%%%%%
%%%%%%%%%%%%%%%%%%%
\counterwithin{figure}{section}
\counterwithin{table}{section}

\section{Preliminary Geometry and Risk Decomposition}
\label{app:preliminary}

\subsection{Deterministic learning-rate filters}
\label{app:toc-filters}

\begin{lemma}[Product and convolution bounds]
\label{lem:general-step-filter}
Let $0<\eta_t\le\eta_0$ and $H_t=\sum_{u=1}^t\eta_u$. For
$0\le s<t\le T$ and $\mu\ge0$ such that $\eta_u\mu\le c_0<1$ for
$s<u\le t$, put $H_{s:t}=H_t-H_s$. Then
\[
 e^{-\mu H_{s:t}/(1-c_0)}
 \le\prod_{u=s+1}^{t}(1-\eta_u\mu)
 \le e^{-\mu H_{s:t}},
\]
and consequently
\[
 1-\prod_{u=s+1}^{t}(1-\eta_u\mu)
 \asymp\min\{1,\mu H_{s:t}\}.
\]
For every fixed $0\le\alpha<1$, with constants depending only on
$\alpha$ and $\eta_0$,
\[
 \sum_{s=0}^{T-1}\eta_{s+1}
 \frac{(1+H_s)^{-\alpha}}{1+H_T-H_{s+1}}
 \lesssim(1+H_T)^{-\alpha}\log(2+H_T).
\]
\end{lemma}

\begin{proof}
For $0\le x\le c_0$, the inequalities
$-x/(1-c_0)\le\log(1-x)\le-x$ give the product bounds by summing.
The filter estimate follows from these bounds and
$1-e^{-x}\asymp\min\{1,x\}$.

For the convolution, let $h=H_T$ and
$g(x)=(1+x)^{-\alpha}/(1+h-x)$. On each interval
$[H_s,H_{s+1}]$, the two factors change by at most fixed ratios depending
on $\eta_0$, so the summand is at most a fixed multiple of
$\int_{H_s}^{H_{s+1}}g(x)\,dx$. Splitting the integral at $h/2$ gives
\[
 \int_0^{h/2}g(x)\,dx\lesssim(1+h)^{-\alpha},\qquad
 \int_{h/2}^{h}g(x)\,dx
 \lesssim(1+h)^{-\alpha}\log(2+h),
\]
where the first estimate uses $\alpha<1$. This proves the claim.
\end{proof}

\subsection{High-probability Gaussian-sketch events}
\label{app:sketch-events}

This subsection makes explicit every component of the sketch event used in
the main theorems.  Work in an eigenbasis of $\Sigma$ and write
\[
    W:=\Sigma_S=S\Sigma S^\top,
    \qquad
    B:=S\Sigma^{1/2},
    \qquad
    Q:=A_*\Sigma^{1/2},
    \qquad
    V_M^{\mathrm{LS}}:=QB^\top W^{-1}.
\]
All constants below are fixed independently of $M$ and $d$.

\paragraph{Spectral event.}
Let $\mu_1(W)\ge\cdots\ge\mu_M(W)$ be the eigenvalues of $W$.  With the
constants supplied by Lemma~\ref{lem:sketch-spectrum}, define
\[
    \mathcal G_M^{\mathrm{spec}}
    :=
    \left\{
        c_{\mathrm{spec}}j^{-a}
        \le \mu_j(W)
        \le C_{\mathrm{spec}}j^{-a}
        \quad\text{for every }1\le j\le M
    \right\}.
\]
This event transfers the population covariance power law to the sketched
covariance, uniformly over all $M$ sketch directions.

\paragraph{Trace event.}
For a fixed sufficiently large $C_\Sigma$, define
\[
    \mathcal G_M^{\mathrm{trace}}
    :=
    \left\{\operatorname{tr}(W)\le C_\Sigma\right\}.
\]
It supplies the dimension-free moment bounds used in the local softmax
geometry and empirical-process arguments.

\paragraph{Uniform ridge-source event.}
Set $q=(b-1)/a$ and, for $u>0$, define
\[
\begin{aligned}
    K_W(u)
    &:=\operatorname{tr}\!\left[
       V_M^{\mathrm{LS}}uW(W+uI)^{-1}
       (V_M^{\mathrm{LS}})^\top
    \right],\\
    F_W(u)
    &:=\left\|
       V_M^{\mathrm{LS}}\mathbf 1_{(0,u]}(W)
    \right\|_W^2.
\end{aligned}
\]
Fix the constants $D_0,u_0,c_{\mathrm r},C_{\mathrm r}>0$ and a
sufficiently large fixed $L$ from
Lemma~\ref{lem:uniform-ridge-source-filter}, and define
\[
\begin{aligned}
    \mathcal G_M^{\mathrm{ridge}}
    :=\bigl\{\;&
       c_{\mathrm r}u^q\le K_W(u)\le C_{\mathrm r}u^q
       &&\text{for all }u\in[D_0M^{-a},u_0],\\
    &  F_W(u)\le C_{\mathrm r}u^q
       &&\text{for all }u\in[D_0M^{-a},u_0],\\
    &  F_W(u)\ge c_{\mathrm r}u^q
       &&\text{for all }u\in[LD_0M^{-a},u_0]
       \;\bigr\}.
\end{aligned}
\]
Thus this event records, uniformly over the full admissible ridge window,
the source energy visible at the early-stopping scale.

\paragraph{Approximation event.}
Fix a sufficiently small constant $\theta\in(0,1)$, as specified by the
head--tail event in the proof of Lemma~\ref{lem:uniform-ridge-source-filter},
put $k=\lfloor\theta M\rfloor$, and define
\[
    H:=\{1,\ldots,k\},
    \qquad
    T_\theta:=\{k+1,\ldots,d\},
    \qquad
    I_M:=\{\lfloor M/2\rfloor+1,\ldots,2M\}.
\]
Let $S_H$ and $S_{T_\theta}$ be the corresponding column blocks of $S$,
let $\Sigma_{T_\theta}:=\operatorname{diag}(\lambda_j:j\in T_\theta)$,
and set
\[
    B_I:=S_{I_M}\Lambda_{I_M}^{1/2},
    \qquad
    Q_I:=Q_{:,I_M},
    \qquad
    P_{B_I}:=B_I^\top(B_IB_I^\top)^{-1}B_I.
\]
The fixed-teacher approximation event used in
Appendix~\ref{app:approx-scaling} is
\[
\begin{aligned}
    \mathcal G_M^{\mathrm{approx}}
    :=\biggl\{\;&
       c_{\mathrm h}I_k\preceq S_H^\top S_H
       \preceq C_{\mathrm h}I_k,\\
    &  \|S_{T_\theta}\Sigma_{T_\theta}S_{T_\theta}^\top\|_{\mathrm{op}}
       \le C_{\mathrm t}\!\left(
          \frac{\sum_{j>k}\lambda_j}{M}
          +\lambda_{k+1}
          +\sqrt{\frac{\sum_{j>k}\lambda_j^2}{M}}
       \right),\\
    &  \|Q_I(I-P_{B_I})\|_F^2
       \ge c_{\mathrm a}\|Q_I\|_F^2
       \;\biggr\}.
\end{aligned}
\]
The first two conditions give the approximation upper bound, while the
last is the fixed-teacher random-row-space condition giving the matching
lower bound.  In particular, this last condition is not implied by the
spectral or ridge event.

Finally, set $\mathcal G_M:=\mathcal G_M^{\mathrm{spec}}\cap
\mathcal G_M^{\mathrm{trace}}$ and define the joint event
\[
    \boxed{
    \mathcal G_M^{\mathrm{all}}
    :=\mathcal G_M^{\mathrm{spec}}
      \cap\mathcal G_M^{\mathrm{trace}}
      \cap\mathcal G_M^{\mathrm{ridge}}
      \cap\mathcal G_M^{\mathrm{approx}}
    =\mathcal G_M
      \cap\mathcal G_M^{\mathrm{ridge}}
      \cap\mathcal G_M^{\mathrm{approx}}.
    }
\]
Results using only local risk geometry retain the weaker event
$\mathcal G_M$; ridge-source results use
$\mathcal G_M\cap\mathcal G_M^{\mathrm{ridge}}$; and the final theorems,
which also invoke $\Delta_M\asymp M^{1-b}$, are stated on
$\mathcal G_M^{\mathrm{all}}$.

To verify the trace bound, write the rows of $S$ as
$s_i^\top=M^{-1/2}g_i^\top$, where $g_i\sim\mathcal N(0,I)$ are
independent. In an eigenbasis of $\Sigma$,
\[
    \operatorname{tr}(\Sigma_S)
    =
    \frac1M\sum_{i=1}^M\sum_{j=1}^{d}\lambda_j g_{ij}^2.
\]
For $0<\theta<(2\|\Sigma\|_{\mathrm{op}})^{-1}$, the Gaussian
chi-square moment-generating function and $-\log(1-x)\le x/(1-x)$ give
\[
    \log\mathbb E\exp\!\left(
        \theta M\operatorname{tr}(\Sigma_S)
    \right)
    =
    -\frac M2\sum_{j=1}^{d}\log(1-2\theta\lambda_j)
    \le
    \frac{M\theta\operatorname{tr}(\Sigma)}
         {1-2\theta\|\Sigma\|_{\mathrm{op}}}.
\]
Chernoff's inequality therefore gives, for a sufficiently large constant
$C_\Sigma$ depending only on $\operatorname{tr}(\Sigma)$ and
$\|\Sigma\|_{\mathrm{op}}$,
\[
    \mathbb P\!\left(\operatorname{tr}(\Sigma_S)>C_\Sigma\right)
    \le e^{-c_\Sigma M}.
\]
Thus the trace event defined above has failure probability at most
$e^{-c_\Sigma M}$.  Intersecting it with the spectral event of
Lemma~\ref{lem:sketch-spectrum} preserves the stated high-probability
form of $\mathcal G_M$.

Under Assumptions~\ref{ass:covariance}--\ref{ass:source}, the
spectral and trace properties defining $\mathcal G_M$ hold with probability
\[
    \mathbb P_S(\mathcal G_M)
    \ge
    1-\exp(-cM)
\]
for some constant $c>0$.  The four displayed event definitions are
consequences of the corresponding Gaussian-sketch lemmas rather than
additional assumptions: Lemma~\ref{lem:sketch-spectrum} proves the
spectral event, Lemma~\ref{lem:uniform-ridge-source-filter} proves the
ridge event, and Appendix~\ref{app:approx-scaling} proves the approximation
event.  Each has failure probability at most $Ce^{-cM}$, so a finite union
bound gives
\[
    \mathbb P_S(\mathcal G_M^{\mathrm{all}})
    \ge 1-Ce^{-cM}.
\]

\subsection{Oracle-aware logit-error decomposition}
\label{app:toc-oracle}

Fix a sketch realization $S$ and distinguish the two population oracles
\[
    V_M^*\in\arg\min_{V\in\mathbb R^{r\times M}}\mathcal R_S(V),
    \qquad
    V_M^{\mathrm{LS}}Sx=\mathbb E[A_*x\mid Sx].
\]
The first is the population cross-entropy optimum in the sketched softmax
class, whereas the second is the conditional-mean linear predictor of the
teacher logits. Define their oracle shift by
\[
    H_M:=V_M^*-V_M^{\mathrm{LS}},
    \qquad W:=S\Sigma S^\top.
\]
Gaussian conditional regression gives
\[
    A_*x=V_M^{\mathrm{LS}}Sx+\xi_M,
    \qquad \xi_M\perp Sx,
    \qquad
    \Delta_M:=\mathbb E_x\|\xi_M\|_2^2.
\]
Here and below, $\mathbb E_x$ denotes expectation over an independent test
input, conditional on the sketch and the training data. Thus the empirical
iterate is fixed inside this expectation. Combining this regression identity
with the exact empirical-GD decomposition from
Appendix~\ref{app:fullbatch-empirical},
\[
    V_T-V_M^*=B_T^{\mathrm{fr}}+Z_T+D_T,
\]
gives the exact logit-error decomposition
\[
\boxed{
V_TSx-A_*x
=-\xi_M+
\bigl(H_M+B_T^{\mathrm{fr}}+Z_T+D_T\bigr)Sx.
}
\tag{A.1}\label{eq:oracle-aware-logit-decomposition}
\]
The five terms are, respectively, the irreducible sketch residual, the
softmax oracle shift, frozen optimization bias, filtered empirical-score
fluctuation, and nonlinear empirical-GD remainder. In particular, only the
first term is orthogonal to the entire student-observable component.

Let $\|E\|_W^2:=\operatorname{tr}(EWE^\top)$. Conditional orthogonality
therefore gives the exact squared-logit identity
\[
\boxed{
\mathbb E_x\|V_TSx-A_*x\|_2^2
=\Delta_M+
\bigl\|H_M+B_T^{\mathrm{fr}}+Z_T+D_T\bigr\|_W^2.
}
\tag{A.2}\label{eq:oracle-aware-squared-logit}
\]
Writing $R_T:=H_M+B_T^{\mathrm{fr}}+Z_T+D_T$, the second term expands as
\[
\begin{aligned}
\|R_T\|_W^2
={}&\|H_M\|_W^2+\|B_T^{\mathrm{fr}}\|_W^2
  +\|Z_T\|_W^2+\|D_T\|_W^2\\
&+2\langle H_M,B_T^{\mathrm{fr}}\rangle_W
 +2\langle H_M,Z_T\rangle_W
 +2\langle H_M,D_T\rangle_W\\
&+2\langle B_T^{\mathrm{fr}},Z_T\rangle_W
 +2\langle B_T^{\mathrm{fr}},D_T\rangle_W
 +2\langle Z_T,D_T\rangle_W.
\end{aligned}
\tag{A.3}\label{eq:oracle-aware-cross-terms}
\]
This is a pathwise identity conditional on the training data. Even after
averaging over that data, the nonlinear remainder and $Z_T$ are not
independent, so the $\langle Z_T,D_T\rangle_W$ term cannot be discarded.
The only cancellation used later is the centeredness of $Z_T$ against terms
that are fixed conditional on $S$.

This is an exact decomposition of logit prediction error, not an exact
additive decomposition of cross-entropy risk. If
$\Phi(h)=\log\sum_c\exp((Uh)_c)$, then the conditional-label identity gives
\[
    \mathcal E_{\mathrm{CE}}(V_T)
    =\mathbb E_x D_\Phi(V_TSx,A_*x).
\]
The global Bregman upper bound and the local Fisher-geometry lower bound
relate this quantity to squared logit error in the regimes used in the
proofs, but the varying softmax curvature prevents identifying the two
expressions term by term.

\begin{remark}[Oracle shift and softmax misspecification]
\label{rem:oracle-shift}
For squared-loss linear regression of $A_*x$ on $Sx$, the population-optimal
linear predictor coincides with $V_M^{\mathrm{LS}}$ under the present
Gaussian setup. In sketched softmax regression, the two oracles generally
differ because marginalizing the teacher probabilities does not preserve the
softmax-linear form. By Lemma~\ref{lem:oracle-compatibility},
\[
    \|H_M\|_W
    =\|V_M^*-V_M^{\mathrm{LS}}\|_W
    \lesssim\Delta_M.
\]
Consequently, its squared prediction contribution is
$O(\Delta_M^2)$, lower order than the leading approximation scale
$\Delta_M$. The classical risk-level approximation--bias--variance
identity below remains valid, but its variance term is not the frozen-score
variance analyzed in Appendix~\ref{app:fullbatch-empirical}. The
oracle-aware decomposition instead keeps the nonlinear oracle shift and GD
remainder explicit.
\end{remark}

\paragraph{Supplementary risk-level identity.}
Let
\[
    \bar V_T:=\mathbb E[V_T\mid S].
\]
All expectations in the following display are conditional on $S$. Convexity
gives the exact cross-entropy identity
\[
\begin{aligned}
\mathbb E[\mathcal E_{\mathrm{CE}}(V_T)\mid S]
={}&\underbrace{\mathcal R_S(V_M^*)-\mathcal R_*}_{\mathrm{Approx}_M}
+\underbrace{\mathcal R_S(\bar V_T)-\mathcal R_S(V_M^*)}_{\mathrm{Bias}_{\mathrm{op}}}\\
&+\underbrace{\mathbb E[\mathcal R_S(V_T)-\mathcal R_S(\bar V_T)\mid S]}_{\mathrm{Var}_{\mathrm{op}}}.
\end{aligned}
\]
The three terms are nonnegative, but this convex-risk identity neither
resolves the logit-error components in \eqref{eq:oracle-aware-logit-decomposition}
nor identifies $\mathrm{Var}_{\mathrm{op}}$ with the frozen empirical-score
variance.

\subsection{Pairwise contrast frame and angular energy}
\label{app:toc-frame}

For each pair $c<d$, set
\[
    a_{cd}:=U^\top(e_c-e_d)\in\mathbb R^r.
\]
The complete-graph contrast identity is
\[
    \boxed{\sum_{c<d}a_{cd}a_{cd}^\top=CI_r.}
\]
Indeed, $\sum_{c<d}(e_c-e_d)(e_c-e_d)^\top=C I_C-\mathbf1_C\mathbf1_C^\top$;
conjugating by $U$ gives the identity. Consequently, for every covariance
eigenvector $v_j$,
\[
    \boxed{
    \sum_{c<d}\lambda_j|a_{cd}^\top A_*v_j|^2
    =C\lambda_j\|A_*v_j\|_2^2\asymp j^{-b}.
    }
\]
Thus the aggregate pairwise teacher-source spectrum has the same source
exponent as the contrast-space prediction spectrum.

For $u,v\in\mathbb R^d$, write
$\langle u,v\rangle_\Sigma:=u^\top\Sigma v$ and
$\|u\|_\Sigma^2:=\langle u,u\rangle_\Sigma$. The teacher and student
pairwise normals are
\[
    w_{cd}^*:=A_*^\top a_{cd},
    \qquad
    w_{cd}(V):=S^\top V^\top a_{cd}.
\]
Under Assumption~\ref{ass:decision-nondegeneracy},
$\|w_{cd}^*\|_\Sigma$ is bounded above and below uniformly over the finite
set of class pairs. Define the $\Sigma$-orthogonal projection onto the
complement of $w_{cd}^*$ by
\[
    \Pi_{cd}^\perp u
    :=u-w_{cd}^*\frac{\langle w_{cd}^*,u\rangle_\Sigma}
    {\|w_{cd}^*\|_\Sigma^2}.
\]
The displayed formula defines a tangent seminorm for every $V$. In the
local neighborhood where each student pairwise normal is positively
aligned with its teacher normal, its value is the aggregate squared angular
prediction error
\[
    \boxed{
    \mathfrak A(V)^2
    :=\sum_{c<d}
    \left\|\Pi_{cd}^\perp
    \bigl(w_{cd}(V)-w_{cd}^*\bigr)\right\|_\Sigma^2.
    }
\]
On this neighborhood, the fixed upper and lower bounds on teacher-normal
lengths make $\mathfrak A(V)^2$ equivalent to the sum of squared local
pairwise angle changes. The positive-alignment restriction is essential:
the displayed projection alone cannot distinguish a positive radial
rescaling from a sign reversal. In the stochastic analysis, the complement
of this local event is handled by the moment bounds in Appendices C--E.

\section{Proofs for the Approximation Error}
\label{app:approximation}

The approximation error depends only on the teacher distribution, the
Gaussian sketch, and the sketched model class. It is therefore
algorithm-independent and applies equally to population and empirical
full-batch gradient descent. Recall that
\[
    V_M^*
    \in
    \arg\min_{V\in\mathbb{R}^{r\times M}}
    \mathcal{R}_S(V),
\]
and define
\[
    \operatorname{Approx}_M
    :=
    \mathcal{R}_S(V_M^*)-\mathcal{R}_*.
\]

Throughout this section, let
\[
    \Phi(h)
    :=
    \log\!\left(
        \sum_{c=1}^{C}\exp((Uh)_c)
    \right),
    \qquad
    h\in\mathbb{R}^{r}.
\]
The corresponding Bregman divergence is
\[
    D_{\Phi}(h,h_*)
    :=
    \Phi(h)-\Phi(h_*)
    -
    \langle \nabla\Phi(h_*),h-h_*\rangle.
\]
Since
\[
    \nabla\Phi(h_*)
    =
    U^\top
    \operatorname{softmax}(Uh_*),
\]
the conditional excess cross-entropy satisfies
\[
    \mathbb{E}_{y\mid x}
    \left[
        \ell(V;Sx,y)-\ell(A_*;x,y)
    \right]
    =
    D_\Phi(VSx,A_*x).
\]
Hence
\[
    \boxed{
    \operatorname{Approx}_M
    =
    \inf_{V\in\mathbb{R}^{r\times M}}
    \mathbb{E}
    D_\Phi(VSx,A_*x).
    }
\]

We first relate this nonlinear approximation error to the optimal
mean-squared approximation of the teacher logits.

Let
\[
    \Sigma_S
    :=
    S\Sigma S^\top
\]
and define
\[
    V_M^{\mathrm{LS}}
    :=
    A_*\Sigma S^\top\Sigma_S^{-1}.
\]
Since $(x,Sx)$ is jointly Gaussian,
\[
    \mathbb{E}[x\mid Sx]
    =
    \Sigma S^\top\Sigma_S^{-1}Sx,
\]
and therefore
\[
    V_M^{\mathrm{LS}}Sx
    =
    \mathbb{E}[A_*x\mid Sx].
\]
We define the irreducible squared-logit approximation error
\[
    \boxed{
    \Delta_M
    :=
    \mathbb{E}
    \left\|
        A_*x-V_M^{\mathrm{LS}}Sx
    \right\|_2^2.
    }
\]
Equivalently,
\[
    \Delta_M
    =
    \operatorname{tr}
    \left(
        A_*\Sigma_\perp A_*^\top
    \right),
\]
where
\[
    \Sigma_\perp
    :=
    \Sigma
    -
    \Sigma S^\top
    \Sigma_S^{-1}
    S\Sigma.
\]

% ============================================================
\subsection{General Upper Bound}
\label{app:approx-upper}

We first show that the softmax approximation error is no larger, up to a
constant, than the optimal squared-logit approximation error.

\begin{proposition}[General approximation upper bound]
\label{prop:approx-upper}
For every sketch matrix $S$ for which $\Sigma_S$ is invertible,
\[
    \boxed{
    \operatorname{Approx}_M
    \le
    C_{\mathrm{sm}}\Delta_M,
    }
\]
where $C_{\mathrm{sm}}$ depends only on the softmax geometry and the fixed
number of classes.
\end{proposition}

\begin{proof}
For
\[
    p(h):=\operatorname{softmax}(Uh),
\]
the Hessian of $\Phi$ is
\[
    \nabla^2\Phi(h)
    =
    U^\top
    \left(
        \operatorname{Diag}(p(h))
        -
        p(h)p(h)^\top
    \right)
    U.
\]
By Lemma~\ref{lem:softmax-geometry},
\[
    0
    \preceq
    \nabla^2\Phi(h)
    \preceq
    C_{\mathrm{sm}}I_r
    \qquad
    \text{for all }h\in\mathbb{R}^r.
\]
Using the integral representation of the Bregman divergence,
\[
    D_\Phi(h,h_*)
    =
    \int_0^1
    (1-t)
    (h-h_*)^\top
    \nabla^2\Phi
    \bigl(
        h_*+t(h-h_*)
    \bigr)
    (h-h_*)
    \,dt,
\]
we obtain
\[
    D_\Phi(h,h_*)
    \le
    C
    \|h-h_*\|_2^2.
\]

Since $V_M^*$ minimizes the population sketched risk,
\[
\begin{aligned}
    \operatorname{Approx}_M
    &=
    \inf_V
    \mathbb{E}
    D_\Phi(VSx,A_*x)
    \\
    &\le
    \mathbb{E}
    D_\Phi
    \bigl(
        V_M^{\mathrm{LS}}Sx,
        A_*x
    \bigr)
    \\
    &\lesssim
    \mathbb{E}
    \left\|
        V_M^{\mathrm{LS}}Sx-A_*x
    \right\|_2^2
    \\
    &=
    \Delta_M.
\end{aligned}
\]
This proves the claim.
\end{proof}

It is useful to record the exact orthogonal decomposition underlying
$\Delta_M$. Define
\[
    \varepsilon
    :=
    x-\mathbb{E}[x\mid Sx].
\]
Since $(x,Sx)$ is jointly Gaussian,
\[
    \varepsilon
    \perp
    Sx.
\]
Consequently, for any $V\in\mathbb{R}^{r\times M}$,
\[
\begin{aligned}
    \mathbb{E}\|A_*x-VSx\|_2^2
    &=
    \mathbb{E}\|A_*\varepsilon\|_2^2
    +
    \mathbb{E}
    \|
        (V-V_M^{\mathrm{LS}})Sx
    \|_2^2
    \\
    &=
    \Delta_M
    +
    \mathbb{E}
    \|
        (V-V_M^{\mathrm{LS}})Sx
    \|_2^2.
\end{aligned}
\tag{B.1}\label{B.1}
\]
In particular,
\[
    \boxed{
    \mathbb{E}\|A_*x-VSx\|_2^2
    \ge
    \Delta_M
    \qquad
    \text{for every }V.
    }
\]

% ============================================================
\subsection{General Lower Bound}
\label{app:approx-lower}

Unlike the upper bound, a global quadratic lower bound for the softmax
Bregman divergence does not hold because the softmax Hessian degenerates
when the logits become saturated. We therefore combine boundedness of the
population-optimal sketched predictor with a truncated strong-convexity
argument.

\begin{proposition}[General approximation lower bound]
\label{prop:approx-lower}
Suppose that $\|A_*\Sigma^{1/2}\|_{\mathrm{op}}\le R$. Then, on the Gaussian-sketch event
$\mathcal G_M$, there exists a constant $c_{\mathrm{app}}>0$, independent
of $M$ and $N$, such that
\[
    \boxed{
    \operatorname{Approx}_M
    \ge
    c_{\mathrm{app}}\Delta_M.
    }
\]
\end{proposition}

\begin{proof}
Let
\[
    h_*
    :=
    A_*x,
    \qquad
    h_M
    :=
    V_M^*Sx,
    \qquad
    \delta
    :=
    h_M-h_*.
\]

By the bounded teacher prediction condition,
\[
    \mathbb{E}\|h_*\|_2^2
    =
    \|A_*\Sigma^{1/2}\|_F^2
    \le C.
\]
Moreover, Lemma~\ref{lem:bounded-sketch-optimum} gives
\[
    \|V_M^*\Sigma_S^{1/2}\|_F
    \le C,
\]
and hence
\[
    \mathbb{E}\|h_M\|_2^2
    \le C.
\]

Choose a sufficiently large constant $L>0$ and define
\[
    \mathcal E_L
    :=
    \{
        \|h_*\|_2\le L,
        \;
        \|h_M\|_2\le L
    \}.
\]
Since both $h_*$ and $h_M$ are fixed-dimensional Gaussian vectors with
uniformly bounded second moments, $L$ can be chosen independently of $M$
such that
\[
    \mathbb{P}(\mathcal E_L^c)
    \le
    \varepsilon_L,
\]
where $\varepsilon_L>0$ is arbitrarily small.

For any $t\in[0,1]$, on $\mathcal E_L$,
\[
    \|
        h_*+t(h_M-h_*)
    \|_2
    \le L.
\]
By Lemma~\ref{lem:softmax-geometry}, the softmax Hessian is uniformly
positive definite on this bounded contrast-logit set:
\[
    \nabla^2\Phi
    \bigl(
        h_*+t(h_M-h_*)
    \bigr)
    \succeq
    c_L I_r.
\]
Hence the Bregman integral representation yields
\[
    D_\Phi(h_M,h_*)
    \ge
    c_L
    \|h_M-h_*\|_2^2
    \qquad
    \text{on }\mathcal E_L.
\]
Therefore,
\[
    \operatorname{Approx}_M
    \ge
    c_L
    \mathbb{E}
    \left[
        \|\delta\|_2^2
        \mathbf{1}_{\mathcal E_L}
    \right].
    \tag{B.2}\label{B.2}
\]

Since $\delta$ is also a fixed-dimensional centered Gaussian vector,
Lemma~\ref{lem:gaussian-fourth-moment} gives
\[
    \mathbb{E}\|\delta\|_2^4
    \le
    C_r
    \left(
        \mathbb{E}\|\delta\|_2^2
    \right)^2.
\]
By Cauchy--Schwarz,
\[
\begin{aligned}
    \mathbb{E}
    \left[
        \|\delta\|_2^2
        \mathbf{1}_{\mathcal E_L^c}
    \right]
    &\le
    \left(
        \mathbb{E}\|\delta\|_2^4
    \right)^{1/2}
    \mathbb{P}(\mathcal E_L^c)^{1/2}
    \\
    &\le
    C_r
    \varepsilon_L^{1/2}
    \mathbb{E}\|\delta\|_2^2.
\end{aligned}
\]
Choosing $L$ sufficiently large so that
\[
    C_r\varepsilon_L^{1/2}
    \le
    \frac12,
\]
we obtain
\[
    \mathbb{E}
    \left[
        \|\delta\|_2^2
        \mathbf{1}_{\mathcal E_L}
    \right]
    \ge
    \frac12
    \mathbb{E}\|\delta\|_2^2.
    \tag{B.3}\label{B.3}
\]
Combining \eqref{B.2} and \eqref{B.3},
\[
    \operatorname{Approx}_M
    \gtrsim
    \mathbb{E}
    \|A_*x-V_M^*Sx\|_2^2.
\]

Finally, applying the orthogonal decomposition \eqref{B.1} with
$V=V_M^*$ gives
\[
    \mathbb{E}
    \|A_*x-V_M^*Sx\|_2^2
    \ge
    \Delta_M.
\]
Therefore
\[
    \operatorname{Approx}_M
    \gtrsim
    \Delta_M,
\]
as claimed.
\end{proof}

Combining Propositions~\ref{prop:approx-upper} and
\ref{prop:approx-lower}, we obtain the general equivalence
\[
    \boxed{
    \operatorname{Approx}_M
    \asymp
    \Delta_M.
    }
    \tag{B.4}\label{B.4}
\]

% ============================================================
\subsection{Power-Law Scaling}
\label{app:approx-scaling}

We now evaluate $\Delta_M$ under
Assumptions~\ref{ass:covariance} and~\ref{ass:source}.

Write
\[
    x=\Sigma^{1/2}g,
    \qquad
    g\sim\mathcal N(0,I),
\]
and define
\[
    B:=S\Sigma^{1/2},
    \qquad
    Q:=A_*\Sigma^{1/2}.
\]
Then
\[
    Sx=Bg.
\]
The conditional mean of $g$ given $Bg$ is
\[
    \mathbb{E}[g\mid Bg]
    =
    P_Bg,
\]
where
\[
    P_B
    :=
    B^\top(BB^\top)^{-1}B
\]
is the orthogonal projector onto the row space of $B$. Hence
\[
    \boxed{
    \Delta_M
    =
    \|Q(I-P_B)\|_F^2.
    }
    \tag{B.5}\label{B.5}
\]

Let
\[
    q_j
    :=
    Qv_j
    =
    \sqrt{\lambda_j}\,A_*v_j.
\]
The source condition becomes
\[
    \boxed{
    \|q_j\|_2^2
    =
    \lambda_j\|A_*v_j\|_2^2
    \asymp
    j^{-b}.
    }
\]

We prove matching upper and lower bounds for \eqref{B.5}.

\paragraph{Upper bound.}
Fix the sufficiently small constant $\theta\in(0,1)$ used in the
definition of $\mathcal G_M^{\mathrm{approx}}$ and set
\[
    k:=\lfloor\theta M\rfloor.
\]
Work in the eigenbasis of $\Sigma$ and split
\[
    x=(x_H,x_T),
    \qquad
    H:=\{1,\ldots,k\},
    \qquad
    T:=\{k+1,k+2,\ldots\}.
\]
Correspondingly, write
\[
    S=(S_H,S_T),
    \qquad
    A_*=(A_H,A_T).
\]

Since $S_H\in\mathbb{R}^{M\times k}$ is a tall Gaussian matrix, standard
Gaussian singular-value concentration gives, with probability at least
$1-\exp(-cM)$,
\[
    cI_k
    \preceq
    S_H^\top S_H
    \preceq
    CI_k.
    \tag{B.6}\label{B.6}
\]
Define the stable left inverse
\[
    L_H
    :=
    (S_H^\top S_H)^{-1}S_H^\top,
\]
so that
\[
    L_HS_H=I_k,
    \qquad
    \|L_H\|_{\mathrm{op}}\lesssim1.
    \tag{B.7}\label{B.7}
\]

Consider the candidate linear predictor
\[
    \widehat h(Sx)
    :=
    A_HL_HSx.
\]
Since
\[
    Sx
    =
    S_Hx_H+S_Tx_T,
\]
we have
\[
    \widehat h(Sx)
    =
    A_Hx_H
    +
    A_HL_HS_Tx_T.
\]
Hence
\[
    A_*x-\widehat h(Sx)
    =
    A_Tx_T
    -
    A_HL_HS_Tx_T,
\]
and therefore
\[
\begin{aligned}
    \Delta_M
    &\le
    \mathbb{E}
    \|A_*x-\widehat h(Sx)\|_2^2
    \\
    &\lesssim
    \operatorname{tr}(A_T\Sigma_TA_T^\top)
    +
    \operatorname{tr}
    \left(
        A_HL_H
        S_T\Sigma_TS_T^\top
        L_H^\top A_H^\top
    \right).
\end{aligned}
\]

The first term is the teacher spectral tail:
\[
\begin{aligned}
    \operatorname{tr}(A_T\Sigma_TA_T^\top)
    &=
    \sum_{j>k}
    \lambda_j\|A_*v_j\|_2^2
    \\
    &\asymp
    \sum_{j>k}j^{-b}
    \asymp
    k^{1-b}.
\end{aligned}
\tag{B.8}\label{B.8}
\]

For the second term, using \eqref{B.7},
\[
\begin{aligned}
    &
    \operatorname{tr}
    \left(
        A_HL_H
        S_T\Sigma_TS_T^\top
        L_H^\top A_H^\top
    \right)
    \\
    &\qquad
    \lesssim
    \|A_H\|_F^2
    \,
    \|S_T\Sigma_TS_T^\top\|_{\mathrm{op}}.
\end{aligned}
\tag{B.9}\label{B.9}
\]
The simultaneous Gaussian head--tail event constructed in the proof of
Lemma~\ref{lem:uniform-ridge-source-filter} gives, for this choice of
$k$, $\|S_T\Sigma_TS_T^\top\|_{\mathrm{op}}\lesssim k^{-a}$ with
probability at least $1-Ce^{-cM}$. Since
$\lambda_{k+1}\gtrsim k^{-a}$, this implies
\[
    \|S_T\Sigma_TS_T^\top\|_{\mathrm{op}}
    \lesssim
    \frac{\sum_{j>k}\lambda_j}{M}
    +
    \lambda_{k+1}
    +
    \sqrt{
        \frac{\sum_{j>k}\lambda_j^2}{M}
    }
    \tag{B.10}\label{B.10}
\]
on the same event.

Under $\lambda_j\asymp j^{-a}$,
\[
    \sum_{j>k}\lambda_j
    \asymp
    k^{1-a},
    \qquad
    \lambda_{k+1}
    \asymp
    k^{-a},
    \qquad
    \sum_{j>k}\lambda_j^2
    \asymp
    k^{1-2a}.
\]
Moreover,
\[
\begin{aligned}
    \|A_H\|_F^2
    &=
    \sum_{j\le k}
    \|A_*v_j\|_2^2
    \\
    &\asymp
    \sum_{j\le k}
    j^{a-b}
    \asymp
    k^{a-b+1},
\end{aligned}
\tag{B.11}\label{B.11}
\]
where the final equivalence uses $b<a+1$.

Combining \eqref{B.9}--\eqref{B.11},
\[
\begin{aligned}
    &
    \operatorname{tr}
    \left(
        A_HL_H
        S_T\Sigma_TS_T^\top
        L_H^\top A_H^\top
    \right)
    \\
    &\qquad
    \lesssim
    \frac{k^{2-b}}{M}
    +
    k^{1-b}
    +
    \frac{k^{3/2-b}}{\sqrt{M}}.
\end{aligned}
\]
Since $k\asymp M$, all three terms are of order $M^{1-b}$. Together with
\eqref{B.8},
\[
    \boxed{
    \Delta_M
    \lesssim
    M^{1-b}.
    }
    \tag{B.12}\label{B.12}
\]

\paragraph{Lower bound.}
To obtain a deterministic-teacher lower bound, we use a dyadic spectral
block. Let
\[
    I_M
    :=
    \{\lfloor M/2\rfloor+1,\ldots,2M\},
\]
which contains $L_M:=2M-\lfloor M/2\rfloor$ coordinates. In particular,
$M/2\le L_M-M\le M$ and $M<L_M\le2M$. Restrict $B$ and $Q$ to this block:
\[
    B_I
    :=
    S_I\Lambda_I^{1/2},
    \qquad
    Q_I
    :=
    A_I\Lambda_I^{1/2}.
\]

Recall that
\[
    \Delta_M
    =
    \mathbb{E}
    \left\|
        Qg-\mathbb{E}[Qg\mid Bg]
    \right\|_2^2.
\]
Giving the learner the complement $g_{I_M^c}$ can only decrease the
conditional mean-squared error. Therefore,
\[
\begin{aligned}
    \Delta_M
    &\ge
    \mathbb{E}
    \left\|
        Q_Ig_I
        -
        \mathbb{E}
        [Q_Ig_I\mid B_Ig_I]
    \right\|_2^2
    \\
    &=
    \|Q_I(I-P_{B_I})\|_F^2,
\end{aligned}
\tag{B.13}\label{B.13}
\]
where
\[
    P_{B_I}
    :=
    B_I^\top
    (B_IB_I^\top)^{-1}
    B_I.
\]

Uniformly for $j\in I_M$, the power-law spectrum satisfies
\[
    \lambda_j
    \asymp
    M^{-a}.
\]
Hence we may write
\[
    \Lambda_I^{1/2}
    =
    M^{-a/2}D_I,
\]
where
\[
    cI
    \preceq
    D_I
    \preceq
    CI.
\]
Consequently,
\[
    \operatorname{row}(B_I)
    =
    \operatorname{row}(S_ID_I).
\]

If $D_I=I$, the row space of
$S_I\in\mathbb{R}^{M\times L_M}$ is a uniformly random
$M$-dimensional subspace of $\mathbb{R}^{L_M}$. For a fixed deterministic
$q\in\mathbb{R}^{L_M}$, the squared relative distance to this row space
has the same law as $X/(X+Y)$, where $X\sim\chi^2_{L_M-M}$ and
$Y\sim\chi^2_M$ are independent. On the event
$X\ge(L_M-M)/2$ and $Y\le2M$, the ratio is at least
\[
\frac{(L_M-M)/2}{(L_M-M)/2+2M}\ge\frac1{10},
\]
using $M/2\le L_M-M\le M$. The chi-square Chernoff bounds give this
event probability at least $1-2e^{-c_1M}$ for a universal $c_1>0$.
Thus we may take $c_0=1/10$ in
\[
    \operatorname{dist}^2
    \left(
        q,
        \operatorname{row}(S_I)
    \right)
    \ge
    c_0\|q\|_2^2
\]
with probability at least $1-2\exp(-c_1M)$. These constants are
uniform in $L_M$ and hence in $M$.

The bounded condition number of $D_I$ transfers this estimate to
$\operatorname{row}(S_ID_I)$. Indeed,
\[
\begin{aligned}
    \operatorname{dist}
    \left(
        q,
        \operatorname{row}(S_ID_I)
    \right)
    &=
    \inf_{w\in\operatorname{row}(S_I)}
    \|q-wD_I\|_2
    \\
    &\ge
    \sigma_{\min}(D_I)
    \operatorname{dist}
    \left(
        qD_I^{-1},
        \operatorname{row}(S_I)
    \right)
    \\
    &\gtrsim
    \|q\|_2.
\end{aligned}
\tag{B.14}\label{B.14}
\]
Since $r$ is fixed, applying \eqref{B.14} simultaneously to the $r$ rows
of $Q_I$ gives
\[
    \boxed{
    \|Q_I(I-P_{B_I})\|_F^2
    \gtrsim
    \|Q_I\|_F^2
    }
    \tag{B.15}\label{B.15}
\]
with probability at least $1-\exp(-cM)$.

Finally, by the source condition and the fact that
$M/2<j\le2M$ on $I_M$,
\[
\begin{aligned}
    \|Q_I\|_F^2
    &=
    \sum_{j=\lfloor M/2\rfloor+1}^{2M}
    \lambda_j\|A_*v_j\|_2^2
    \\
    &\asymp
    \sum_{j=\lfloor M/2\rfloor+1}^{2M}
    j^{-b}
    \\
    &\asymp
    M^{1-b}.
\end{aligned}
\tag{B.16}\label{B.16}
\]
Combining \eqref{B.13}, \eqref{B.15}, and \eqref{B.16},
\[
    \boxed{
    \Delta_M
    \gtrsim
    M^{1-b}.
    }
    \tag{B.17}\label{B.17}
\]

Together with \eqref{B.12},
\[
    \boxed{
    \Delta_M
    \asymp
    M^{1-b}
    }
    \tag{B.18}\label{B.18}
\]
on the intersection of the Gaussian events used in \eqref{B.6},
\eqref{B.10}, and \eqref{B.15}.  Denote this fixed-teacher approximation
event by $\mathcal G_M^{\mathrm{approx}}$.  The tail estimates established
above and a finite union bound give
\[
    \mathbb P_S(\mathcal G_M^{\mathrm{approx}})
    \ge 1-Ce^{-cM}.
\]
In particular, the random-row-space lower bound needed for \eqref{B.17} is
an explicit component of $\mathcal G_M^{\mathrm{approx}}$; it is not
inferred from the sketch-spectrum or ridge-source events.

Combining \eqref{B.4} and \eqref{B.18} proves the desired approximation
scaling law.

\begin{theorem}[Approximation scaling law]
\label{thm:approximation-scaling}
Under Assumptions~\ref{ass:covariance}--\ref{ass:source}, the joint
event $\mathcal G_M\cap\mathcal G_M^{\mathrm{approx}}$ has probability at
least $1-Ce^{-cM}$.  On this event there are constants $c,C>0$,
independent of $M$ and $N$, such that
\[
    \boxed{
    cM^{1-b}
    \le
    \operatorname{Approx}_M
    \le
    CM^{1-b}.
    }
\]
Equivalently,
\[
    \boxed{
    \operatorname{Approx}_M
    \asymp
    M^{1-b}.
    }
\]
\end{theorem}
% ============================================================
\subsection{Compatibility of the conditional-mean oracle}
\label{app:toc-oracle-compatibility}

Let
\[
    \Omega_M:=A_*\Sigma A_*^\top
    -A_*\Sigma S^\top\Sigma_S^{-1}S\Sigma A_*^\top,
    \qquad \Delta_M=\operatorname{tr}(\Omega_M).
\]
The Gaussian conditional-mean representation gives
\[
    A_*x=m+\xi_M,
    \qquad m:=V_M^{\mathrm{LS}}Sx,
    \qquad \xi_M\perp Sx,
    \qquad \xi_M\sim\mathcal N(0,\Omega_M).
\]

\begin{lemma}[Oracle compatibility]
\label{lem:oracle-compatibility}
Suppose $\|A_*\Sigma^{1/2}\|_{\mathrm{op}}\le R$. On $\mathcal G_M$,
\[
    \boxed{
    \|H_M\|_{\Sigma_S}
    \lesssim\Delta_M,
    }
\]
where $H_M:=V_M^*-V_M^{\mathrm{LS}}$ is the oracle shift from
Appendix~\ref{app:preliminary} and
$\|E\|_{\Sigma_S}^2:=\operatorname{tr}(E\Sigma_SE^\top)$.
\end{lemma}

\begin{proof}
Write
\[
    C(h):=U^\top\bigl(\operatorname{Diag}(p(h))-p(h)p(h)^\top\bigr)U.
\]
For fixed $C$, the first and second derivatives of this matrix-valued
softmax covariance map are uniformly bounded on $\mathbb R^r$; this
follows by differentiating the softmax coordinates, whose derivatives are
polynomials in probabilities in $[0,1]$. Taylor's formula at $m$, the
independence and centering of $\xi_M$, and the uniform second-derivative
bound imply
\[
\left\|\mathbb E[C(m+\xi_M)\mid m]-C(m)\right\|_{\mathrm{op}}
\le C\,\mathbb E[\|\xi_M\|_2^2\mid m].
\]
Taking expectations yields
\[
    \|B_*-B(V_M^{\mathrm{LS}})\|_{\mathrm{op}}
    \le C\Delta_M.
\]

By the population Stein identity and
$V_M^{\mathrm{LS}}\Sigma_S=A_*\Sigma S^\top$,
\[
\nabla\mathcal R_S(V_M^{\mathrm{LS}})
=\bigl(B(V_M^{\mathrm{LS}})-B_*\bigr)V_M^{\mathrm{LS}}\Sigma_S.
\]
Moreover,
\[
\|V_M^{\mathrm{LS}}\Sigma_S^{1/2}\|_F^2
=\mathbb E\|\mathbb E[A_*x\mid Sx]\|_2^2
\le\mathbb E\|A_*x\|_2^2\le rR^2.
\]
Thus
\[
\|\nabla\mathcal R_S(V_M^{\mathrm{LS}})\Sigma_S^{-1/2}\|_F
\lesssim\Delta_M.
\]

Both $V_M^*$ and $V_M^{\mathrm{LS}}$ have uniformly bounded prediction
norm. The first follows from Lemma~\ref{lem:bounded-sketch-optimum}; the
second was just shown. The same truncated-Gaussian argument used in that
lemma gives
$\nabla^2\mathcal R_S(V)\succeq c(I_r\otimes\Sigma_S)$ on their joining
segment: the segment has bounded prediction norm, hence the student logits
are bounded on a fixed-probability Gaussian event, where the softmax
Fisher matrix has a uniform positive lower bound. Strong monotonicity on
this segment, together with $\nabla\mathcal R_S(V_M^*)=0$, gives
\[
\begin{aligned}
c\|V_M^{\mathrm{LS}}-V_M^*\|_{\Sigma_S}^2
&\le\left\langle
\nabla\mathcal R_S(V_M^{\mathrm{LS}}),
V_M^{\mathrm{LS}}-V_M^*
\right\rangle\\
&\le\|\nabla\mathcal R_S(V_M^{\mathrm{LS}})\Sigma_S^{-1/2}\|_F
\|V_M^{\mathrm{LS}}-V_M^*\|_{\Sigma_S}.
\end{aligned}
\]
Canceling the final factor proves the claim.
\end{proof}

\begin{lemma}[Angular sketch approximation]
\label{lem:angular-sketch-approximation}
Under Assumption~\ref{ass:decision-nondegeneracy} and the bound
$\|A_*\Sigma^{1/2}\|_{\mathrm{op}}\le R$, on $\mathcal G_M$ and as $M\to\infty$,
\[
    \boxed{
    \sum_{c<d}\left\|\Pi_{cd}^\perp
    \bigl(w_{cd}(V_M^*)-w_{cd}^*\bigr)\right\|_\Sigma^2
    \asymp\Delta_M.
    }
\]
In particular, this quantity is $\asymp M^{1-b}$ under
Theorem~\ref{thm:approximation-scaling}.
\end{lemma}

\begin{proof}
For the conditional-mean oracle, write
\[
    w_{cd}^*=w_{cd}^{\mathrm{LS}}+r_{cd,M},
    \qquad
    w_{cd}^{\mathrm{LS}}:=S^\top(V_M^{\mathrm{LS}})^\top a_{cd}.
\]
Gaussian conditional regression gives
$r_{cd,M}\perp_\Sigma\operatorname{Range}(S^\top)$. Hence
\[
\sum_{c<d}\|r_{cd,M}\|_\Sigma^2
=\sum_{c<d}a_{cd}^\top\Omega_M a_{cd}
=C\operatorname{tr}(\Omega_M)=C\Delta_M.
\]
Orthogonality also yields
$\langle r_{cd,M},w_{cd}^*\rangle_\Sigma=\|r_{cd,M}\|_\Sigma^2$, so
\[
\|\Pi_{cd}^\perp r_{cd,M}\|_\Sigma^2
=\|r_{cd,M}\|_\Sigma^2
-\frac{\|r_{cd,M}\|_\Sigma^4}{\|w_{cd}^*\|_\Sigma^2}.
\]
The pairwise teacher-normal norms are uniformly bounded below, while
$\max_{c<d}\|r_{cd,M}\|_\Sigma^2\le C\Delta_M=o(1)$. Summing therefore
shows that the angular error of $V_M^{\mathrm{LS}}$ is asymptotic to
$C\Delta_M$.

Finally, Lemma~\ref{lem:oracle-compatibility} and the contrast-frame
identity imply
\[
\sum_{c<d}\|w_{cd}(V_M^*)-w_{cd}(V_M^{\mathrm{LS}})\|_\Sigma^2
=C\|V_M^*-V_M^{\mathrm{LS}}\|_{\Sigma_S}^2
=O(\Delta_M^2).
\]
The triangle inequality in the direct sum of pairwise contrast spaces
then shows that replacing $V_M^{\mathrm{LS}}$ by $V_M^*$ changes the
squared angular error by $o(\Delta_M)$. The approximation scaling
$\Delta_M\asymp M^{1-b}$ follows from
Theorem~\ref{thm:approximation-scaling}.
\end{proof}

\section{Full-Batch GD: Population and Frozen Bias}
\label{app:fullbatch-bias}

This appendix separates the nonlinear population-GD bias from the frozen
linear filter used in the empirical-GD decomposition. Throughout this
appendix, condition on a sketch in
$\mathcal G_M\cap\mathcal G_M^{\mathrm{ridge}}$ and set
\[
\begin{aligned}
    W&:=\Sigma_S,\qquad V^*:=V_M^*,\qquad
    G:=\nabla^2\mathcal R_S(V^*),\\
    q&:=\frac{b-1}{a},\qquad
    H_t:=\sum_{u=1}^t\eta_u.
\end{aligned}
\]
For the frozen Fisher dynamics, define
\[
    \Psi(t,s):=\prod_{u=s+1}^{t}(I-\eta_uG),
    \qquad \Psi(s,s):=I,
    \qquad \Psi_t:=\Psi(t,0).
\]

\subsection{Genuine nonlinear population gradient descent}
\label{app:toc-population-bias}

We first record the standard comparator inequality used for both the
population and empirical GD trajectories.

\begin{lemma}[Last-iterate comparator inequality for GD]
\label{lem:gd-comparator}
Let $F$ be a convex, $L$-smooth function on a finite-dimensional Hilbert
space. Consider
\[
    V_t=V_{t-1}-\eta_t\nabla F(V_{t-1}),
    \qquad V_0=0,
\]
where $0<\eta_t\le L^{-1}$, and define
$H_T:=\sum_{t=1}^T\eta_t$. Then, for every fixed comparator $U$,
\[
    F(V_T)-F(U)\le \frac{\|U\|_F^2}{2H_T}.
\]
\end{lemma}

\begin{proof}
By convexity and smoothness,
\[
\begin{aligned}
    F(V_t)-F(U)
    &\le
    \langle \nabla F(V_{t-1}),V_t-U\rangle_F
    +\frac L2\|V_t-V_{t-1}\|_F^2.
\end{aligned}
\]
Using
$\nabla F(V_{t-1})=(V_{t-1}-V_t)/\eta_t$ and the three-point identity,
\[
\begin{aligned}
    F(V_t)-F(U)
    &\le
    \frac{\|V_{t-1}-U\|_F^2-\|V_t-U\|_F^2}{2\eta_t}\\
    &\quad-
    \left(\frac{1}{2\eta_t}-\frac L2\right)
    \|V_t-V_{t-1}\|_F^2\\
    &\le
    \frac{\|V_{t-1}-U\|_F^2-\|V_t-U\|_F^2}{2\eta_t},
\end{aligned}
\]
where the last inequality uses $\eta_t\le L^{-1}$. Multiplying by
$2\eta_t$ and summing yields
\[
    2\sum_{t=1}^T\eta_t\bigl[F(V_t)-F(U)\bigr]
    \le \|V_0-U\|_F^2-\|V_T-U\|_F^2
    \le \|U\|_F^2.
\]
The same step-size condition and the descent lemma make $F(V_t)$
nonincreasing. Hence
\[
    H_T\bigl[F(V_T)-F(U)\bigr]
    \le \sum_{t=1}^T\eta_t\bigl[F(V_t)-F(U)\bigr],
\]
which proves the claim.
\end{proof}

Consider the deterministic recursion
\[
    V_0^{\mathrm{pop}}=0,\qquad
    V_t^{\mathrm{pop}}
    =V_{t-1}^{\mathrm{pop}}-\eta_t\nabla\mathcal R_S(V_{t-1}^{\mathrm{pop}}),
    \qquad t=1,\ldots,T.
\]

\begin{theorem}[Nonlinear population GD bias]
\label{thm:population-gd-bias}
Under the setup of Section~\ref{sec:preliminaries}, suppose
Assumptions~\ref{ass:covariance}--\ref{ass:source} hold and
$M\ge D H_T^{1/a}$. For all sufficiently large $H_T$, on
$\mathcal G_M\cap\mathcal G_M^{\mathrm{ridge}}$,
\[
    \boxed{
    \mathcal R_S(V_T^{\mathrm{pop}})-\mathcal R_S(V^*)
    \asymp
    \|V_T^{\mathrm{pop}}-V^*\|_G^2
    \asymp H_T^{-q}.
    }
\]
The constants are uniform in $M,T$ in this non-saturated range.
\end{theorem}

\begin{proof}
\emph{Upper bound.} For $u\in[D_0M^{-a},u_0]$, define the ridge
comparator
\[
    V_u:=QB^\top(\Sigma_S+uI)^{-1},\qquad B=S\Sigma^{1/2},\quad
    Q=A_*\Sigma^{1/2}.
\]
The singular-value formulas underlying
Lemma~\ref{lem:uniform-ridge-source-filter} give
\[
    \|V_u\|_F^2\lesssim u^{q-1},
    \qquad
    \|V_u-V_M^{\mathrm{LS}}\|_{\Sigma_S}^2\lesssim u^q.
\]
Oracle compatibility gives $\|H_M\|_{\Sigma_S}\lesssim\Delta_M$ by
Lemma~\ref{lem:oracle-compatibility}; hence
\[
    \|V_u-V^*\|_{\Sigma_S}^2\lesssim u^q+\Delta_M^2.
\]
The global softmax-curvature bound gives, for every $V$,
$0\preceq\nabla^2\mathcal R_S(V)
\preceq C_{\mathrm{sm}}(I_r\otimes\Sigma_S)$ and hence
$\|\nabla^2\mathcal R_S(V)\|_{\mathrm{op}}
\le C_{\mathrm{sm}}\|\Sigma_S\|_{\mathrm{op}}$. Its prediction-norm form and
the preceding display imply
$\mathcal R_S(V_u)-\mathcal R_S(V^*)\lesssim u^q+\Delta_M^2$.
Moreover, on the sketch event, $\|\Sigma_S\|_{\mathrm{op}}$ is uniformly bounded,
and the stable learning-rate condition in Section~\ref{subsec:optimization}
makes every GD step no larger than the
reciprocal smoothness constant.

Applying Lemma~\ref{lem:gd-comparator} to $\mathcal R_S$ with comparator
$V_u$ now yields directly
\[
\begin{aligned}
    \mathcal R_S(V_T^{\mathrm{pop}})-\mathcal R_S(V^*)
    &\le \frac{\|V_u\|_F^2}{2H_T}
       +\mathcal R_S(V_u)-\mathcal R_S(V^*)\\
    &\lesssim \frac{u^{q-1}}{H_T}+u^q+\Delta_M^2.
\end{aligned}
\]
Take $u=H_T^{-1}$, which lies in the stated source range
after increasing $D$. Then $M\ge D H_T^{1/a}$ and the approximation estimate give
$\Delta_M\lesssim M^{1-b}\lesssim H_T^{-q}$ and thus
$\Delta_M^2\lesssim H_T^{-q}$. Consequently,
\[
    \mathcal R_S(V_T^{\mathrm{pop}})-\mathcal R_S(V^*)
    \lesssim H_T^{-q}.
\]

\emph{Lower bound.} Let $Wu_j=\mu_j u_j$ and set
$v_{t,j}=V_t^{\mathrm{pop}}u_j$, $v_j^*=V^*u_j$. The exact Stein identity
and stationarity at $V^*$ give
\[
    v_{t+1,j}
    =(I_r-\eta_{t+1}\mu_jB(V_t^{\mathrm{pop}}))v_{t,j}
      +\eta_{t+1}\mu_jB_Mv_j^*.
\]
Every population iterate lies in the initial-risk sublevel set since the
risk decreases. Every population iterate lies in a fixed prediction-norm ball. Restricting
the Gaussian logits to a fixed-probability bounded-logit event and applying
Lemma~\ref{lem:softmax-geometry} therefore gives
$cI_r\preceq B(V_t^{\mathrm{pop}})\preceq CI_r$ uniformly. Choose
$\eta_{\max}$ small enough that all factors are contractions. Variation of
constants can be used without assuming that the time-dependent matrices
$B(V_t^{\mathrm{pop}})$ commute. Namely, put
\[
 A_{t,j}:=I_r-\eta_{t+1}\mu_jB(V_t^{\mathrm{pop}}),\qquad
 \Phi_j(t,s):=A_{t-1,j}\cdots A_{s,j}\quad(t>s),
\]
with $\Phi_j(s,s)=I_r$. The uniform Fisher sandwich gives
$\|\Phi_j(t,s)\|_{\mathrm{op}}
\le\exp[-c\mu_j(H_t-H_s)]$. Since
$v_{0,j}=0$, the exact recursion expands as
\[
 v_{T,j}=\mu_j\sum_{s=0}^{T-1}\eta_{s+1}
      \Phi_j(T,s+1)B_Mv_j^*.
\]
Bounding the weighted sum once by $H_T$ and once by its exponential
convolution gives
\[
    \|v_{T,j}\|_2
    \le C\min\{1,H_T\mu_j\}\|v_j^*\|_2.
\]
This calculation uses the exact nonlinear Stein recursion and a uniform
matrix sandwich only; it makes neither a separability assumption nor a
time-independent-Hessian assumption.
For $\mu_j\le\epsilon/H_T$, choose fixed $\epsilon$ sufficiently small
to obtain $\|v_{T,j}-v_j^*\|_2\ge\frac12\|v_j^*\|_2$. Choose the
constant $D$ in $M\ge D H_T^{1/a}$ so that
$D\ge(LD_0/\epsilon)^{1/a}$, where $L$ is the Tauberian endpoint constant
in Lemma~\ref{lem:uniform-ridge-source-filter}; for sufficiently large
$H_T$, this places $\epsilon/H_T$ in its lower-source interval.
The Stein relation gives $V^*=B_M^{-1}B_*V_M^{\mathrm{LS}}$. The matrices
$B_M$ and $B_*$, and their inverses, are uniformly bounded: for either
matrix, restrict the Gaussian logits to a fixed-probability bounded-logit
event and apply Lemma~\ref{lem:softmax-geometry}. Thus the low-frequency
source energy of $V^*$ is comparable to that of $V_M^{\mathrm{LS}}$. The Tauberian conclusion of
Lemma~\ref{lem:uniform-ridge-source-filter} now yields
\[
\begin{aligned}
    \|V_T^{\mathrm{pop}}-V^*\|_W^2
    &\ge c\sum_{\mu_j\le\epsilon/H_T}
          \mu_j\|v_j^*\|_2^2\\
    &\ge c\|V_M^{\mathrm{LS}}\mathbf1_{(0,\epsilon/H_T]}(W)\|_W^2
     \gtrsim H_T^{-q}.
\end{aligned}
\]
On the bounded population sublevel set, the Hessian sandwich in
Lemma~\ref{lem:bounded-sketch-optimum} (whose proof applies to every
bounded prediction-norm ball) makes excess risk and squared $G$-distance
uniformly equivalent. This proves both lower bounds and completes the
proof.
\end{proof}

\begin{lemma}[Post-cutoff population and frozen-bias bounds]
\label{lem:post-cutoff-bias}
Let $B_T^{\mathrm{fr}}:=-\Psi_TV^*$.
There is a fixed $c_{\mathrm{sat}}>0$ such that, whenever
$H_T\ge c_{\mathrm{sat}}M^a$,
\[
    \mathcal R_S(V_T^{\mathrm{pop}})-\mathcal R_S(V^*)
    \lesssim M^{1-b},
    \qquad
    \|V_T^{\mathrm{pop}}-V^*\|_G^2\lesssim M^{1-b},
\]
and
\[
    \boxed{\|B_T^{\mathrm{fr}}\|_G^2\lesssim M^{1-b}.}
\]
Thus the population parameter bias may decay further after the spectral
cutoff, but it is uniformly below the sketch approximation scale.
\end{lemma}

\begin{proof}
Let $D$ be large enough for Theorem~\ref{thm:population-gd-bias} and
Lemma~\ref{lem:fisher-ridge-frozen-bias}, and choose a fixed
$c_{\mathrm{sat}}>0$ so small that
$D(c_{\mathrm{sat}})^{1/a}<1$.  For large $M$, choose an integer $t_M$
such that
\[
    \frac{c_{\mathrm{sat}}}{2}M^a
    \le H_{t_M}\le c_{\mathrm{sat}}M^a.
\]
Then $M\ge D H_{t_M}^{1/a}$, so the non-saturated results at time
$t_M$ give
\[
\mathcal R_S(V_{t_M}^{\mathrm{pop}})-\mathcal R_S(V^*)
\lesssim H_{t_M}^{-q}\asymp M^{1-b},
\qquad
\|\Psi_{t_M}V^*\|_G^2\lesssim M^{1-b}.
\]
Population GD decreases $\mathcal R_S$ at every step.  Hence the first
risk bound holds for every $T\ge t_M$.  All such iterates remain in the
fixed population sublevel set, where
Lemma~\ref{lem:bounded-sketch-optimum} makes excess risk uniformly
equivalent to squared $G$-distance, proving the second bound.  Finally,
$\Psi(T,t_M)$ is a self-adjoint contraction commuting with $G$, and spectral
calculus gives
\[
    \|\Psi_TV^*\|_G\le \|\Psi_{t_M}V^*\|_G,
    \qquad T\ge t_M.
\]
This proves the frozen-bias assertion.  Enlarging constants covers the
bounded initial values of $M$.
\end{proof}

\begin{lemma}[Population GD localization]
\label{lem:population-gd-localization}
Under the assumptions of Theorem~\ref{thm:population-gd-bias}, for all
$1\le t\le T$ in the non-saturated range,
\[
    \|V_t^{\mathrm{pop}}-V^*\|_G^2\lesssim(1+H_t)^{-q}.
\]
Consequently, population GD enters any fixed local Fisher neighborhood
around $V^*$ once its cumulative horizon exceeds a fixed threshold
independent of $M,N$, provided the non-saturation condition holds.
\end{lemma}

\begin{proof}
Repeat the ridge-comparator upper-bound argument above with
$u=(1+H_t)^{-1}$ and horizon $t$. The non-saturation condition places $u$
above the lower source endpoint (after adjusting constants). When $u$ is
above the upper source endpoint, the cumulative horizon is bounded and
the bounded population sublevel set gives a uniform constant bound. The
risk--Fisher equivalence on that set completes the estimate. Since
$q>0$, the right-hand side tends to zero uniformly in the non-saturated
range.
\end{proof}

\subsection{Frozen-linear spectral bias}
\label{app:toc-frozen-bias}

Define the deterministic frozen bias
\[
    B_T^{\mathrm{fr}}:=-\Psi_T V^*.
\]

\begin{lemma}[Fisher ridge transfer and frozen bias]
\label{lem:fisher-ridge-frozen-bias}
On $\mathcal G_M\cap\mathcal G_M^{\mathrm{ridge}}$, there are fixed
$D_1,u_1>0$ such that, simultaneously for
$D_1M^{-a}\le u\le u_1$,
\[
    \left\langle V^*,uG(G+uI)^{-1}V^*\right\rangle_F\asymp u^q.
\]
The constants $D_1,u_1$ absorb the fixed Fisher and class-space
comparisons. Consequently, if $M\ge D H_T^{1/a}$ for a sufficiently
large fixed $D$ and $\eta_{\max}\|G\|_{\mathrm{op}}\le1/2$, then
\[
    \boxed{\|B_T^{\mathrm{fr}}\|_G^2\asymp H_T^{-q}.}
\]
\end{lemma}

\begin{proof}
Use the self-adjoint Fisher operator $G$ on $\mathbb R^{r\times M}$ and
its separable comparison $G_0[E]=B_MEW$. The Fisher sandwich gives
$c_FG_0\preceq G\preceq C_FG_0$, and the eigenvalues of the fixed
class matrices $B_M,B_*$ and their inverses are bounded uniformly. For
$f_u(x)=ux/(x+u)$, operator monotonicity gives
$f_u(c_FG_0)\preceq f_u(G)\preceq f_u(C_FG_0)$. Diagonalizing $B_M$
and using $V^*=B_M^{-1}B_*V_M^{\mathrm{LS}}$ shows that the endpoint
quadratic forms reduce to $K_W(u/b)$, up to fixed multiplicative factors,
where $b$ ranges over a compact interval bounded away from zero. Choose
$D_1$ large enough and $u_1$ small enough that every such argument
$u/b$ lies in $[D_0M^{-a},u_0]$ whenever
$D_1M^{-a}\le u\le u_1$. The uniform ridge lemma then yields the claimed
Fisher ridge law.

Let $\nu_G$ be the spectral measure of $G$ induced by $V^*$ and put
$F_G(x)=\int_{(0,x]}\lambda\,d\nu_G(\lambda)$. Applying the same
Tauberian argument with the explicit interval above gives
\[
F_G(x)\lesssim x^q\quad(D_1M^{-a}\le x\le u_1),\qquad
F_G(x)\gtrsim x^q\quad(LD_1M^{-a}\le x\le u_1),
\]
for fixed sufficiently large $L$. Choose the non-saturation constant $D$
so that $M\ge D H_T^{1/a}$ implies
$D_1M^{-a}\le(2H_T)^{-1}$. Then the spectral mass below this lower
cutoff is at most $CD_1^qM^{1-b}\lesssim H_T^{-q}$. On the interval
$[D_1M^{-a},u_1]$, put
$\psi_T(\lambda):=\prod_{t=1}^T(1-\eta_t\lambda)$. By
Lemma~\ref{lem:general-step-filter}, since $0\le\eta_t\lambda\le1/2$,
\[
    e^{-2H_T\lambda}\le\psi_T(\lambda)\le e^{-H_T\lambda}.
\]
Split the integral
$\int\lambda\psi_T(\lambda)^2\,d\nu_G(\lambda)$ into dyadic bands
around $H_T^{-1}$. The band below $H_T^{-1}$ is bounded by
$F_G(H_T^{-1})\lesssim H_T^{-q}$; the band
$(2^j/H_T,2^{j+1}/H_T]$ contributes at most
$Ce^{-c2^j}F_G(2^{j+1}/H_T)\lesssim
 e^{-c2^j}2^{q(j+1)}H_T^{-q}$. The sum is $O(H_T^{-q})$. The
remaining spectrum above $u_1/2$ contributes at most
$e^{-cH_Tu_1}\|V^*\|_G^2=O(H_T^{-q})$.

For the lower bound, take fixed small $\epsilon>0$ and increase $D$ so
that $M\ge D H_T^{1/a}$ implies
$\epsilon/H_T\ge LD_1M^{-a}$; also require
$H_T\ge\epsilon/u_1$. Then
$F_G(\epsilon/H_T)\gtrsim(\epsilon/H_T)^q$. For
$\lambda\le\epsilon/H_T$, the product filter obeys
$\psi_T(\lambda)^2\ge c_\epsilon$.
Thus the frozen-bias quadratic form is at least
$c_\epsilon F_G(\epsilon/H_T)\gtrsim H_T^{-q}$. Finally,
$\|B_T^{\mathrm{fr}}\|_G^2=
\int\lambda\psi_T(\lambda)^2\,d\nu_G(\lambda)$, proving the two-sided
claim with all source endpoints inside the stated admissible interval.
\end{proof}

\begin{lemma}[Frozen bias retains angular source energy]
\label{lem:frozen-angular-bias}
Under Assumption~\ref{ass:decision-nondegeneracy}, let
$B_T^{\mathrm{fr}}:=-\Psi_TV^*$ as in the frozen
linearization.
Define the tangent energy of a parameter perturbation $E$ by
\[
    \mathfrak a(E)^2
    :=
    \sum_{c<d}
    \left\|
        \Pi_{cd}^{\perp}
        \bigl(S^\top E^\top a_{cd}\bigr)
    \right\|_{\Sigma}^{2}.
\]
On $\mathcal G_M\cap\mathcal G_M^{\mathrm{ridge}}$, if
$M\ge D H_T^{1/a}$ and $H_T$ is sufficiently large, then
\[
    \boxed{
    \mathfrak a(B_T^{\mathrm{fr}})^2\asymp H_T^{-q}.
    }
\]
\end{lemma}

\begin{proof}
Write $Wu_j=\mu_j u_j$, $b_t=\Psi_tV^*$, and $b_{t,j}=b_tu_j$.
Differentiating the population Stein identity at $V^*$ gives
\[
    G[E]=B_MEW+DB(V^*)[E]V^*W,
\]
and hence
\[
    b_{t+1,j}
    =(I_r-\eta_{t+1}\mu_jB_M)b_{t,j}
      -\eta_{t+1}\mu_j DB(V^*)[b_t]v_j^*.
\]
The stable-step condition makes every factor in $\Psi_t$ a contraction in
the $G$-norm.
Since $\|V^*\|_G$ is uniformly bounded and $G\asymp I_r\otimes W$,
we have $\sup_t\|b_t\|_W\le C$. Differentiating the finite-dimensional
softmax covariance under the Gaussian expectation gives
\[
    \|DB(V^*)[E]\|_{\mathrm{op}}\le C\|E\|_W.
\]
The matrices $B_M$ are uniformly positive definite. Variation of
constants in the mode recursion retains the derivative coupling
explicitly. With $A_{t,j}:=I_r-\eta_{t+1}\mu_jB_M$ and
$\Phi_j(t,s):=A_{t-1,j}\cdots A_{s,j}$,
\[
 b_{t,j}=\Phi_j(t,0)v_j^*
 -\mu_j\sum_{s=0}^{t-1}\eta_{s+1}\Phi_j(t,s+1)
       DB(V^*)[b_s]v_j^*,
 \qquad
 \|\Phi_j(t,s)\|_{\mathrm{op}}
 \le e^{-c\mu_j(H_t-H_s)}.
\]
The bounds $\sup_s\|b_s\|_W\le C$ and
$\|DB(V^*)[b_s]\|_{\mathrm{op}}\le C\|b_s\|_W$, followed by the
geometric-series estimate, therefore give
\[
    \sup_t\|b_{t,j}\|_2\le C\|v_j^*\|_2.
\]
Moreover, the exact increment identity
\[
 b_{t+1,j}-b_{t,j}
 =-\eta_{t+1}\mu_j\{B_Mb_{t,j}+DB(V^*)[b_t]v_j^*\}
\]
and the preceding uniform mode bound yield, after summing from $0$ to
$T-1$,
\[
    \|b_{T,j}-v_j^*\|_2
    \le CH_T\mu_j\|v_j^*\|_2.
\]
Thus the class-coordinate coupling through $DB(V^*)$ has not been
dropped or replaced by a separable surrogate.
Choose a fixed sufficiently small $\epsilon>0$ and put
$u_T=\epsilon/H_T$. Increase the fixed non-saturation constant $D$ so
that $M\ge D H_T^{1/a}$ implies
$u_T\ge LD_0M^{-a}$; take $H_T\ge\epsilon/u_0$ as well. Thus
$u_T$ is inside the explicit lower-source interval. For $\mu_j\le u_T$
this shows
$\|b_{T,j}\|_2\ge\frac12\|v_j^*\|_2$. The low-frequency Tauberian
bound in Lemma~\ref{lem:uniform-ridge-source-filter}, transferred from
$V_M^{\mathrm{LS}}$ to $V^*=B_M^{-1}B_*V_M^{\mathrm{LS}}$, gives
\[
    \|B_T^{\mathrm{fr}}\mathbf1_{(0,u_T]}(W)\|_W^2
    \gtrsim H_T^{-q},
    \qquad
    \|B_T^{\mathrm{fr}}\|_W^2\lesssim H_T^{-q}.
\]

It remains to show that pairwise radial projection does not remove this
low-frequency energy. Whiten the input space and write
\[
    B=S\Sigma^{1/2},\qquad Q=A_*\Sigma^{1/2},\qquad
    h_{cd}=Q^\top a_{cd},\qquad
    e_{cd}=B^\top(B_T^{\mathrm{fr}})^\top a_{cd}.
\]
Let $\mathsf P_u$ project onto
$\operatorname{span}\{B^\top u_j:\mu_j\le u\}$. The vectors
$B^\top u_j$ are mutually orthogonal and have squared norm $\mu_j$;
thus
\[
    \mathsf P_u B^\top E^\top a_{cd}
    =B^\top\mathbf1_{(0,u]}(W)E^\top a_{cd}.
\]
The contrast-frame identity and Gaussian conditional regression imply
\[
    \sum_{c<d}\|\mathsf P_{u_T}e_{cd}\|_2^2
    =C\|B_T^{\mathrm{fr}}\mathbf1_{(0,u_T]}(W)\|_W^2
    \gtrsim H_T^{-q},
\]
while
\[
    \sum_{c<d}\|\mathsf P_{u_T}h_{cd}\|_2^2
    =C\|V_M^{\mathrm{LS}}\mathbf1_{(0,u_T]}(W)\|_W^2
    \lesssim H_T^{-q}.
\]
For each pair, set
\[
    \alpha_{cd}:=\frac{\langle e_{cd},h_{cd}\rangle}{\|h_{cd}\|_2^2}.
\]
Assumption~\ref{ass:decision-nondegeneracy} bounds the denominators below,
and the frozen-bias upper bound gives
$\max_{c<d}|\alpha_{cd}|\lesssim H_T^{-q/2}$. Therefore, in the direct
sum of pair spaces,
\[
\begin{aligned}
\left(\sum_{c<d}
  \|\mathsf P_{u_T}\Pi_{cd}^{\perp}e_{cd}\|_2^2\right)^{1/2}
&\ge
\left(\sum_{c<d}\|\mathsf P_{u_T}e_{cd}\|_2^2\right)^{1/2}
 -\max_{c<d}|\alpha_{cd}|
  \left(\sum_{c<d}\|\mathsf P_{u_T}h_{cd}\|_2^2\right)^{1/2}\\
&\ge cH_T^{-q/2}-CH_T^{-q}
\ge c'H_T^{-q/2}
\end{aligned}
\]
for all sufficiently large $H_T$. Since orthogonal projection is a
contraction, this proves $\mathfrak a(B_T^{\mathrm{fr}})^2\gtrsim
H_T^{-q}$. The upper bound follows from
$\mathfrak a(E)^2\le C\|E\|_W^2$ and the frozen-bias upper bound.
\end{proof}

\section{Empirical Full-Batch GD: Score Fluctuation and Nonlinear Response}
\label{app:fullbatch-empirical}

Fix a sketch in $\mathcal G_M\cap\mathcal G_M^{\mathrm{ridge}}$. Let
$W=\Sigma_S$, $V^*=V_M^*$, $G=\nabla^2\mathcal R_S(V^*)$,
$k_T=\min\{M,H_T^{1/a}\}$, and
\[
    Q_{T,N}:=H_T^{-q}+\frac{k_T}{N}.
\]
For the all-horizon argument, also put
\[
    A_M:=M^{1-b},\qquad
    \rho_{M,t}:=(1+H_t)^{-q}+A_M,
    \qquad
    \overline Q_{M,N,T}:=A_M+H_T^{-q}+\frac{k_T}{N}.
\]
All expectations in this appendix are conditional on the sketch.
For $0\le s\le t\le T$, define
\[
    \Phi(t,s):=\prod_{u=s+1}^{t}(I-\eta_uG),
    \qquad \Phi(s,s):=I,
    \qquad H_t:=\sum_{u=1}^t\eta_u.
\]
All factors commute. The finite-sum identities in the next subsection also
hold for any fixed sketch and deterministic schedule whenever $V^*$ exists;
they do not require the sketch event or an invertible $G$.

\subsection{Exact decomposition}
\label{app:toc-empirical-decomposition}

Define the fixed empirical score and the two remainders by
\[
\begin{aligned}
    \zeta_N&:=\nabla\widehat R_N(V^*),\\
    \mathcal N(e)&:=\nabla\mathcal R_S(V^*+e)-Ge,\\
    \mathcal E_N(e)&:=\nabla\widehat R_N(V^*+e)-\nabla\mathcal R_S(V^*+e)-\zeta_N.
\end{aligned}
\]
Since $\nabla\mathcal R_S(V^*)=0$, empirical GD obeys the exact recursion
\[
    e_{t+1}=(I-\eta_{t+1}G)e_t-\eta_{t+1}\zeta_N
    -\eta_{t+1}\mathcal N(e_t)-\eta_{t+1}\mathcal E_N(e_t),
    \qquad e_t:=V_t-V^*.
\]
Iteration from $e_0=-V^*$ gives
\[
\boxed{e_T=B_T^{\mathrm{fr}}+Z_T+D_T,}
\]
where
\[
\begin{aligned}
    B_T^{\mathrm{fr}}&=-\Phi(T,0)V^*,\\
    Z_T&=-\sum_{s=0}^{T-1}\eta_{s+1}\Phi(T,s+1)\zeta_N,\\
    D_T&=-\sum_{s=0}^{T-1}\eta_{s+1}\Phi(T,s+1)
       [\mathcal N(e_s)+\mathcal E_N(e_s)],\\
    f_T(\lambda)&:=1-\prod_{t=1}^T(1-\eta_t\lambda).
\end{aligned}
\]
This identity holds for any deterministic step sequence and treats data reuse
exactly; $\zeta_N$ is a single fixed empirical score, not fresh noise at
each iteration. When $G$ is invertible, the filtered score can equivalently
be written as $Z_T=-f_T(G)G^{-1}\zeta_N$.

\subsection{Uniform empirical localization}
\label{app:toc-localization}

Put $g_i=W^{-1/2}z_i$ and $A=VW^{1/2}$. For $\|A\|_F$
bounded, write
\[
\widehat H_N(A)=\frac1N\sum_{i=1}^N
 \widetilde C(Ag_i)\otimes g_ig_i^\top,
\qquad
H(A)=\mathbb E_g[\widetilde C(Ag)\otimes gg^\top].
\]
The Kronecker product denotes the Hessian operator on the corresponding
vectorized matrix space; equivalently, its quadratic form at a matrix $D$
is $N^{-1}\sum_i(Dg_i)^\top\widetilde C(Ag_i)(Dg_i)$.

\begin{lemma}[Uniform empirical Hessian and risk event]
\label{lem:uniform-empirical-hessian}
Fix $R_0,K<\infty$. If $M\log(N+T)=o(N)$, then, conditional on any sketch
with bounded $\operatorname{tr}(W)$ and spectrum as in
Lemma~\ref{lem:sketch-spectrum}, with probability at least
$1-(N+T)^{-K}$,
\[
\sup_{\|A\|_F\le R_0+1}
\|\widehat H_N(A)-H(A)\|_{\mathrm{op}}
\le \epsilon_N,
\qquad
\epsilon_N=C_K\sqrt{\frac{M\log(N+T)}{N}}.
\]
On the same event, for a sufficiently large fixed $R_0$, empirical risk
has a strict barrier above $\widehat R_N(0)$ on
$\|VW^{1/2}\|_F=R_0$, and all full-batch GD iterates satisfy
\[
    \sup_{0\le t\le T}\|V_tW^{1/2}\|_F\le R_0.
\]
\end{lemma}

\begin{proof}
Let $g_i=W^{-1/2}z_i$. For fixed $A$ in the stated ball and fixed unit
$D\in\mathbb R^{r\times M}$, the Hessian quadratic-form summand
\[
X_i(A,D):=(Dg_i)^\top\widetilde C(Ag_i)(Dg_i)
\]
satisfies $0\le X_i(A,D)\le C\|Dg_i\|_2^2$ and
$\|X_i(A,D)\|_{\psi_1}\le C$, uniformly in $M,A,D$. Bernstein's
inequality gives deviation at most
$C(\sqrt{x/N}+x/N)$ with failure at most $2e^{-x}$.

Set $L_N=C_K(\sqrt M+\sqrt{\log(N+T)})$. Gaussian chi-square tails
and a union bound give
$\max_i\|g_i\|_2\le L_N$ except with probability at most
$(N+T)^{-K-3}$. On this event, for any $A,A'$ and unit $D,D'$, the
quadratic forms and their expectations obey
\[
|\widehat q(A,D)-\widehat q(A',D')|
\le C L_N^3\|A-A'\|_F+C L_N^2\|D-D'\|_F,
\]
and the same bound holds for the population forms. Choose Frobenius nets
of mesh $\delta_A=\epsilon_N/(CL_N^3)$ for the $A$-ball and
$\delta_D=\epsilon_N/(CL_N^2)$ for the unit $D$-sphere. Their combined
log-cardinality is $O(M\log(N+T))$, since $r$ is fixed, $M\le N$
eventually, and $L_N,\epsilon_N^{-1}$ are at most polynomial in $N+T$.
Apply Bernstein to the product net with
$x=C_KM\log(N+T)$. The linear Bernstein term is absorbed by the square-root
term under $M\log(N+T)=o(N)$. The net extension above proves the displayed
uniform operator bound; the union-bound and truncation failure
probabilities are at most $(N+T)^{-K-1}$ after increasing constants.

For the risk event, the centered loss increment
$\ell(A;g,y)-\ell(0;g,y)$ is bounded in absolute value by
$C\|Ag\|_2$, and has a uniform sub-exponential norm on the parameter
ball because $Ag$ is an $r$-dimensional Gaussian with covariance trace
at most $(R_0+1)^2$.
A net of mesh $\epsilon_N/(CL_N)$ in that ball has log-cardinality
$O(M\log(N+T))$. Bernstein on this net, followed by the Lipschitz
extension $C\|A-A'\|_F N^{-1}\sum_i\|g_i\|_2$, yields
\[
\sup_{\|A\|_F\le R_0+1}
\left|[\widehat R_N(A)-\widehat R_N(0)]-
      [\mathcal R_S(A)-\mathcal R_S(0)]\right|
\le C_K\sqrt{\frac{M\log(N+T)}N}=o(1)
\]
with the same polynomial confidence. A net argument for the gradient at
zero similarly gives
$\|\nabla_A\widehat R_N(0)-\nabla_A\mathcal R_S(0)\|_F
\le C_K\sqrt{(M+\log(N+T))/N}$. Gaussian Stein regression and the
bounded teacher signal give $\|\nabla_A\mathcal R_S(0)\|_F\le C$.
Also, Gaussian sample-covariance concentration gives
$\|N^{-1}\sum_i g_ig_i^\top\|_{\mathrm{op}}\le C$ with the same
confidence. Since softmax curvature is globally bounded, this is a global
smoothness bound for $\widehat R_N$ in $A$ coordinates. It follows that
$\|\nabla_A\widehat R_N(A)\|_F\le C_{R_0}$ on the ball
$\|A\|_F\le R_0+1$.

Population coercivity supplies a fixed positive gap above
$\mathcal R_S(0)$ on a sufficiently large sphere $\|A\|_F=R_0$;
the uniform risk bound transfers this to a strict empirical barrier above
$\widehat R_N(0)$. In $A=VW^{1/2}$ coordinates the actual update is the
preconditioned step
$A_{t+1}=A_t-\eta_{t+1}\nabla_A\widehat R_N(A_t)W$. The empirical smoothness
constant in these coordinates is bounded by a fixed $L$, and
$\|W\|_{\mathrm{op}}\le\operatorname{tr}(W)\le C$. Thus, if
$\eta_{\max} L\|W\|_{\mathrm{op}}\le1$, the descent lemma gives nonincreasing
risk along the entire update segment; indeed
\[
\langle\nabla_A\widehat R_N,\nabla_A\widehat R_NW\rangle
\ge0,\qquad
\|\nabla_A\widehat R_NW\|_F^2
\le\|W\|_{\mathrm{op}}
\langle\nabla_A\widehat R_N,\nabla_A\widehat R_NW\rangle.
\]
The gradient bound on the enlarged ball also gives segment length at most
$\eta_{\max} C_{R_0}\|W\|_{\mathrm{op}}\le1$ after reducing the fixed step
size. If an iterate first crossed the sphere, its segment up to the first
intersection would lie in the enlarged ball and have risk no greater than
$\widehat R_N(0)$, contradicting the strict barrier. Hence all iterates
remain in the sphere. In the whitened $A$ coordinates, the population
Hessian and $G$ are both uniformly comparable to the identity throughout
this ball. Therefore the operator deviation above gives, after decreasing
the fixed constants when $\epsilon_N$ is small,
\[
    cG\preceq \nabla^2\widehat R_N(V)\preceq CG
    \qquad\text{for every }\|VW^{1/2}\|_F\le R_0.
\]
The ball was chosen to contain the origin, $V^*$, every ridge comparator
used below, and their joining segments. Its containment of the empirical
iterates is a consequence of the risk barrier and descent argument above,
not an additional localization assumption.
\end{proof}

The covering argument also gives, for the score
$s_N:=\|\zeta_N\|_{G^{-1}}$,
\begin{equation}
\boxed{
\mathbb E[s_N^2\mid S]\asymp\frac{rM}{N},\qquad
\mathbb E[s_N^{2p}\mid S]\lesssim\left(\frac{M}{N}\right)^p,
\quad p=2,4.
}
\label{eq:fullbatch-score-moments}
\end{equation}
Indeed $\mathbb E[\zeta_N\mid S]=0$ and
$\operatorname{Cov}(\operatorname{vec}\zeta_N\mid S)=\Gamma/N$ with
$cG\preceq\Gamma\preceq CG$. After whitening by $G^{-1/2}$, a single
score has moments of order $M^p$ because the softmax residual is bounded
and $g_i$ is standard Gaussian. Hilbert-space Rosenthal inequalities
then give \eqref{eq:fullbatch-score-moments}.

\subsection{An actual empirical-GD trajectory bound}
\label{app:toc-trajectory}

\begin{lemma}[Empirical GD path bound]
\label{lem:fullbatch-trajectory-bound}
On the event of Lemma~\ref{lem:uniform-empirical-hessian}, if
$\epsilon_N$ is sufficiently small, then for every $0\le t\le T$,
\[
    \boxed{\|e_t\|_G^2\lesssim\rho_{M,t}+s_N^2.}
\]
\end{lemma}

\begin{proof}
The uniform Hessian deviation integrated along the segment from $V^*$ to
$V^*+e$ gives, throughout the localized ball,
\[
\widehat R_N(V^*+e)-\widehat R_N(V^*)
=\mathcal R_S(V^*+e)-\mathcal R_S(V^*)
 +\langle\zeta_N,e\rangle_F+\rho_N(e),
\qquad |\rho_N(e)|\le C\epsilon_N\|e\|_G^2.
\]
Population curvature and Fisher equivalence therefore imply the lower
bound
$c\|e\|_G^2-s_N\|e\|_G-C\epsilon_N\|e\|_G^2$ for this empirical
excess.

Let $D_0$ be the lower endpoint in
Lemma~\ref{lem:uniform-ridge-source-filter} and define
\[
    u_t:=\max\{(1+H_t)^{-1},D_0M^{-a}\}.
\]
For times with $u_t\le u_0$, the ridge comparator
$V_{u_t}=QB^\top(W+u_tI)^{-1}$ lies in the admissible source range for
every relation between $M$ and $t$.
The source estimates, oracle compatibility, and Fisher equivalence give
\[
\|V_{u_t}-V^*\|_G^2\le Cu_t^q,
\qquad \|V_{u_t}\|_F^2\le Cu_t^{q-1}.
\]
These comparators lie in a fixed prediction-norm ball. Choose the fixed
localization radius $R_0$ large enough to contain that ball, $V^*$, and
the origin; then their joining segments with $V^*$ lie in the empirical
localization event.

On the empirical covariance/Hessian event, the global softmax-curvature
bound makes $\widehat R_N$ globally $L$-smooth in Frobenius norm, with
$L\le C\|N^{-1}\sum_i z_iz_i^\top\|_{\mathrm{op}}$.
The stable learning-rate condition in Section~\ref{subsec:optimization}
ensures $\eta_s\le L^{-1}$ for every GD step.
Lemma~\ref{lem:gd-comparator}, applied to $\widehat R_N$ over the first
$t$ steps with comparator $V_{u_t}$, therefore gives directly
\[
\widehat R_N(V_t)-\widehat R_N(V_{u_t})
\le\frac{\|V_{u_t}\|_F^2}{2H_t}.
\]
The same empirical objective is used at every step. The optimization
term is $O(u_t^q)$ in both branches of the cap. Indeed, if
$u_t=(1+H_t)^{-1}$, then
\[
 \frac{u_t^{q-1}}{H_t}
 =u_t^q\frac{1+H_t}{H_t}\le C u_t^q
\]
outside the already-separated bounded cumulative-time interval. If
$u_t=D_0M^{-a}$, cap activity implies
$(1+H_t)^{-1}\le u_t$, hence $H_t^{-1}\le C u_t$ and again
$u_t^{q-1}/H_t\le Cu_t^q$. Thus the capped comparator remains valid
after the spectral cutoff rather than being extrapolated below the ridge
window. Applying the integrated Hessian identity
to the comparator also gives
\[
\widehat R_N(V_{u_t})-\widehat R_N(V^*)
\le C(u_t^q+s_Nu_t^{q/2}+\epsilon_Nu_t^q).
\]
Combine these upper bounds with the preceding lower bound at $e=e_t$.
For sufficiently small $\epsilon_N$, absorb its quadratic term and use
$2s_N\|e_t\|_G\le c\|e_t\|_G^2/2+Cs_N^2$ and
$s_Nu_t^{q/2}\le (u_t^q+s_N^2)/2$. This yields
$\|e_t\|_G^2\le C(u_t^q+s_N^2)$. Since
\[
    u_t^q
    \asymp\max\{(1+H_t)^{-q},M^{1-b}\}
    \asymp\rho_{M,t},
\]
this is the asserted all-horizon rate.

If $u_t>u_0$, then $H_t$ belongs to a fixed bounded initial interval.
Localization bounds $\|e_t\|_G$ there, while $(1+H_t)^{-q}$ is bounded
below by a positive fixed constant.
The case $t=0$ follows from the bounded prediction norm of $V^*$.
Combining the ranges proves the claim for every $0\le t\le T$.
\end{proof}

\subsection{Frozen score variance and nonlinear response}
\label{app:toc-score-response}

\begin{lemma}[Frozen empirical-score variance]
\label{lem:fullbatch-frozen-variance}
Let $Z_T=-f_T(G)G^{-1}\zeta_N$. On the sketch event,
\[
\boxed{
\mathbb E\|Z_T\|_G^2\asymp\frac{k_T}{N},\qquad
\mathbb E\|Z_T\|_G^4\lesssim\left(\frac{k_T}{N}\right)^2,
\qquad
G^{1/2}\operatorname{Cov}(\operatorname{vec}Z_T\mid S)G^{1/2}
\preceq\frac CN I.
}
\]
\end{lemma}

\begin{proof}
Let $\Gamma$ be the covariance of one score at $V^*$, so
$\operatorname{Cov}(\zeta_N\mid S)=\Gamma/N$. The covariance identity and
Lemma~\ref{lem:quasi-fisher} give
\[
\mathbb E\|Z_T\|_G^2
=\frac1N\operatorname{tr}
 [G f_T(G)G^{-1}\Gamma G^{-1}f_T(G)]
\asymp\frac1N\operatorname{tr}[f_T(G)^2].
\]
Lemma~\ref{lem:sketch-effective-dimension} makes this comparable to
$k_T/N$. Whitening the covariance further gives
\[
G^{1/2}\operatorname{Cov}(Z_T\mid S)G^{1/2}
=\frac1Nf_T(G)G^{-1/2}\Gamma G^{-1/2}f_T(G)
\preceq C N^{-1}I.
\]

For the fourth moment, write $Z_T=N^{-1}\sum_{i=1}^N X_i$ as a sum of
independent centered Hilbert-space score vectors after applying the fixed
filter $f_T(G)G^{-1}$. Quasi-Fisher equivalence gives
$\mathbb E\|X_i\|_G^2\lesssim k_T$. To bound a single fourth moment,
write the per-example score as $\delta_i\otimes W^{1/2}g_i$, where
$\|\delta_i\|_2\le C$ and $g_i\sim N(0,I_M)$. For each of the fixed
$r$ class coordinates the filtered feature is a deterministic linear
image $L_cg_i$. Their total Hilbert--Schmidt energy satisfies
$\sum_{c=1}^r\|L_c\|_F^2\lesssim\operatorname{tr}f_T(G)^2\lesssim k_T$,
by the Fisher sandwich. Since $\delta_i$ is bounded and Gaussian fourth
moments obey $\mathbb E\|L_cg_i\|_2^4\le3\|L_c\|_F^4$, we get
$\mathbb E\|X_i\|_G^4\lesssim k_T^2$. The Hilbert-space fourth-moment
inequality
\[
\mathbb E\Big\|\sum_{i=1}^N X_i\Big\|_G^4
\le C\left[\left(\sum_i\mathbb E\|X_i\|_G^2\right)^2
                 +\sum_i\mathbb E\|X_i\|_G^4\right]
\]
therefore yields $\mathbb E\|Z_T\|_G^4\lesssim(k_T/N)^2$.
\end{proof}

\begin{lemma}[Nonlinear response bound]
\label{lem:fullbatch-nonlinear-response}
Assume the event of Lemma~\ref{lem:uniform-empirical-hessian} and the
non-saturated range $M\ge DH_T^{1/a}$. Then pathwise on that event,
\[
\boxed{
\|D_T\|_G\lesssim\log(2+H_T)
\left[H_T^{-q}+s_N^2+
\epsilon_N(H_T^{-q/2}+s_N)\right].
}
\]
If additionally
\[
    \frac{M^2\log^3(N+T)}{Nk_T}\longrightarrow0,
\]
then, for $p=2,4$,
\[
    \|D_T\mathbf1_{\mathcal B}\|_{L^p(G)}
    =o(\sqrt{Q_{T,N}}),
\]
where $\mathcal B$ is the good event of
Lemma~\ref{lem:uniform-empirical-hessian}. The complement is handled
separately in Theorem~\ref{thm:fullbatch-estimation-risk}.
\end{lemma}

\begin{proof}
The global softmax remainder lemma gives
$\|\mathcal N(e)\|_{G^{-1}}\lesssim\|e\|_G^2$. On the empirical event,
$\|\mathcal E_N(e)\|_{G^{-1}}\le\epsilon_N\|e\|_G$. Spectral calculus,
$\eta_{\max}\|G\|_{\mathrm{op}}\le1/2$, and
Lemma~\ref{lem:general-step-filter} give
\[
    \|\Phi(T,s+1)f\|_G
    \le\frac{C}{1+H_T-H_{s+1}}\|f\|_{G^{-1}}.
\]
Therefore
\[
\|D_T\|_G\lesssim\sum_{s=0}^{T-1}\eta_{s+1}
\frac{\|e_s\|_G^2+\epsilon_N\|e_s\|_G}
     {1+H_T-H_{s+1}}.
\]
For $0\le\alpha<1$, Lemma~\ref{lem:general-step-filter} gives
\[
\sum_{s=0}^{T-1}\eta_{s+1}
\frac{(1+H_s)^{-\alpha}}{1+H_T-H_{s+1}}
\lesssim (1+H_T)^{-\alpha}\log(2+H_T).
\]
Substitution of Lemma~\ref{lem:fullbatch-trajectory-bound} yields the
pathwise assertion. Taking $L^p$ norms and using
\eqref{eq:fullbatch-score-moments} gives
\[
\|D_T\mathbf1_{\mathcal B}\|_{L^p(G)}\lesssim\log(2+H_T)
\left[H_T^{-q}+\frac MN+
\epsilon_N\left(H_T^{-q/2}+\sqrt{\frac MN}\right)\right].
\]
Since $Q_{T,N}\ge k_T/N$, division by $\sqrt{Q_{T,N}}$ bounds the ratio
by
\[
C\log(2+H_T)\left[
H_T^{-q/2}+\frac{M}{\sqrt{Nk_T}}
+\epsilon_N\sqrt{\frac{M}{k_T}}\right]=o(1)
\]
under the displayed sample-size condition. The score eighth-moment bound
handles $p=4$.
\end{proof}

\begin{lemma}[Post-cutoff response when sampling fluctuation is visible]
\label{lem:post-cutoff-response}
Suppose
\[
    H_T\ge c_{\mathrm{sat}}M^a,
    \qquad
    \frac{M}{N}\gtrsim M^{1-b},
    \qquad
    \frac{M\log^3(N+T)}{N}\longrightarrow0,
\]
where $c_{\mathrm{sat}}$ is as in
Lemma~\ref{lem:post-cutoff-bias}.  On the event $\mathcal B$ of
Lemma~\ref{lem:uniform-empirical-hessian}, for $p=2,4$,
\[
    \boxed{
    \|D_T\mathbf1_{\mathcal B}\|_{L^p(G)}
    =o\!\left(\sqrt{\frac MN}\right).
    }
\]
\end{lemma}

\begin{proof}
The convolution argument in
Lemma~\ref{lem:fullbatch-nonlinear-response}, now using the all-horizon
path bound of Lemma~\ref{lem:fullbatch-trajectory-bound}, gives
\[
\|D_T\|_G\lesssim\log(2+H_T)
\left[
\rho_{M,T}+s_N^2+
\epsilon_N(\rho_{M,T}^{1/2}+s_N)
\right].
\]
Indeed, the extra constant summand $A_M$ in $\rho_{M,t}$ contributes at
most $CA_M\log(2+H_T)$ to the same resolvent convolution.  In the
post-cutoff range, $H_T^{-q}\lesssim A_M$, so
$\rho_{M,T}\lesssim A_M$.  Taking $L^p$ norms and using the score
moments therefore yields
\[
\|D_T\mathbf1_{\mathcal B}\|_{L^p(G)}
\lesssim\log(N+T)
\left[
A_M+\frac MN+
\epsilon_N\left(\sqrt{A_M}+\sqrt{\frac MN}\right)
\right].
\]
Put $v_M=M/N$. For both $p=2$ and $p=4$, the score bounds
\eqref{eq:fullbatch-score-moments} give
$\|s_N^2\|_{L^p}\lesssim v_M$ and
$\|s_N\|_{L^p}\lesssim\sqrt{v_M}$; the former uses no more than the
displayed eighth moment. Since $A_M\lesssim v_M$ and
$\epsilon_N\lesssim\sqrt{v_M\log(N+T)}$, division term by term by
$\sqrt{v_M}$ gives
\[
\begin{aligned}
&C\left[
\frac{A_M}{\sqrt{v_M}}\log(N+T)
+\sqrt{v_M}\log(N+T)
+\epsilon_N\sqrt{\frac{A_M}{v_M}}\log(N+T)
+\epsilon_N\log(N+T)
\right]\\
&\qquad\le C\left[
\sqrt{v_M}\log(N+T)
+\sqrt{v_M\log^3(N+T)}
\right]=o(1).
\end{aligned}
\]
This proves the claim for both stated moments. The same empirical score
generated by the training sample is used throughout; no fresh or
independent score is introduced along the trajectory.
\end{proof}

\subsection{Unconditional estimation-risk consequence}
\label{app:toc-estimation-risk}

\begin{theorem}[Finite-iteration empirical GD risk]
\label{thm:fullbatch-estimation-risk}
Suppose $H_T\to\infty$, $M\ge DH_T^{1/a}$, and
$M^2\log^3(N+T)/(Nk_T)\to0$. Conditional on
$S\in\mathcal G_M\cap\mathcal G_M^{\mathrm{ridge}}$,
\[
\boxed{
\mathbb E[\mathcal R_S(V_T)-\mathcal R_S(V^*)\mid S]
\asymp H_T^{-q}+\frac{k_T}{N}.
}
\]
The same scale holds for $\mathbb E\|V_T-V^*\|_G^2$. The frozen term
$Z_T$ has variance $\asymp k_T/N$, but this theorem does not identify it
with the exact centered-iterate variance.
\end{theorem}

\begin{proof}
Let $\mathcal B$ denote the good event from
Lemma~\ref{lem:uniform-empirical-hessian}, and put
$Y_T=B_T^{\mathrm{fr}}+Z_T$. Conditional on the sketch,
$B_T^{\mathrm{fr}}$ is deterministic and $\mathbb E[Z_T\mid S]=0$.
Consequently the cross term vanishes exactly,
\[
 \mathbb E\|Y_T\|_G^2
 =\|B_T^{\mathrm{fr}}\|_G^2+\mathbb E\|Z_T\|_G^2,
\]
and the frozen-filter and variance lemmas, together with
$(x+y)^4\le8(x^4+y^4)$, give
\[
\mathbb E\|Y_T\|_G^2\asymp Q_{T,N},\qquad
\mathbb E\|Y_T\|_G^4\lesssim Q_{T,N}^2.
\]
The restricted response estimate gives
$\|D_T\mathbf1_{\mathcal B}\|_{L^2(G)}=o(\sqrt{Q_{T,N}})$ and
$\|D_T\mathbf1_{\mathcal B}\|_{L^4(G)}=o(\sqrt{Q_{T,N}})$. Since
$\Pr(\mathcal B^c)\le(N+T)^{-K}$, Cauchy--Schwarz gives
\[
 \mathbb E[\|Y_T\|_G^2\mathbf1_{\mathcal B^c}]
 \le(\mathbb E\|Y_T\|_G^4)^{1/2}\Pr(\mathcal B^c)^{1/2}
 =o(Q_{T,N}).
\]
Thus $\|Y_T\mathbf1_{\mathcal B}\|_{L^2(G)}\asymp\sqrt{Q_{T,N}}$.
The exact decomposition $e_T=Y_T+D_T$ and the reverse triangle
inequality in $L^2(\mathcal B;G)$ give the matching lower bound; the
ordinary triangle inequality gives the upper and fourth-moment bounds.
No independence between $D_T$ and $Z_T$, and no cancellation of their
cross term, is used.

To control the complement without a circular appeal to a moment bound
for $e_T$, use the pointwise score estimate, valid at every iterate,
\[
\|V_T\|_F\le CH_T\frac1N\sum_{i=1}^N\|z_i\|_2.
\]
On the sketch event $\operatorname{tr}(W)\le C$, every fixed moment of
the sample average is bounded uniformly in $M,N$. Also
$\|e_T\|_G\le C(1+\|V_T\|_F)$ by the Fisher upper bound and
$\|W\|_{\mathrm{op}}\le C$. Thus, for $p=2,4$,
\[
\mathbb E[\|e_T\|_G^p\mathbf1_{\mathcal B^c}]
\le C(1+T^p)(N+T)^{-K/2}.
\]
Likewise, $\|Z_T\|_G\le s_N$ and the eighth score moment give, for
$p=2,4$,
\[
\mathbb E[\|Y_T\|_G^p\mathbf1_{\mathcal B^c}]
\le C(H_T^{-pq/2}+(M/N)^{p/2})(N+T)^{-K/2}.
\]
The sample-size condition implies $M\le N$ eventually, while
$Q_{T,N}\ge cM^{-(b-1)}\ge c(N+T)^{-(b-1)}$ in this regime.
Since $H_T\le\eta_0T$, choosing $K>4(b+1)+2$ makes the complement contributions
$o(Q_{T,N})$ for $p=2$ and $o(Q_{T,N}^2)$ for $p=4$. The identity
$D_T=e_T-Y_T$ gives the same complement bounds for $D_T$. We have proved
unconditionally
\[
\mathbb E\|e_T\|_G^2\asymp Q_{T,N},\qquad
\mathbb E\|e_T\|_G^4\lesssim Q_{T,N}^2,
\]
and $\mathbb E\|D_T\|_G^2=o(Q_{T,N})$,
$\mathbb E\|D_T\|_G^4=o(Q_{T,N}^2)$.

On $\mathcal B$, population risk--Fisher equivalence applies to $e_T$.
On $\mathcal B^c$, nonnegativity of excess risk and the global softmax
loss bound give
$\mathcal R_S(V_T)-\mathcal R_S(V^*)\le C(1+\|V_T\|_W)$; the same
iterate moment estimate makes its expectation $o(Q_{T,N})$. Hence the
excess-risk expectation has the same order as $Q_{T,N}$.
\end{proof}

\begin{theorem}[All-horizon empirical GD total scale]
\label{thm:fullbatch-total-risk}
Suppose $H_T\to\infty$ and
\[
    \frac{M^2\log^3(N+T)}{Nk_T}\longrightarrow0.
\]
Conditional on
$S\in\mathcal G_M\cap\mathcal G_M^{\mathrm{ridge}}$, one has
\[
\boxed{
A_M+
\mathbb E[\mathcal R_S(V_T)-\mathcal R_S(V^*)\mid S]
\asymp
\overline Q_{M,N,T}
\asymp
k_T^{1-b}+\frac{k_T}{N}.
}
\]
Moreover,
\[
\mathbb E[\mathcal R_S(V_T)-\mathcal R_S(V^*)\mid S]
\lesssim\overline Q_{M,N,T},
\qquad
\mathbb E\|V_T-V^*\|_G^2
\lesssim\overline Q_{M,N,T}.
\]
\[
\mathbb E\|V_T-V^*\|_G^4
\lesssim\overline Q_{M,N,T}^2.
\]
In the post-cutoff regime $H_T\ge c_{\mathrm{sat}}M^a$, if in
addition $M/N\gtrsim M^{1-b}$, then the parameter excess itself obeys
\[
\mathbb E[\mathcal R_S(V_T)-\mathcal R_S(V^*)\mid S]
\asymp\frac MN,
\qquad
\mathbb E\|V_T-V^*\|_G^2\asymp\frac MN.
\]
No two-sided claim is made for the post-cutoff parameter bias when the
algorithm-independent approximation scale dominates.
\end{theorem}

\begin{proof}
First note the deterministic identity
\[
A_M+H_T^{-q}\asymp
\max\{M^{1-b},H_T^{-q}\}
=\min\{M,H_T^{1/a}\}^{1-b}=k_T^{1-b}.
\tag{D.2}\label{eq:all-horizon-algebra}
\]

Reduce the fixed $c_{\mathrm{sat}}$ from
Lemma~\ref{lem:post-cutoff-bias}, if necessary, so that
$H_T<c_{\mathrm{sat}}M^a$ ensures
$M\ge DH_T^{1/a}$ for the fixed $D$ in
Theorem~\ref{thm:fullbatch-estimation-risk}.  That theorem gives
\[
\mathbb E[\mathcal R_S(V_T)-\mathcal R_S(V^*)\mid S]
\asymp H_T^{-q}+\frac{k_T}{N}
\]
and the corresponding fourth-moment upper bound.  Since
$A_M\lesssim H_T^{-q}$ in this range, all assertions follow.

It remains to consider $H_T\ge c_{\mathrm{sat}}M^a$.  Here
\[
    k_T\asymp M,\qquad H_T^{-q}\lesssim A_M,\qquad
    \overline Q_{M,N,T}\asymp A_M+v_M,\quad v_M:=M/N.
\]
On the event $\mathcal B$ of
Lemma~\ref{lem:uniform-empirical-hessian}, the all-horizon path bound
gives
\[
    \|e_T\|_G^2\lesssim A_M+s_N^2.
\]
The score moments therefore imply, for $p=2,4$,
\[
    \mathbb E[\|e_T\|_G^p\mathbf1_{\mathcal B}]
    \lesssim(A_M+v_M)^{p/2}.
\]
The complement is lower order.  Indeed, the pointwise iterate estimate
used in Theorem~\ref{thm:fullbatch-estimation-risk} gives
\[
\mathbb E[\|e_T\|_G^p\mathbf1_{\mathcal B^c}]
\le C(1+T^p)(N+T)^{-K/2}.
\]
The sample condition implies $M=o(N)$ and hence $M\le N+T$ eventually,
while $A_M=M^{1-b}\ge(N+T)^{1-b}$. Since $1+T^p\le C(N+T)^p$, it is
enough to choose the arbitrary confidence exponent so that
$K>p(b+1)$; choosing once and for all $K>4(b+1)+2$ works simultaneously
for $p=2,4$. With this choice the last display is
$o((A_M+v_M)^{p/2})$.  Population risk is uniformly equivalent to the
$G$-norm on $\mathcal B$; on $\mathcal B^c$, the global softmax-loss
bound and the same iterate estimate make the expected contribution
$o(A_M+v_M)$.  This proves the parameter-risk upper bound and the fourth
moment assertion in the post-cutoff range.

The approximation summand $A_M$ already supplies the lower bound when
$v_M\lesssim A_M$.  Suppose therefore that $v_M\gtrsim A_M$.  Put
$Y_T=B_T^{\mathrm{fr}}+Z_T$.  Lemmas
\ref{lem:post-cutoff-bias} and~\ref{lem:fullbatch-frozen-variance} give
\[
    \mathbb E\|Y_T\|_G^2
    =\|B_T^{\mathrm{fr}}\|_G^2+
      \mathbb E\|Z_T\|_G^2
    \gtrsim v_M,
    \qquad
    \mathbb E\|Y_T\|_G^4\lesssim v_M^2.
\]
Here the second-moment identity is exact because
$B_T^{\mathrm{fr}}$ is deterministic conditional on $S$ and
$\mathbb E[Z_T\mid S]=0$; the fourth-moment bound follows from
$(x+y)^4\le8(x^4+y^4)$. Hence Cauchy--Schwarz and the chosen confidence
exponent give
\[
 \mathbb E[\|Y_T\|_G^2\mathbf1_{\mathcal B^c}]
 \le(\mathbb E\|Y_T\|_G^4)^{1/2}\Pr(\mathcal B^c)^{1/2}
 =o(v_M),
\]
so $\|Y_T\mathbf1_{\mathcal B}\|_{L^2(G)}\ge c\sqrt{v_M}$.
Lemma~\ref{lem:post-cutoff-response} gives
$\|D_T\mathbf1_{\mathcal B}\|_{L^2(G)}=o(\sqrt{v_M})$. The exact
identity $e_T=Y_T+D_T$ and the reverse triangle inequality in
$L^2(\mathcal B;G)$ now yield
\[
 \|e_T\mathbf1_{\mathcal B}\|_{L^2(G)}
 \ge \|Y_T\mathbf1_{\mathcal B}\|_{L^2(G)}
      -\|D_T\mathbf1_{\mathcal B}\|_{L^2(G)}
 \ge c'\sqrt{v_M}.
\]
Thus $\mathbb E[\|e_T\|_G^2\mathbf1_{\mathcal B}]\gtrsim v_M$.
This step neither assumes independence of $D_T$ and $Z_T$ nor attempts
to control their cross term.
Risk--Fisher equivalence on $\mathcal B$ gives the same lower bound for
the parameter excess risk.  Together with the upper bound, this proves
the last display of the theorem and the total lower bound in the
variance-visible case.  Combining the two cases and using
\eqref{eq:all-horizon-algebra} completes the proof.
\end{proof}

\section{Multiclass Decision Geometry and Metric Transfer}
\label{app:decision-geometry}

Throughout, condition on $S\in\mathcal G_M^{\mathrm{all}}$ and write
$W:=\Sigma_S$, $V^*:=V_M^*$, $G:=\nabla^2\mathcal R_S(V^*)$, and
$k_T:=\min\{M,H_T^{1/a}\}$. Expectations over the training sample are
conditional on $S$.

\subsection{Gaussian pairwise boundary estimates}
\label{app:toc-boundary}

For a fixed $V$ in the local prediction neighborhood of the teacher, write
\[
    w_{cd}(V)=\alpha_{cd}w_{cd}^*+t_{cd},
    \qquad
    \langle t_{cd},w_{cd}^*\rangle_\Sigma=0,
    \qquad
    \alpha_{cd}\in[1/2,3/2].
\]
Then $\|t_{cd}\|_\Sigma$ is the corresponding summand in
$\mathfrak A(V)$.

\begin{lemma}[Uniform multiclass boundary transfer]
\label{lem:multiclass-boundary-transfer}
Under Assumption~\ref{ass:decision-nondegeneracy} and the bounded teacher
prediction condition of Remark~\ref{rem:teacher-bound}, there are constants $\epsilon,c,C>0$
such that for every $V$ satisfying
\[
    \sum_{c<d}\|w_{cd}(V)-w_{cd}^*\|_\Sigma^2\le\epsilon^2,
\]
we have
\[
    \boxed{
    c\mathfrak A(V)\le D(V)\le C\mathfrak A(V),
    \qquad
    c\mathfrak A(V)^2\le\mathcal E_{01}(V)
    \le C\mathfrak A(V)^2.
    }
\]
\end{lemma}

\begin{lemma}[Aggregate angular rigidity]
\label{lem:angular-rigidity}
Let $Q\in\mathbb R^{r\times d}$ satisfy
$\kappa_{\mathrm{dec}}I_r\preceq QQ^\top\preceq R^2I_r$. For
$H\in\mathbb R^{r\times d}$, put $h_{cd}=Q^\top a_{cd}$ and
\[
    \mathcal A_Q(H)^2
    :=\sum_{c<d}\|\Pi_{h_{cd}}^\perp H^\top a_{cd}\|_2^2.
\]
There is $C_{\mathrm{rig}}<\infty$, depending only on
$C,R,\kappa_{\mathrm{dec}}$, such that
\[
    \inf_{\beta\in\mathbb R}\|H-\beta Q\|_F
    \le C_{\mathrm{rig}}\mathcal A_Q(H).
\]
Moreover, $\mathcal A_Q(H)=0$ if and only if $H=\beta Q$ for some
$\beta\in\mathbb R$.
\end{lemma}

\begin{proof}
Write $K=QQ^\top$ and decompose
\[
    L=HQ^\top K^{-1},\qquad H_0=H-LQ.
\]
Then $H_0Q^\top=0$. The row space of $H_0$ is orthogonal to that of
$Q$, while $Q^\top L^\top a_{cd}$ and $h_{cd}$ lie in the row space of
$Q$. Hence the two components remain orthogonal after projection onto
$h_{cd}^\perp$, and the contrast-frame identity gives
\[
\mathcal A_Q(H)^2
=\sum_{c<d}\|\Pi_{h_{cd}}^\perp Q^\top L^\top a_{cd}\|_2^2
 +C\|H_0\|_F^2.
\]
The first sum depends only on $K$ and $L$; its summand is
\[
    a_{cd}^\top LKL^\top a_{cd}
    -\frac{(a_{cd}^\top LKa_{cd})^2}{a_{cd}^\top Ka_{cd}}.
\]
Its kernel requires $L^\top a_{cd}$ to be collinear with $a_{cd}$ for
every pair, because $Q^\top$ is injective. For $C\ge3$, the root
identity $a_{cd}+a_{de}=a_{ce}$ and linear independence of the first
two roots force the eigenvalues of $L^\top$ on these three root lines
to agree. Connectivity of the complete graph then forces one common
eigenvalue on all root lines, which span $\mathbb R^r$. Thus $L=\beta I_r$.
For $C=2$, $r=1$ and this is automatic.

The first sum is therefore a positive-definite quadratic form on the
finite-dimensional quotient by $\operatorname{span}\{I_r\}$. Its
coefficients depend continuously on $K$ in the compact set
$\{K:\kappa_{\mathrm{dec}}I_r\preceq K\preceq R^2I_r\}$, so its
smallest quotient eigenvalue is uniformly positive. Finally,
\[
\|H-\beta Q\|_F^2
=\|H_0\|_F^2+
  \operatorname{tr}((L-\beta I_r)K(L-\beta I_r)^\top)
\le \|H_0\|_F^2+R^2\|L-\beta I_r\|_F^2.
\]
This proves both assertions.
\end{proof}

\begin{lemma}[Uniform Gaussian boundary-patch moments]
\label{lem:boundary-patch-moments}
For each pair $c<d$, let $\mathcal P_{cd}$ be a fixed compact set with
nonempty relative interior in $a_{cd}^\perp$. For $C=2$, take
$\mathcal P_{12}=\{0\}$. Let $g\sim N(0,I_d)$,
$\kappa_{\mathrm{dec}}I_r\preceq QQ^\top\preceq R^2I_r$, and set
\[
    n_{cd}=\frac{Q^\top a_{cd}}{\|Q^\top a_{cd}\|_2},
    \qquad Y=(I-n_{cd}n_{cd}^\top)g.
\]
For every $t\perp Q^\top a_{cd}$ and $p=1,2$, constants $c_p,C_p>0$
exist, independent of $d,Q,t$, such that
\[
    c_p\|t\|_2^p\le
    \mathbb E\left[|t^\top Y|^p\mathbf1_{\{QY\in\mathcal P_{cd}\}}\right]
    \le C_p\|t\|_2^p.
\]
\end{lemma}

\begin{proof}
Decompose $t=t_0+Q^\top v$, where $t_0\in\ker Q$ and
$t_0\perp\operatorname{Range}(Q^\top)$. The condition $t\perp Q^\top a_{cd}$
implies $v^\top K a_{cd}=0$, with $K=QQ^\top$. The Gaussian variable
$t_0^\top Y=t_0^\top g$ is independent of $QY$ and of
$(Q^\top v)^\top Y=v^\top QY$. Thus
\[
\mathbb E[(t^\top Y)^2\mathbf1_{\{QY\in\mathcal P_{cd}\}}]
=\|t_0\|_2^2\Pr(QY\in\mathcal P_{cd})
 +\mathbb E[(v^\top QY)^2\mathbf1_{\{QY\in\mathcal P_{cd}\}}].
\]
On $a_{cd}^\perp$, $QY$ has a nondegenerate Gaussian law. Its
covariance is the restriction of
$K-Ka_{cd}a_{cd}^\top K/(a_{cd}^\top Ka_{cd})$ to that hyperplane;
its eigenvalues are uniformly bounded above and below over the allowed
compact set of $K$. Therefore $\Pr(QY\in\mathcal P_{cd})$ is uniformly
positive. In the second term, no nonzero $v$ satisfying
$v^\top Ka_{cd}=0$ vanishes as a linear functional on all of
$a_{cd}^\perp$: such a $v$ would be proportional to $a_{cd}$ and then
$v^\top Ka_{cd}=0$ forces $v=0$. Since the patch has relative interior,
the truncated quadratic form is positive definite on this tangent
subspace. Compactness of $K$ and of its unit $K$-norm tangent vectors
makes its lower eigenvalue uniform. As
$\|t\|_2^2=\|t_0\|_2^2+v^\top Kv$, this proves the second-moment lower
bound. For $C=2$, the row-space tangent subspace is zero and the
independent $t_0$ term gives the same conclusion.

The upper second-moment bound follows from $Y$ being standard Gaussian on
$n_{cd}^\perp$. Also $\mathbb E|t^\top Y|^4=3\|t\|_2^4$. Interpolation
$\mathbb E|Z|^2\le(\mathbb E|Z|)^{2/3}(\mathbb E|Z|^4)^{1/3}$,
with $Z=t^\top Y\mathbf1_{\{QY\in\mathcal P_{cd}\}}$, gives the
first-moment lower bound; Cauchy--Schwarz gives its upper bound.
\end{proof}

\begin{proof}[Proof of Lemma~\ref{lem:multiclass-boundary-transfer}]
Let $g=\Sigma^{-1/2}x\sim N(0,I_d)$,
$Q=A_*\Sigma^{1/2}$, $\widehat Q=VS\Sigma^{1/2}$, and
$H=\widehat Q-Q$. The contrast-frame identity gives
\[
\sum_{c<d}\|H^\top a_{cd}\|_2^2=C\|H\|_F^2,
\]
so the stated local neighborhood implies $\|H\|_F\le\epsilon/\sqrt C$.
The teacher covariance $K=QQ^\top$ obeys the bounds in
Lemma~\ref{lem:angular-rigidity}. By that lemma choose $\beta$ minimizing
$\|H-\beta Q\|_F$, and write $F=H-\beta Q$. Since
$\mathfrak A(V)\le\sqrt C\|H\|_F$, rigidity gives
$\|F\|_F\lesssim\mathfrak A(V)\lesssim\epsilon$. Also
$\|Q\|_F\ge\sqrt{r\kappa_{\mathrm{dec}}}$, whence
$|\beta|\lesssim\epsilon$. Choose $\epsilon$ small enough that
$1+\beta\in[1/2,3/2]$. The classifier of
$\widehat Q=(1+\beta)Q+F$ is unchanged by positive rescaling, so it is
the classifier of $Q+\widetilde F$, where
$\widetilde F=F/(1+\beta)$. Its aggregate tangent norm is
$\widetilde{\mathfrak A}=\mathfrak A(V)/(1+\beta)$ and
$\|\widetilde F\|_F\lesssim\mathfrak A(V)$.

For each pair $c<d$, choose $y_{cd}^0\in a_{cd}^\perp$ so that the
$c$th and $d$th coordinates of $Uy_{cd}^0$ tie and exceed every other
coordinate by (say) $4\gamma$. This is possible because the rows of $U$
are the vertices of a regular simplex, so each pair is an edge of their
convex hull; for $C=2$ the competitor condition is vacuous. Choose a
small compact relative ball $\mathcal P_{cd}$ around $y_{cd}^0$ on which
the gap is at least $3\gamma$. The covariance family for $Qg$ is compact,
and the Gaussian density restricted to each such patch is uniformly
bounded above and below. The construction and its constants depend only
on the fixed model parameters, not on $d,M,N,T$.

Let $m=\binom C2$ and choose a pair attaining the largest tangent norm
$\|\widetilde t_{cd}\|_2$ of $\widetilde F^\top a_{cd}$; then
$\|\widetilde t_{cd}\|_2\ge\widetilde{\mathfrak A}/\sqrt m$. Fix this
pair and write $h=Q^\top a_{cd}$, $n=h/\|h\|_2$, $g=Xn+Y$, and
$Z=\widetilde t_{cd}^\top Y$. Here $X\sim N(0,1)$ is independent of $Y$.
Write $\widetilde F^\top a_{cd}=\rho h+\widetilde t_{cd}$; then
$|\rho|\lesssim\mathfrak A(V)$ and $1+\rho\in[1/2,3/2]$ after reducing
$\epsilon$. The student pairwise margin is
$(1+\rho)\|h\|_2X+Z$.

Consider the event in the tangential Gaussian coordinates
\[
\mathcal G=\{QY\in\mathcal P_{cd},\ |Z|\le z_0,
                    \|U\widetilde F Y\|_\infty\le\gamma/4\}.
\]
The patch moment lemma gives moments of $Z$ of orders one and two of
size $\|\widetilde t_{cd}\|_2^p$. The discarded large-$Z$ part is at
most $C\|\widetilde t_{cd}\|_2^p e^{-c/\mathfrak A(V)^2}$ by Gaussian
tail integration. Since $\|\widetilde F Y\|$ has covariance operator
bounded by $\|\widetilde F\|_F^2\lesssim\mathfrak A(V)^2$, Gaussian
tail bounds and Cauchy--Schwarz show that imposing the last condition
also removes at most $C\|\widetilde t_{cd}\|_2^p e^{-c/\mathfrak A(V)^2}$ from either moment. For sufficiently small fixed $\epsilon$, therefore,
\[
\mathbb E[|Z|^p\mathbf1_{\mathcal G}]
\gtrsim\|\widetilde t_{cd}\|_2^p,\qquad p=1,2.
\]

For $QY\in\mathcal P_{cd}$ and $|X|\le x_0$, with fixed sufficiently
small $x_0$, the teacher's other class scores remain below the tied pair
by at least $2\gamma$, since $\|Qn\|_2\le R$. Reduce $\epsilon$ further
so that $|X|\|U\widetilde F n\|_\infty\le\gamma/4$ on this interval.
The competitor scores then remain below classes $c,d$ for every
$Y$ in $\mathcal G$. Choose $z_0$ so that
$|Z|/((1+\rho)\|h\|_2)\le x_0$, using the uniform lower bound on
$\|h\|_2$. For each such $Y$, the interval in $X$ between $0$ and
$-Z/((1+\rho)\|h\|_2)$ has length comparable to $|Z|$; throughout it
the teacher and student top classes are respectively $c,d$ in one order
or the other. The standard Gaussian density of $X$ is bounded below on
$[-x_0,x_0]$. Integrating over this interval gives contributions at
least $c|Z|$ to disagreement and $cZ^2$ to excess $0$--$1$ risk, because
on the compact patch the teacher probability gap is comparable to
$|X|$. Averaging over $\mathcal G$ and using
$\|\widetilde t_{cd}\|\gtrsim\mathfrak A(V)$ proves the two lower bounds.

For the upper bounds, for each pair the teacher and student margins are
$X_{cd}$ and $\alpha_{cd}X_{cd}+Z_{cd}$, where the centered Gaussian
variables $X_{cd}$ and $Z_{cd}$ are independent and
$\operatorname{Var}(Z_{cd})=\|t_{cd}\|_\Sigma^2$. The exact Gaussian
sign-change formula bounds the pairwise reversal probability by
$C\|t_{cd}\|_\Sigma$. A top-class disagreement entails at least one
pairwise reversal, so a union bound gives $D(V)\lesssim\mathfrak A(V)$.
The conditional classification regret on a pairwise reversal is bounded
by the teacher pairwise margin; Gaussian integration gives
\[
\mathbb E[|X_{cd}|\mathbf1_{\{|X_{cd}|\le C|Z_{cd}|\}}]
\lesssim\mathbb E Z_{cd}^2\asymp\|t_{cd}\|_\Sigma^2.
\]
Summing over the fixed number of pairs proves the excess-risk upper bound.
\end{proof}

\subsection{Angular energy of the frozen empirical score}
\label{app:toc-angular-score}

For $E\in\mathbb R^{r\times M}$, define the whitened pairwise map
\[
    \mathcal T(E)
    :=\bigl(B^\top E^\top a_{cd}\bigr)_{c<d},
    \qquad B=S\Sigma^{1/2},
\]
and let $\Pi:=\bigoplus_{c<d}\Pi_{cd}^{\perp}$, where each projector is
written in whitened input coordinates. Thus
$\|\Pi\mathcal T(E)\|^2=\mathfrak a(E)^2$.

\begin{lemma}[Filtered score retains angular variance]
\label{lem:angular-score-variance}
Under the assumptions of Lemma~\ref{lem:fullbatch-frozen-variance} and
Assumption~\ref{ass:decision-nondegeneracy}, with
$H_T\to\infty$,
\[
    \boxed{
    \mathbb E\|\Pi\mathcal T(Z_T)\|^2
    \asymp \frac{k_T}{N},
    \qquad
    \mathbb E\|\Pi\mathcal T(Z_T)\|^4
    \lesssim \left(\frac{k_T}{N}\right)^2.
    }
\]
\end{lemma}

\begin{proof}
Regard $\mathcal T$ as a linear map from parameter Frobenius space to the
direct sum of pairwise whitened input spaces. The contrast-frame identity
gives, for every $E$,
\[
    \|\mathcal T(E)\|^2
    =C\|E\|_W^2.
\]
Together with $G\asymp I_r\otimes W$, this implies that
$L:=\mathcal T G^{-1/2}$ has all singular values bounded above and below
by fixed positive constants.

Write $\Gamma=\operatorname{Cov}(\operatorname{vec}\nabla\ell(V^*;z,y)\mid S)$.
The filtered-score covariance and the quasi-Fisher equivalence give
\[
\operatorname{Cov}(G^{1/2}Z_T\mid S)
 =\frac1N f_T(G)G^{-1/2}\Gamma G^{-1/2}f_T(G)
 \asymp \frac1N f_T(G)^2.
\]
The projector $\Pi$ removes one radial coordinate for each class pair,
so $I-\Pi$ has the fixed rank $m=C(C-1)/2$. Therefore
\[
\begin{aligned}
\mathbb E\|\Pi\mathcal T(Z_T)\|^2
&\asymp \frac1N\operatorname{tr}
  \bigl(\Pi Lf_T(G)^2L^*\Pi\bigr)\\
&\le \frac C N\operatorname{tr}f_T(G)^2,
\end{aligned}
\]
and, for the lower bound,
\[
\operatorname{tr}(\Pi Lf_T(G)^2L^*\Pi)
\ge c\operatorname{tr}f_T(G)^2
    -m\|L\|_{\mathrm{op}}^2.
\]
The last estimate follows by subtracting the trace on the fixed-rank
complement of $\Pi$ and using $0\preceq f_T(G)^2\preceq I$. The
effective-dimension estimate gives
$\operatorname{tr}f_T(G)^2\asymp r k_T$, and $k_T\to\infty$ in the
stated regime. Absorbing the fixed rank loss proves the second-moment
claim.

For the fourth moment, the contrast-frame identity and the Fisher
sandwich give the deterministic operator bound
$\|\Pi\mathcal T(E)\|\le C\|E\|_G$. Hence
Lemma~\ref{lem:fullbatch-frozen-variance} immediately yields
\[
 \mathbb E\|\Pi\mathcal T(Z_T)\|^4
 \le C\mathbb E\|Z_T\|_G^4
 \lesssim\left(\frac{k_T}{N}\right)^2.
\]
\end{proof}

\subsection{Transfer to random iterates}
\label{app:toc-random-transfer}

\begin{lemma}[Expected classification metrics from angular moments]
\label{lem:expected-metric-transfer}
Let $V$ be a random iterate whose total teacher-prediction error
has second moment $Q\to0$ and fourth moment $O(Q^2)$. Suppose
$\mathbb E\mathfrak A(V)^2\asymp Q$. Then
\[
    \boxed{
    \mathbb E D(V)\asymp\sqrt Q,
    \qquad
    \mathbb E\mathcal E_{01}(V)\asymp Q.
    }
\]
\end{lemma}

\begin{proof}
Let $\mathcal U_\epsilon$ be the local event in
Lemma~\ref{lem:multiclass-boundary-transfer}, let $R$ be the total
teacher-prediction error norm defining this event, and put
$A=\mathfrak A(V)$. Since $A\le R$, the fourth-moment hypothesis and
Markov's inequality give
\[
 \Pr(\mathcal U_\epsilon^c)
 \le \epsilon^{-4}\mathbb E R^4=O(Q^2),
 \qquad
 \mathbb E[A^2\mathbf1_{\mathcal U_\epsilon^c}]
 \le \epsilon^{-2}\mathbb E R^4=O(Q^2).
\]
Thus $\mathbb E[A^2\mathbf1_{\mathcal U_\epsilon}]\asymp Q$, while
$\mathbb E[A^4\mathbf1_{\mathcal U_\epsilon}]\le\mathbb E R^4=O(Q^2)$.
Hölder's inequality in the form
\[
 \mathbb E[A^2\mathbf1_{\mathcal U_\epsilon}]
 \le
 \bigl(\mathbb E[A\mathbf1_{\mathcal U_\epsilon}]\bigr)^{2/3}
 \bigl(\mathbb E[A^4\mathbf1_{\mathcal U_\epsilon}]\bigr)^{1/3}
\]
therefore gives
$\mathbb E[A\mathbf1_{\mathcal U_\epsilon}]\gtrsim\sqrt Q$; the upper
bound follows from Cauchy--Schwarz. On $\mathcal U_\epsilon$, apply the
two deterministic comparisons in
Lemma~\ref{lem:multiclass-boundary-transfer}. On its complement, both
classification metrics are bounded by one, so their contributions are
$O(Q^2)$. This proves the two claimed expectations.
\end{proof}

\begin{theorem}[Optimization-limited angular and classification scaling]
\label{thm:fullbatch-classification-resolved}
Suppose the assumptions and asymptotic regime of
Theorem~\ref{thm:fullbatch-estimation-risk} and
Assumption~\ref{ass:decision-nondegeneracy} hold. Conditional on
$S\in\mathcal G_M^{\mathrm{all}}$, put
\[
    Q_{T,N}:=H_T^{-q}+\frac{k_T}{N}.
\]
Then
\[
    \boxed{
    \mathbb E[\mathfrak A(V_T)^2\mid S]
    \asymp \Delta_M+Q_{T,N},
    }
\]
and
\[
    \boxed{
    \mathbb E[\mathcal E_{01}(V_T)\mid S]
    \asymp \Delta_M+Q_{T,N},
    \qquad
    \mathbb E[D(V_T)\mid S]
    \asymp (\Delta_M+Q_{T,N})^{1/2}.
    }
\]
In particular, since $\Delta_M\asymp M^{1-b}$ on the sketch event,
these are the same total scaling rates as the cross-entropy theorem.
\end{theorem}

\begin{proof}
All vectors in this proof are in whitened input coordinates. For each
pair, let
\[
    h_{cd}:=Q^\top a_{cd},\qquad
    r_{cd}:=h_{cd}-B^\top(V_M^{\mathrm{LS}})^\top a_{cd},
    \qquad Q=A_*\Sigma^{1/2}.
\]
Gaussian conditional regression gives $r_{cd}\perp\operatorname{Range}(B^\top)$,
$\langle r_{cd},h_{cd}\rangle=\|r_{cd}\|^2$, and
\[
    \sum_{c<d}\|r_{cd}\|^2=C\Delta_M.
\]
The teacher-normal lower bound and $\Delta_M\to0$ imply
\[
    \|\Pi r\|^2
    =\sum_{c<d}\left(\|r_{cd}\|^2-
       \frac{\|r_{cd}\|^4}{\|h_{cd}\|^2}\right)
    \asymp\Delta_M,
\]
where $r=(r_{cd})_{c<d}$.

Let $H_M:=V^*-V_M^{\mathrm{LS}}$ be the oracle shift from
Appendix~\ref{app:preliminary}, and set $Y:=B_T^{\mathrm{fr}}+Z_T$.
The oracle-aware logit decomposition
\eqref{eq:oracle-aware-logit-decomposition}, written pairwise, gives
\[
    \bigl(B^\top V_T^\top a_{cd}-h_{cd}\bigr)_{c<d}
    =\mathcal T(H_M+Y+D_T)-r.
\]
Oracle compatibility gives $\|H_M\|_W\lesssim\Delta_M$. By
Lemma~\ref{lem:frozen-angular-bias},
Lemma~\ref{lem:angular-score-variance}, and
$\mathbb E[Z_T\mid S]=0$,
\[
    \mathbb E\|\Pi\mathcal T(Y)\|^2
    =\mathfrak a(B_T^{\mathrm{fr}})^2
      +\mathbb E\|\Pi\mathcal T(Z_T)\|^2
    \asymp Q_{T,N}.
\]
On the empirical-Hessian event, the nonlinear-response estimate gives
$\|D_T\|_{L_2(W)}=o(\sqrt{Q_{T,N}})$. The same holds without restricting
to this event: on its complement, use
$D_T=e_T-B_T^{\mathrm{fr}}-Z_T$, the unconditional fourth-moment bound
for $e_T$, the fourth-moment bound for $Z_T$, and the polynomially small
failure probability from Lemma~\ref{lem:uniform-empirical-hessian} with
Cauchy--Schwarz. This complement contributes $o(Q_{T,N})$ to
$\mathbb E\|D_T\|_W^2$. The contrast-frame identity and oracle
compatibility now imply
\[
    \|\Pi\mathcal T(H_M+D_T)\|_{L_2}
    \le C\bigl(\Delta_M+\|D_T\|_{L_2(W)}\bigr)
    =o(\sqrt{Q_{T,N}}),
\]
using $\Delta_M\lesssim H_T^{-q}\le Q_{T,N}$. Hence
$\mathbb E\|\Pi\mathcal T(H_M+Y+D_T)\|^2\asymp Q_{T,N}$.

For any student-observable vector $s_{cd}\in\operatorname{Range}(B^\top)$,
orthogonality of $r_{cd}$ gives
\[
    \langle\Pi_{cd}^{\perp}r_{cd},
             \Pi_{cd}^{\perp}s_{cd}\rangle
    =-\frac{\|r_{cd}\|^2\langle s_{cd},h_{cd}\rangle}
            {\|h_{cd}\|^2}.
\]
Summing over the fixed number of pairs and using the teacher-normal lower
bound yields
\[
\left|\mathbb E\left\langle\Pi r,
       \Pi\mathcal T(H_M+Y+D_T)\right\rangle\right|
\le C\Delta_M\,
       \|H_M+Y+D_T\|_{L_2(W)}
\lesssim\Delta_M\sqrt{Q_{T,N}}
 =o(\Delta_M+Q_{T,N}).
\]
Expanding the squared norm proves the angular-energy assertion.

For the moment condition in Lemma~\ref{lem:expected-metric-transfer},
let $d_T=(B^\top V_T^\top a_{cd}-h_{cd})_{c<d}$. Orthogonality of the
regression residuals to $\operatorname{Range}(B^\top)$ gives
\[
    \|d_T\|^2
    =C\|V_T-V_M^{\mathrm{LS}}\|_W^2+C\Delta_M.
\]
Oracle compatibility and $G\asymp I_r\otimes W$ imply
$\|V_T-V_M^{\mathrm{LS}}\|_{L_4(W)}
\lesssim Q_{T,N}^{1/2}+\Delta_M$. The fourth-moment estimate in
Theorem~\ref{thm:fullbatch-estimation-risk} therefore yields
\[
    \mathbb E\|d_T\|^4
    \lesssim (Q_{T,N}+\Delta_M)^2.
\]
Because $\mathfrak A(V_T)\le\|d_T\|$, this also proves
\[
    \mathbb E[\mathfrak A(V_T)^4\mid S]
    \lesssim (Q_{T,N}+\Delta_M)^2.
\]
In particular, Markov's inequality bounds the probability of leaving the
local decision-geometry neighborhood by $O((Q_{T,N}+\Delta_M)^2)$.
Applying Lemma~\ref{lem:expected-metric-transfer} with
$Q=\Delta_M+Q_{T,N}$ proves both classification statements.
\end{proof}

\begin{theorem}[All-horizon angular and classification scaling]
\label{thm:fullbatch-classification}
Suppose the assumptions and asymptotic regime of
Theorem~\ref{thm:fullbatch-total-risk} and
Assumption~\ref{ass:decision-nondegeneracy} hold. Conditional on
$S\in\mathcal G_M^{\mathrm{all}}$, put
\[
    \mathcal Q_{M,N,T}:=k_T^{1-b}+\frac{k_T}{N}.
\]
Then
\[
\boxed{
\mathbb E[\mathfrak A(V_T)^2\mid S]\asymp\mathcal Q_{M,N,T},
\qquad
\mathbb E[\mathcal E_{01}(V_T)\mid S]\asymp\mathcal Q_{M,N,T},
}
\]
and
\[
\boxed{
\mathbb E[D(V_T)\mid S]\asymp\mathcal Q_{M,N,T}^{1/2}.
}
\]
\end{theorem}

\begin{proof}
Reduce $c_{\mathrm{sat}}$ from
Lemma~\ref{lem:post-cutoff-bias}, if necessary, so that
$H_T<c_{\mathrm{sat}}M^a$ implies $M\ge DH_T^{1/a}$ for every
optimization-limited result used below. In that range,
Theorem~\ref{thm:fullbatch-classification-resolved} and
$\Delta_M\asymp M^{1-b}\lesssim H_T^{-q}$ give the claim, since
\[
\Delta_M+H_T^{-q}+\frac{k_T}{N}
\asymp k_T^{1-b}+\frac{k_T}{N}.
\]

Now suppose $H_T\ge c_{\mathrm{sat}}M^a$. Then
$k_T\asymp M$, $H_T^{-q}\lesssim M^{1-b}$, and, writing
$v_M=M/N$,
\[
    \mathcal Q_{M,N,T}\asymp\Delta_M+v_M.
\]
Use the notation from the preceding proof: $r$ is the
conditional-regression residual, $H_M=V^*-V_M^{\mathrm{LS}}$, and
$\mathcal T$ and $\Pi$ are the student-observable pairwise map and the
teacher-tangent projection. Thus
\[
\bigl(B^\top V_T^\top a_{cd}-h_{cd}\bigr)_{c<d}
=\mathcal T(H_M+B_T^{\mathrm{fr}}+Z_T+D_T)-r.
\tag{E.1}\label{eq:post-cutoff-angular-decomp}
\]
To make the regression geometry explicit, for one pair write
$h=s^{\mathrm{LS}}+r$ with
$s^{\mathrm{LS}}\in\operatorname{Range}(B^\top)$ and
$r\perp\operatorname{Range}(B^\top)$. Then $r\perp s^{\mathrm{LS}}$,
$\langle h,r\rangle=\|r\|^2$, and, for every
$s\in\operatorname{Range}(B^\top)$,
\[
 \langle\Pi_h^\perp r,\Pi_h^\perp s\rangle
 =-\frac{\|r\|^2\langle h,s\rangle}{\|h\|^2}.
\]
Summing this identity over the fixed set of pairs, using the uniform
lower bound on $\|h_{cd}\|$, and applying Cauchy--Schwarz and the
contrast-frame identity gives
$\|\Pi r\|^2\asymp\Delta_M$ and, for every student-observable $H$,
\[
\left|\langle\Pi r,\Pi\mathcal T(H)\rangle\right|
\le C\Delta_M\|H\|_W.
\tag{E.2}\label{eq:post-cutoff-near-orthogonality}
\]
The all-horizon moment bounds in
Theorem~\ref{thm:fullbatch-total-risk}, oracle compatibility, and the
contrast-frame identity imply
\[
\mathbb E\|H_M+B_T^{\mathrm{fr}}+Z_T+D_T\|_W^2
\lesssim\mathcal Q_{M,N,T}.
\]
Consequently, if $\Delta_M\gtrsim v_M$, expansion of
\eqref{eq:post-cutoff-angular-decomp} and
\eqref{eq:post-cutoff-near-orthogonality} gives
\[
\mathbb E\mathfrak A(V_T)^2
\ge c\Delta_M-C\Delta_M\sqrt{\mathcal Q_{M,N,T}}
\gtrsim\mathcal Q_{M,N,T}.
\]

It remains to treat $v_M\gtrsim\Delta_M$. Put
$Y_T^\angle=H_M+B_T^{\mathrm{fr}}+Z_T$. Since $Z_T$ is centered,
Lemma~\ref{lem:angular-score-variance} gives
\[
\mathbb E\|\Pi\mathcal T(Y_T^\angle)\|^2
=\|\Pi\mathcal T(H_M+B_T^{\mathrm{fr}})\|^2
\mathrel{+}\mathbb E\|\Pi\mathcal T(Z_T)\|^2
\gtrsim v_M.
\]
Lemmas~\ref{lem:post-cutoff-bias} and
\ref{lem:fullbatch-frozen-variance}, together with oracle
compatibility, give
\[
 \mathbb E\|\Pi\mathcal T(Y_T^\angle)\|^4
 \le C\left(
   \|H_M\|_W^4+\|B_T^{\mathrm{fr}}\|_W^4
   +\mathbb E\|Z_T\|_W^4\right)
 \lesssim v_M^2,
\]
where $v_M\gtrsim\Delta_M$ and the post-cutoff frozen bias is
$O(\Delta_M)$ in squared norm. Cauchy--Schwarz with the polynomially
small failure probability makes the bad-event second-moment contribution
$o(v_M)$. On the good event,
Lemma~\ref{lem:post-cutoff-response} gives
$\|D_T\|_{L^2(W)}=o(\sqrt{v_M})$. The reverse triangle inequality
in the restricted $L^2$ space then shows
\[
\mathbb E\|\Pi\mathcal T(H_M+B_T^{\mathrm{fr}}+Z_T+D_T)\|^2
\gtrsim v_M.
\]
Adding the residual $r$ and applying
\eqref{eq:post-cutoff-near-orthogonality} changes this by at most
$C\Delta_M\sqrt{\mathcal Q_{M,N,T}}=o(\mathcal Q_{M,N,T})$.
This proves the matching angular lower bound when the sampling term is
visible. The angular upper bound follows directly from the oracle-aware
logit decomposition \eqref{eq:oracle-aware-squared-logit} and
Theorem~\ref{thm:fullbatch-total-risk}.

The same decomposition and the fourth-moment assertion of that theorem
give
\[
\mathbb E\left[
\sum_{c<d}\|w_{cd}(V_T)-w_{cd}^*\|_\Sigma^2
\right]^2
\lesssim\mathcal Q_{M,N,T}^2.
\]
Lemma~\ref{lem:expected-metric-transfer}, with
$Q=\mathcal Q_{M,N,T}$, now yields the two classification statements.
\end{proof}

\section{Proofs of the Full-Batch Scaling Theorems}
\label{app:scaling-proofs}

\begin{proof}[Proof of Theorem~\ref{thm:population-gd-scaling}]
Reduce $c_{\mathrm{sat}}$ from
Lemma~\ref{lem:post-cutoff-bias}, if necessary, so that
$H_T<c_{\mathrm{sat}}M^a$ implies the non-saturation condition in
Theorem~\ref{thm:population-gd-bias}.  In this range, that theorem and
the exact approximation--estimation decomposition give
\[
\mathcal E_{\mathrm{CE}}(V_T^{\mathrm{pop}})
\asymp M^{1-b}+H_T^{-q}
\asymp H_T^{-q}
\asymp k_T^{1-b}.
\]
If $H_T\ge c_{\mathrm{sat}}M^a$, then
Lemma~\ref{lem:post-cutoff-bias} gives
\[
\mathcal R_S(V_T^{\mathrm{pop}})-\mathcal R_S(V_M^*)
\lesssim M^{1-b}.
\]
Together with
$\operatorname{Approx}_M\asymp M^{1-b}$ and the nonnegativity of the
parameter excess, this yields
\[
\mathcal E_{\mathrm{CE}}(V_T^{\mathrm{pop}})
\asymp M^{1-b}
\asymp k_T^{1-b}.
\]
The two cases prove the all-horizon claim.
\end{proof}

\begin{proof}[Proof of Theorem~\ref{thm:metric-scaling}]
By the exact approximation--estimation decomposition,
\[
\mathbb E\mathcal E_{\mathrm{CE}}(V_T)
=\operatorname{Approx}_M+
 \mathbb E[\mathcal R_S(V_T)-\mathcal R_S(V_M^*)\mid S].
\]
The approximation theorem gives
$\operatorname{Approx}_M\asymp M^{1-b}$ on the sketch event.
The all-horizon result, Theorem~\ref{thm:fullbatch-total-risk}, gives
\[
M^{1-b}+
\mathbb E[\mathcal R_S(V_T)-\mathcal R_S(V_M^*)\mid S]
\asymp k_T^{1-b}+\frac{k_T}{N}.
\]
Substitution proves the claim.
\end{proof}

\begin{proof}[Proof of Theorem~\ref{thm:classification-scaling}]
On the stated sketch event, $\Delta_M\asymp M^{1-b}$ by the
approximation theorem. The angular-energy theorem
\ref{thm:fullbatch-classification} gives
\[
\mathbb E\mathcal E_{01}(V_T)\asymp
k_T^{1-b}+\frac{k_T}{N},
\qquad
\mathbb E D(V_T)\asymp
\left(k_T^{1-b}+\frac{k_T}{N}\right)^{1/2}.
\]
This is exactly the claimed pair of rates.
\end{proof}

\begin{proof}[Proof of Corollary~\ref{cor:rate-optimal-tuning}]
Put $k:=k_T$.  Theorem~\ref{thm:metric-scaling} gives
\[
    k^{1-b}+\frac{k}{N}
    \gtrsim N^{-(b-1)/b}.
\]
Choosing $M\asymp N^{1/b}$ and $H_T\asymp N^{a/b}$ makes
$k_T\asymp N^{1/b}$ and both terms of order
$N^{-(b-1)/b}$. Moreover,
\[
    \frac{M^2\log^3(N+T)}{Nk_T}
    \asymp N^{1/b-1}\log^3(N+T)\longrightarrow0,
\]
which tends to zero under the corollary's sample-size condition. For example,
a fixed sufficiently small constant step size gives $T\asymp N^{a/b}$ and
satisfies it. The classification and disagreement rates follow from
Theorems~\ref{thm:metric-scaling} and
\ref{thm:classification-scaling}.
\end{proof}

\section{Auxiliary Lemmas}
\label{app:auxiliary}

This appendix collects structural lemmas used in the approximation,
full-batch bias, empirical-score, and decision-geometry analyses.

% ============================================================
\subsection{Softmax Geometry}
\label{app:toc-softmax}

For $h\in\mathbb{R}^{r}$, define
\[
    p(h):=\operatorname{softmax}(Uh),
\]
and
\[
    \widetilde C(h)
    :=
    U^\top
    \left(
        \operatorname{Diag}(p(h))
        -
        p(h)p(h)^\top
    \right)
    U.
\]

\begin{lemma}[Softmax curvature on the contrast space]
\label{lem:softmax-geometry}
For every $h\in\mathbb{R}^{r}$,
\[
    0
    \preceq
    \widetilde C(h)
    \preceq
    C I_r.
\]
Moreover, for every $L<\infty$, there exists $c_L>0$ such that
\[
    \|h\|_2\le L
    \qquad\Longrightarrow\qquad
    \widetilde C(h)
    \succeq
    c_L I_r.
\]
Consequently, the log-partition function
\[
    \Phi(h)
    =
    \log\sum_{c=1}^{C}\exp((Uh)_c)
\]
satisfies
\[
    D_\Phi(h_1,h_2)
    \lesssim
    \|h_1-h_2\|_2^2
\]
globally, and
\[
    D_\Phi(h_1,h_2)
    \gtrsim_L
    \|h_1-h_2\|_2^2
\]
whenever the line segment joining $h_1$ and $h_2$ is contained in
$\{h:\|h\|_2\le L\}$.
\end{lemma}

\begin{proof}
The Hessian of $\Phi$ is
\[
    \nabla^2\Phi(h)
    =
    \widetilde C(h).
\]
The global upper bound follows because
$\operatorname{Diag}(p)-pp^\top$ is the covariance matrix of a categorical
random variable and hence has uniformly bounded operator norm.

For the lower bound, if $\|h\|_2\le L$, then all centered class logits
$Uh$ are uniformly bounded. Hence every class probability satisfies
\[
    p_c(h)\ge c_{L,C}>0.
\]
For every $z\in\mathbf{1}_C^\perp$,
\[
    z^\top
    \left(
        \operatorname{Diag}(p)-pp^\top
    \right)
    z
    \ge
    c_{L,C}\|z\|_2^2.
\]
Since $U$ is an orthonormal basis of $\mathbf{1}_C^\perp$, the claimed
lower bound follows.
\end{proof}

% ============================================================
\subsection{Gaussian Moment Bounds}
\label{app:toc-gaussian-moments}

\begin{lemma}[Fixed-dimensional Gaussian fourth moment]
\label{lem:gaussian-fourth-moment}
Let $g$ be a centered Gaussian vector in $\mathbb{R}^{r}$, where $r$ is
fixed. Then
\[
    \mathbb{E}\|g\|_2^4
    \le
    C_r
    \left(
        \mathbb{E}\|g\|_2^2
    \right)^2.
\]
Consequently, for every event $\mathcal E$,
\[
    \mathbb{E}
    \left[
        \|g\|_2^2
        \mathbf{1}_{\mathcal E^c}
    \right]
    \le
    C_r
    \mathbb{P}(\mathcal E^c)^{1/2}
    \mathbb{E}\|g\|_2^2.
\]
\end{lemma}

\begin{proof}
The first claim follows from Isserlis' formula. The second follows by
Cauchy--Schwarz.
\end{proof}

% ============================================================
\subsection{Bounded Population Sublevel Sets}
\label{app:toc-sublevels}

\begin{lemma}[Uniform boundedness of the sketched population optimum]
\label{lem:bounded-sketch-optimum}
Suppose $\|A_*\Sigma^{1/2}\|_{\mathrm{op}}\le R$. Then there exists a constant
$R_{\mathrm{pop}}<\infty$, independent of $M$ and $N$, such that
\[
    \boxed{
    \|V_M^*\Sigma_S^{1/2}\|_F
    \le
    R_{\mathrm{pop}}.
    }
\]
More generally, every population sublevel set
\[
    \left\{
        V:
        \mathcal R_S(V)
        \le
        \mathcal R_S(0)
    \right\}
\]
is uniformly bounded in the prediction norm
$\|V\Sigma_S^{1/2}\|_F$. Moreover, for every fixed $R_0<\infty$,
there are constants $0<c_{R_0}<C_{R_0}<\infty$ such that, whenever
$\|V\Sigma_S^{1/2}\|_F\le R_0$,
\[
    c_{R_0}(I_r\otimes\Sigma_S)
    \preceq
    \nabla^2\mathcal R_S(V)
    \preceq
    C_{R_0}(I_r\otimes\Sigma_S).
\]
\end{lemma}

\begin{proof}
Put $s:=\|V\Sigma_S^{1/2}\|_F$ and $h_*:=A_*x$. By the teacher-bound
assumption, $\mathbb E\|h_*\|_2^2\le rR^2$. Choose $L$ so large that
$\mathbb P(\|h_*\|_2>L)$ is small enough, uniformly in $M$. On the event
$\mathcal E:=\{\|h_*\|_2\le L\}$, every teacher class probability is
bounded below by a constant $\kappa_L>0$. For centered student logits
$a=UVz$, the expected loss conditional on $x$ then satisfies
\[
    \Phi(Vz)-p_*(x)^\top a
    \ge
    \kappa_L\sum_{c=1}^C(\max_k a_k-a_c)
    \ge c_L\|Vz\|_2
    \qquad\text{on }\mathcal E.
\]
The random vector $Vz$ is an $r$-dimensional centered Gaussian with
$\mathbb E\|Vz\|_2^2=s^2$ and
$\mathbb E\|Vz\|_2\ge c_rs$. Cauchy--Schwarz gives
\[
    \mathbb E[\|Vz\|_2\mathbf 1_{\mathcal E}]
    \ge
    c_rs-s\,\mathbb P(\mathcal E^c)^{1/2}
    \ge c s,
\]
after increasing $L$. The loss is nonnegative outside $\mathcal E$,
so $\mathcal R_S(V)\ge cs$. Thus every sublevel set below
$\mathcal R_S(0)=\log C$ is uniformly bounded in prediction norm, and in
particular the same is true of $V_M^*$.

For the Hessian lower bound, write $z=\Sigma_S^{1/2}g$ with
$g\sim\mathcal N(0,I_M)$ and let $P$ project onto the row space of
$V\Sigma_S^{1/2}$. Its rank is at most $r$, and the assumed bound $s\le R_0$ bounds
$\|V\Sigma_S^{1/2}\|_{\mathrm{op}}$. Choose a fixed $L_0>0$.
On $\mathcal E_V:=\{\|Pg\|_2\le L_0\}$, the student logits are bounded,
so Lemma~\ref{lem:softmax-geometry} gives
$\widetilde C(Vz)\succeq cI_r$. Since $P$ has rank at most the fixed
$r$, Gaussian rotational invariance gives
\[
    \mathbb E[gg^\top\mathbf 1_{\mathcal E_V}]
    \succeq c_{r,L_0}I_M.
\]
Therefore, for every perturbation $E$,
\[
    \mathbb E[(Ez)^\top\widetilde C(Vz)(Ez)]
    \ge c\|E\Sigma_S^{1/2}\|_F^2.
\]
The global upper bound on $\widetilde C$ gives the reverse inequality.
\end{proof}

% ============================================================
\subsection{Gaussian Stein Identity}
\label{app:toc-stein}

Define
\[
    B(V)
    :=
    U^\top
    \mathbb{E}
    \left[
        \operatorname{Diag}(p_V(z))
        -
        p_V(z)p_V(z)^\top
    \right]
    U.
\]

\begin{lemma}[Population Stein identity]
\label{lem:stein-softmax}
For every $V\in\mathbb{R}^{r\times M}$,
\[
    \boxed{
    \nabla\mathcal R_S(V)
    =
    B(V)V\Sigma_S
    -
    B_*A_*\Sigma S^\top,
    }
    \tag{G.1}\label{eq:stein-identity}
\]
where
\[
    B_*
    :=
    U^\top
    \mathbb{E}
    \left[
        \operatorname{Diag}(p_*(x))
        -
        p_*(x)p_*(x)^\top
    \right]
    U.
\]
Consequently, at the sketched population optimum,
\[
    \boxed{
    B_MV_M^*\Sigma_S
    =
    B_*A_*\Sigma S^\top,
    }
\]
where
\[
    B_M:=B(V_M^*).
\]
If $\Sigma_S$ is invertible,
\[
    \boxed{
    V_M^*
    =
    B_M^{-1}B_*V_M^{\mathrm{LS}},
    \qquad
    V_M^{\mathrm{LS}}
    =
    A_*\Sigma S^\top\Sigma_S^{-1}.
    }
\]
\end{lemma}

\begin{proof}
For $z\sim\mathcal N(0,\Sigma_S)$, Gaussian Stein's identity gives
\[
    \mathbb{E}
    \left[
        U^\top p_V(z)z^\top
    \right]
    =
    B(V)V\Sigma_S.
\]
Similarly,
\[
\begin{aligned}
    \mathbb{E}
    \left[
        U^\top yz^\top
    \right]
    &=
    \mathbb{E}
    \left[
        U^\top p_*(x)x^\top
    \right]
    S^\top
    \\
    &=
    B_*A_*\Sigma S^\top.
\end{aligned}
\]
Subtracting the two terms gives \eqref{eq:stein-identity}. The remaining identities
follow from the first-order optimality condition.
\end{proof}

% ============================================================
\subsection{Fisher Geometry}
\label{app:toc-fisher}

Let
\[
    G_M
    :=
    \nabla^2\mathcal R_S(V_M^*).
\]

\begin{lemma}[Fisher sandwich]
\label{lem:fisher-sandwich}
Suppose $\|A_*\Sigma^{1/2}\|_{\mathrm{op}}\le R$. On $\mathcal G_M$ there exist
constants $0<c<C<\infty$, independent of $M$ and $N$, such that
\[
    \boxed{
    c(I_r\otimes\Sigma_S)
    \preceq
    G_M
    \preceq
    C(I_r\otimes\Sigma_S).
    }
\]
In particular, the nonzero spectrum of $G_M$ is comparable, up to fixed
class-space multiplicity, to that of $\Sigma_S$.
\end{lemma}

\begin{proof}
Write $z=W^{1/2}g$, $W=\Sigma_S$, and let
$A=V_M^*W^{1/2}$. The bounded-population-sublevel lemma gives
$\|A\|_F\le R_0$ uniformly. For any matrix $D$, the Hessian quadratic
form is
\[
\langle D,G_MD\rangle
=\mathbb E[(DW^{1/2}g)^\top
  \widetilde C(Ag)(DW^{1/2}g)].
\]
The global softmax curvature bound immediately gives the upper estimate
$C\|DW^{1/2}\|_F^2$.

For the lower estimate, choose $L$ so large that
$\Pr(\|Ag\|_2>L)\le R_0^2/L^2$ is at most a sufficiently small
constant, uniformly in $A$. For every unit $v\in\mathbb R^M$,
Cauchy--Schwarz and $\mathbb E(v^\top g)^4=3$ give
\[
\mathbb E[(v^\top g)^2\mathbf1_{\{\|Ag\|\le L\}}]
\ge 1-\sqrt3\Pr(\|Ag\|>L)^{1/2}\ge\frac12.
\]
Thus the truncated covariance of $g$ is bounded below by $I_M/2$;
this controls directions both inside and orthogonal to the row space of
$A$. On this event Lemma~\ref{lem:softmax-geometry} gives
$\widetilde C(Ag)\succeq c_LI_r$, and hence
$\langle D,G_MD\rangle\ge c\|DW^{1/2}\|_F^2$. This proves the
Fisher sandwich with constants independent of $M,N,T,d$.
\end{proof}

\subsection{Finite-Rank Fisher Correction}
\label{app:toc-fisher-correction}

\begin{lemma}[Finite-rank Fisher correction]
\label{lem:fisher-finite-rank}
Define
\[
    G_{0,M}[E]
    :=
    B_ME\Sigma_S.
\]
Then
\[
    \boxed{
    G_M
    =
    G_{0,M}
    +
    \mathcal K_M,
    }
\]
where
\[
    \mathcal K_M[E]
    =
    DB(V_M^*)[E]
    V_M^*\Sigma_S.
    \]
Moreover,
\[
    \operatorname{rank}(\mathcal K_M)
    \le
    r^2,
\]
and
\[
    \operatorname{Range}(\mathcal K_M)
    \subseteq
    \left\{
        HV_M^*\Sigma_S:
        H\in\mathbb{R}^{r\times r}
    \right\}.
\]
\end{lemma}

\begin{proof}
Differentiate the Stein identity
\[
    \nabla\mathcal R_S(V)
    =
    B(V)V\Sigma_S
    -
    B_*A_*\Sigma S^\top
\]
at $V=V_M^*$. Since $DB(V_M^*)[E]\in\mathbb{R}^{r\times r}$, the range
of the correction lies in an at-most $r^2$ dimensional subspace.
\end{proof}

% ============================================================
\subsection{Quasi-Fisher Noise Covariance}
\label{app:toc-quasi-fisher}

Define the score at the sketched population optimum by
\[
    \upsilon_M
    :=
    -
    \nabla_V
    \ell(V_M^*;z,y),
\]
and let
\[
    C_M
    :=
    \mathbb{E}
    \left[
        \operatorname{vec}(\upsilon_M)
        \operatorname{vec}(\upsilon_M)^\top
    \right].
\]

\begin{lemma}[Quasi-Fisher covariance equivalence]
\label{lem:quasi-fisher}
Suppose $\|A_*\Sigma^{1/2}\|_{\mathrm{op}}\le R$. On $\mathcal G_M$,
\[
    \boxed{
    cG_M
    \preceq
    C_M
    \preceq
    CG_M.
    }
\]
\end{lemma}

\begin{proof}
Since $V_M^*$ is a population minimizer,
$\mathbb E[\upsilon_M\mid S]=0$, so $C_M$ is the score covariance. Notice,
however, that misspecification generally makes the conditional score mean
nonzero.  If
\[
    \bar p(z):=\mathbb E[y\mid z],
    \qquad p_M(z):=p_{V_M^*}(z),
\]
then
\[
    \mathbb E[\upsilon_M\mid z,S]
    =U^\top\bigl(\bar p(z)-p_M(z)\bigr)z^\top,
\]
which need not vanish even though its expectation over $z$ does. Write
$B=S\Sigma^{1/2}$, $Q=A_*\Sigma^{1/2}$, $W=BB^\top$, and let
$\epsilon\sim N(0,I_d)$ be the full-input whitened Gaussian. Then
$g=W^{-1/2}z=W^{-1/2}B\epsilon\sim N(0,I_M)$. Gaussian regression gives
\[
    Q\epsilon=C_Sg+\xi,
    \qquad C_S:=QB^\top W^{-1/2},
    \qquad \xi\perp g,
\]
where $\xi$ is centered Gaussian. Conditional expectation is an
orthogonal projection in $L^2$, so
$\|C_S\|_F^2\le\|Q\|_F^2$ and
$\mathbb E\|\xi\|_2^2\le\|Q\|_F^2$. The teacher prediction bound and fixed
$r$ bound $\|Q\|_F$ uniformly. Choose fixed $L$ so that
$\Pr(\|C_Sg\|_2>L)$ is sufficiently small and
$\Pr(\|\xi\|_2\le L)\ge1/2$. For every unit $v\in\mathbb R^M$,
Cauchy--Schwarz and $\mathbb E(v^\top g)^4=3$ imply
\[
\mathbb E[(v^\top g)^2\mathbf1_{\{\|C_Sg\|\le L\}}]
\ge 1-\sqrt{3\Pr(\|C_Sg\|>L)}\ge\frac12.
\]
The event involving $\xi$ is independent of $g$. On the intersection of
these two events, $\|Q\epsilon\|_2\le2L$, and softmax curvature on the
fixed contrast space gives
$U^\top\operatorname{Cov}(y\mid x)U\succeq c_LI_r$.

For any test matrix $D\in\mathbb R^{r\times M}$, conditioning the squared
score on $z$ gives the exact decomposition
\[
\begin{aligned}
\langle D,C_MD\rangle
&=\mathbb E[(Dz)^\top U^\top\operatorname{Cov}(y\mid z)U(Dz)]\\
&\quad+\mathbb E\!\left[
  \bigl((Dz)^\top U^\top(\bar p(z)-p_M(z))\bigr)^2
  \right]\\
&\ge\mathbb E[(Dz)^\top U^\top\operatorname{Cov}(y\mid z)U(Dz)]\\
&\ge \mathbb E[(Dz)^\top U^\top\operatorname{Cov}(y\mid x)U(Dz)]\\
&\ge c\,\mathbb E[\|DW^{1/2}g\|_2^2
                 \mathbf1_{\{\|C_Sg\|\le L\}}
                 \mathbf1_{\{\|\xi\|\le L\}}]
\ge c'\|DW^{1/2}\|_F^2.
\end{aligned}
\]
Here the second inequality is the law of total covariance,
$\operatorname{Cov}(y\mid z)\succeq
\mathbb E[\operatorname{Cov}(y\mid x)\mid z]$; the omitted conditional-mean
covariance is positive semidefinite.  In the final step, apply the
truncated covariance bound to each row of
$DW^{1/2}$ and use independence of $\xi$ and $g$. Conversely, bounded
softmax residuals control the full score, including its conditional mean,
and imply
$\langle D,C_MD\rangle\le C\|DW^{1/2}\|_F^2$. The Fisher sandwich
$G_M\asymp I_r\otimes W$ now yields
$cG_M\preceq C_M\preceq CG_M$, with constants independent of
$M,N,T,d$.
\end{proof}

% ============================================================
\subsection{Local Risk Geometry}
\label{app:toc-risk-geometry}

\begin{lemma}[Local risk--Fisher equivalence]
\label{lem:local-risk-fisher}
There exists a sufficiently small constant $\delta_{\mathrm{loc}}>0$ such
that, whenever
\[
    \|E\|_{G_M}\le\delta_{\mathrm{loc}},
\]
we have
\[
    c\|E\|_{G_M}^2
    \le
    \mathcal R_S(V_M^*+E)
    -
    \mathcal R_S(V_M^*)
    \le
    C\|E\|_{G_M}^2.
\]
Moreover,
\[
    cG_M
    \preceq
    \nabla^2\mathcal R_S(V_M^*+E)
    \preceq
    CG_M
\]
throughout the same neighborhood.
\end{lemma}

\begin{proof}
Write $H(V):=\nabla^2\mathcal R_S(V)$ and
\[
    \|E\|_{\Sigma_S}:=\|E\Sigma_S^{1/2}\|_F.
\]
The derivative of the finite-dimensional softmax covariance is uniformly
bounded, so for a constant $L_{\mathrm{sm}}$ depending only on the fixed
class dimension,
\[
    |u^\top(\widetilde C(a)-\widetilde C(b))v|
    \le L_{\mathrm{sm}}\|a-b\|_2\|u\|_2\|v\|_2.
\]
For arbitrary perturbations $D,E,F$, Gaussian moment bounds and H\"older's
inequality therefore give
\[
\begin{aligned}
    |\langle F,(H(V+D)-H(V))E\rangle_F|
    &\le L_{\mathrm{sm}}\,
       \mathbb E[\|Dz\|_2\|Ez\|_2\|Fz\|_2] \\
    &\le C\|D\|_{\Sigma_S}
          \|E\|_{\Sigma_S}
          \|F\|_{\Sigma_S}.
\end{aligned}
\]
The Fisher sandwich and duality between the $G_M$ and $G_M^{-1}$ norms
convert this to
\[
    \|(H(V+D)-H(V))E\|_{G_M^{-1}}
    \le C\|D\|_{G_M}\|E\|_{G_M}
\]
when $V=V_M^*$. In particular, for every $E$ in the stated ball and
$0\le t\le1$,
\[
    (1-Ct\|E\|_{G_M})\|F\|_{G_M}^2
    \le \langle F,H(V_M^*+tE)F\rangle_F
    \le (1+Ct\|E\|_{G_M})\|F\|_{G_M}^2.
\]
Choose $\delta_{\mathrm{loc}}$ so that $C\delta_{\mathrm{loc}}\le1/2$.
This proves the Hessian comparison throughout the ball. Since
$\nabla\mathcal R_S(V_M^*)=0$, Taylor's integral formula,
\[
    \mathcal R_S(V_M^*+E)-\mathcal R_S(V_M^*)
    =\int_0^1(1-t)\langle E,H(V_M^*+tE)E\rangle_F\,dt,
\]
then proves the risk comparison.
\end{proof}

% ============================================================
\subsection{Global Softmax Remainder}
\label{app:toc-global-remainder}

\begin{lemma}[Global softmax remainder]
\label{lem:local-softmax-remainder}
Let
\[
    F(V):=\nabla\mathcal R_S(V).
\]
For
\[
    E:=V-V_M^*,
\]
write
\[
    F(V_M^*+E)
    =
    G_ME+\mathcal N(E).
\]
For every $E,E'\in\mathbb R^{r\times M}$,
\[
    \|\mathcal N(E)\|_{G_M^{-1}}
    \le
    C\|E\|_{G_M}^2,
\]
and
\[
\begin{aligned}
    &
    \|\mathcal N(E)-\mathcal N(E')\|_{G_M^{-1}}
    \\
    &\qquad
    \le
    C
    \left(
        \|E\|_{G_M}
        +
        \|E'\|_{G_M}
    \right)
    \|E-E'\|_{G_M}.
\end{aligned}
\]
\end{lemma}

\begin{proof}
Let $H(V):=\nabla^2\mathcal R_S(V)$. The softmax-covariance Lipschitz
bound and Gaussian moment estimate established in the proof of
Lemma~\ref{lem:local-risk-fisher} imply, for arbitrary $D,E$,
\[
    \|(H(V_M^*+D)-G_M)E\|_{G_M^{-1}}
    \le C\|D\|_{G_M}\|E\|_{G_M}.
\]
The fundamental theorem of calculus gives
\[
    \mathcal N(E)
    =\int_0^1
       (H(V_M^*+tE)-G_M)E\,dt,
\]
so
\[
    \|\mathcal N(E)\|_{G_M^{-1}}
    \le C\|E\|_{G_M}^2.
\]
For $D:=E-E'$, another application along the segment from $E'$ to $E$
yields
\[
    \mathcal N(E)-\mathcal N(E')
    =\int_0^1
       (H(V_M^*+E'+tD)-G_M)D\,dt.
\]
Since the segment from $E'$ to $E$ lies in the parameter space,
\[
    \|E'+tD\|_{G_M}
    \le (1-t)\|E'\|_{G_M}+t\|E\|_{G_M}.
\]
The same Hessian estimate now gives the asserted Lipschitz bound.
\end{proof}

\subsection{Stein-Structured Softmax Remainder}
\label{app:toc-stein-remainder}

\begin{lemma}[Stein-structured softmax remainder]
\label{lem:stein-structured-remainder}
Let $\Delta B(E):=B(V_M^*+E)-B_M$, and define
\[
    \mathcal A(E):=\Delta B(E)B_M^{-1},
\]
and
\[
    \mathcal R(E):=
    \Delta B(E)-DB(V_M^*)[E]
    -\Delta B(E)B_M^{-1}DB(V_M^*)[E].
\]
Then, for every $E$,
\[
    \boxed{
    \mathcal N(E)
    =
    \mathcal A(E)G_M[E]
    +
    \mathcal R(E)V_M^*\Sigma_S.
    }
\]
Moreover, with constants independent of $M,N$,
\[
\begin{aligned}
    \|\mathcal A(E)\|_{\mathrm{op}}
    &\le C\|E\|_{G_M},\\
    \|\mathcal A(E)-\mathcal A(E')\|_{\mathrm{op}}
    &\le C\|E-E'\|_{G_M},\\
    \|\mathcal R(E)\|_{\mathrm{op}}
    &\le C\|E\|_{G_M}^2,\\
    \|\mathcal R(E)-\mathcal R(E')\|_{\mathrm{op}}
    &\le C(\|E\|_{G_M}+\|E'\|_{G_M})\|E-E'\|_{G_M}.
\end{aligned}
\]
\end{lemma}

\begin{proof}
The Stein identity and its derivative give
\[
\nabla\mathcal R_S(V_M^*+E)
=B(V_M^*+E)(V_M^*+E)\Sigma_S
-B_MV_M^*\Sigma_S,
\]
and
\[
G_M[E]=B_ME\Sigma_S+DB(V_M^*)[E]V_M^*\Sigma_S.
\]
Subtracting these expressions first yields
\[
\mathcal N(E)
=\Delta B(E)E\Sigma_S
+(\Delta B(E)-DB(V_M^*)[E])V_M^*\Sigma_S.
\]
The prediction-norm bound on $V_M^*$ implies that $V_M^*z$ is a
fixed-dimensional Gaussian vector with uniformly bounded second moment.
On an event of uniformly positive probability its norm is bounded, so the
softmax-curvature lower bound gives $B_M\succeq cI_r$. Thus $B_M^{-1}$
is uniformly bounded. Now
\[
E\Sigma_S
=B_M^{-1}G_M[E]
-B_M^{-1}DB(V_M^*)[E]V_M^*\Sigma_S.
\]
Substitution proves the claimed factorization.

For fixed class dimension, the first and second derivatives of the
softmax-covariance map are uniformly bounded. Gaussian moment bounds
therefore imply
\[
\begin{aligned}
\|\Delta B(E)\|_{\mathrm{op}}
&\le C\|E\|_{\Sigma_S},\\
\|\Delta B(E)-\Delta B(E')\|_{\mathrm{op}}
&\le C\|E-E'\|_{\Sigma_S},\\
\|\Delta B(E)-DB(V_M^*)[E]\|_{\mathrm{op}}
&\le C\|E\|_{\Sigma_S}^2,\\
\|[\Delta B(E)-DB(V_M^*)[E]]-[\Delta B(E')-DB(V_M^*)[E'] ]\|_{\mathrm{op}}
&\le C(\|E\|_{\Sigma_S}+\|E'\|_{\Sigma_S})\|E-E'\|_{\Sigma_S}.
\end{aligned}
\]
The same difference bound for the product term in $\mathcal R$ follows by
adding and subtracting one factor at a time.
The Fisher sandwich makes $\|E\|_{\Sigma_S}$ equivalent to
$\|E\|_{G_M}$, and $\|B_M^{-1}\|_{\mathrm{op}}\le C$. These facts
give all asserted coefficient estimates.
\end{proof}

\subsection{Uniform Ridge Source Filters}
\label{app:toc-ridge}

Let $B=S\Sigma^{1/2}$, $Q=A_*\Sigma^{1/2}$, and $W=BB^\top=\Sigma_S$.
Write $P_B=B^\top(BB^\top)^{-1}B$ and
\[
    V_M^{\mathrm{LS}}=QB^\top W^{-1},\qquad
    \Delta_M=\|Q(I-P_B)\|_F^2.
\]

\begin{lemma}[Uniform Gaussian-sketch ridge source filter]
\label{lem:uniform-ridge-source-filter}
Suppose Assumptions~\ref{ass:covariance} and~\ref{ass:source} hold and
$q=(b-1)/a\in(0,1)$. There are constants $D_0,u_0,c,C>0$, independent of
$M$, and a sketch event $\mathcal G_M^{\mathrm{ridge}}$ with
\[
    \mathbb P_S(\mathcal G_M^{\mathrm{ridge}})\ge 1-Ce^{-cM},
\]
such that, on this event, simultaneously for
$D_0M^{-a}\le u\le u_0$,
\[
    c u^q\le K_W(u)\le C u^q,
    \qquad
    K_W(u):=\operatorname{tr}\!\left[
       V_M^{\mathrm{LS}}uW(W+uI)^{-1}(V_M^{\mathrm{LS}})^\top
    \right].
\]
Moreover, write
\[
    F_W(u):=\|V_M^{\mathrm{LS}}\mathbf1_{(0,u]}(W)\|_W^2.
\]
The Tauberian bounds below give
\[
    F_W(u)\lesssim u^q\quad
    (D_0M^{-a}\le u\le u_0),
\]
and, for a fixed sufficiently large $L$,
\[
    F_W(u)\gtrsim u^q\quad
    (LD_0M^{-a}\le u\le u_0).
\]
Thus both bounds hold on the explicitly smaller interval
$[LD_0M^{-a},u_0]$. In particular, for fixed sufficiently small
$\epsilon>0$, if $H_T\ge\epsilon/u_0$ and
$M\ge (LD_0/\epsilon)^{1/a}H_T^{1/a}$, then
\[
    F_W\bigl(\epsilon/H_T\bigr)\gtrsim H_T^{-q}.
\]
\end{lemma}

\begin{proof}
The singular-value decomposition of $B$ gives the exact identities
\[
\begin{aligned}
    \Delta_M+K_W(u)
    &=\min_{V\in\mathbb R^{r\times M}}
       \{\|Q-VB\|_F^2+u\|V\|_F^2\}\\
    &=u\operatorname{tr}[Q(B^\top B+uI)^{-1}Q^\top].
\end{aligned}
\tag{G.2}\label{eq:ridge-variational}
\]
Indeed, the minimizer is $V_u=QB^\top(W+uI)^{-1}$; on each nonzero
singular direction of $B$, the minimized contribution is
$u|q_j|^2/(\sigma_j^2+u)$, while the null-space contribution is exactly
$\Delta_M$.

We first describe one event that works for all relevant scales. In the
eigenbasis of $\Sigma$, write $S_I$ for the sketch columns in a block
$I$. Fix a sufficiently small $\rho>0$. For every integer $1\le k\le\rho M$, the standard Gaussian singular-value
tails applied to the $M\times k$ matrix $S_{1:k}$ give
\[
\Pr\{cI_k\not\preceq S_{1:k}^\top S_{1:k}\preceq CI_k\}
\le 2e^{-cM}
\]
when $\rho$ is sufficiently small. A union bound over at most $M$ values
of $k$ costs only $Me^{-cM}$. For dyadic blocks
$I_\ell=(\ell,2\ell]\cap\{1,\ldots,d\}$, Gaussian upper singular-value
tails give
\[
\Pr\{\|S_{I_\ell}\|_{\mathrm{op}}^2>C(1+\ell/M)\}
\le 2e^{-c(M+\ell)}.
\]
The sum over dyadic $\ell\le M$ is $O(\log(M)e^{-cM})$ and the sum over
$\ell>M$ is $O(e^{-c'M})$, uniformly in the endpoint $d$. Thus all head
and block bounds hold on one event with failure probability $Ce^{-cM}$.
Splitting $\{j>k\}$ into dyadic blocks and using $\lambda_j\asymp j^{-a}$
then yields, simultaneously for $k\le\rho M$,
\[
    \|S_{>k}\operatorname{diag}(\lambda_{>k})S_{>k}^\top\|_{\mathrm{op}}
    \le Ck^{-a}.
\]
For blocks with $\ell\le M$, sum $C\ell^{-a}$; for $\ell>M$, sum
$C\ell^{1-a}/M=O(M^{-a})$. This is a finite sum because the ambient
dimension $d$ is finite.

Let $\mathcal H_M$ be this simultaneous head--tail event, intersected
with the trace event in Section~\ref{sec:assumptions}. Its failure
probability is at most $Ce^{-cM}$, uniformly in $d$. Set
$\mathcal G_M^{\mathrm{ridge}}=\mathcal H_M\cap\mathcal G_M$; a union
bound with the already-proved spectrum event preserves the same
probability order. In particular, the reconstruction below gives
directly $\Delta_M\le CM^{1-b}$, which is the only approximation-tail
estimate needed for the ridge lower bound.

For the upper bound, choose $k\asymp u^{-1/a}$ with constants so that
$k\le\rho M$. Let $A_H=A_*|_{1:k}$ and
$L_H=(S_{1:k}^\top S_{1:k})^{-1}S_{1:k}^\top$, and set $V_k=A_HL_H$.
The source condition and the head--tail event imply
\[
\|A_H\|_F^2\le Ck^{a-b+1},\qquad
\|L_H\|_{\mathrm{op}}\le C,
\]
where $L_H=(S_{1:k}^\top S_{1:k})^{-1}S_{1:k}^\top$. In the first $k$
input coordinates, $V_kB$ equals $Q$ exactly. On the remaining
coordinates, the teacher tail has squared Frobenius norm at most
$Ck^{1-b}$, and the reconstruction leakage satisfies
\[
\|A_HL_HS_{>k}\operatorname{diag}(\sqrt{\lambda_{>k}})\|_F^2
\le C\|A_H\|_F^2
 \|S_{>k}\operatorname{diag}(\lambda_{>k})S_{>k}^\top\|_{\mathrm{op}}
\le Ck^{1-b}.
\]
Therefore $\|Q-V_kB\|_F^2\le Ck^{1-b}$ and
$\|V_k\|_F^2\le Ck^{a-b+1}$. Taking $k=\lfloor\rho M\rfloor$ in the
same construction gives $\Delta_M\le CM^{1-b}$. Using \eqref{eq:ridge-variational}
and subtracting $\Delta_M\ge0$ gives
\[
    K_W(u)\le C(k^{1-b}+uk^{a-b+1})\le Cu^q.
\]

For the lower bound, choose a dyadic block $I_k=(k,2k]$ with
$k\asymp u^{-1/a}$, and put $Q_{I_k}=Q|_{I_k}$ and
$B_{I_k}=S_{I_k}\operatorname{diag}(\sqrt{\lambda_j})_{j\in I_k}$.
The source condition gives $\|Q_{I_k}\|_F^2\asymp k^{1-b}$ and the same
sketch event gives $B_{I_k}^\top B_{I_k}\preceq Ck^{-a}I$. We spell out
the restriction argument, since it is the point at which a source block
is converted into a lower bound for the full inverse. If $A=B^\top B$
and $q_c^\top$ is a row of $Q$, then
\[
 q_c^\top(A+uI)^{-1}q_c
 =\sup_x\{2q_c^\top x-x^\top(A+uI)x\}.
\]
Restricting this supremum to vectors supported on $I_k$ gives
\[
 q_c^\top(A+uI)^{-1}q_c
 \ge q_{c,I_k}^\top
     (A_{I_kI_k}+uI)^{-1}q_{c,I_k}
 =q_{c,I_k}^\top
     (B_{I_k}^\top B_{I_k}+uI)^{-1}q_{c,I_k}.
\]
Summing over the $r$ rows of $Q$ and using
$B_{I_k}^\top B_{I_k}+uI\preceq(Ck^{-a}+u)I$ yields
\[
\begin{aligned}
    \Delta_M+K_W(u)
    &\ge u\operatorname{tr}[Q_{I_k}
       (B_{I_k}^\top B_{I_k}+uI)^{-1}Q_{I_k}^\top]\\
    &\ge \frac{u}{Ck^{-a}+u}\|Q_{I_k}\|_F^2
     \ge ck^{1-b}\asymp u^q.
\end{aligned}
\]
Here a dyadic $k$ can be chosen with fixed comparison constants
$c_0u^{-1/a}\le k\le C_0u^{-1/a}$. After increasing $D_0$, the lower
endpoint $u\ge D_0M^{-a}$ ensures $2k\le\rho M$; the standing condition
$d\ge2M$ then ensures that the entire block $I_k$ is present.
Since $u\ge D_0M^{-a}$ and $aq=b-1$,
$\Delta_M\le CM^{1-b}\le CD_0^{-q}u^q$. Choose $D_0$ large enough
that $CD_0^{-q}\le c/2$, where $c$ is the dyadic-block lower constant.
Then $K_W(u)\ge(c/2)u^q$. Thus the approximation-tail subtraction is
uniform over the same continuum of $u$ on which the common head--tail
event was constructed.

For the Tauberian statements, write
\[
    K_W(u)=\int_{(0,\infty)}\frac{u\lambda}{\lambda+u}\,d\nu_W(\lambda),
    \qquad
    F_W(u)=\int_{(0,u]}\lambda\,d\nu_W(\lambda).
\]
If $D_0M^{-a}\le u\le u_0$, then for $\lambda\le u$ the ridge kernel
is at least $\lambda/2$, whence
$F_W(u)\le2K_W(u)\lesssim u^q$.
For the lower bound, take $v=u/L$ and assume
$LD_0M^{-a}\le u\le u_0$. On $\{\lambda>u\}$,
\[
\frac{v\lambda}{\lambda+v}
\le\frac{2}{L}\frac{u\lambda}{\lambda+u}.
\]
Consequently,
$K_W(v;\lambda>u)\le(2/L)K_W(u)\le C L^{q-1}v^q$.
Choose $L$ fixed large enough that this is at most
$(c/2)v^q$, where $c$ is the lower ridge constant. Then
$K_W(v;\lambda\le u)\ge(c/2)v^q$. Since
$v\lambda/(\lambda+v)\le\lambda$, this implies
$F_W(u)\ge(c/2)L^{-q}u^q$. The endpoint requirements used here are
exactly $v=u/L\ge D_0M^{-a}$ and $u\le u_0$; no upper estimate at
$Lu$ is needed. Substitution $u=\epsilon/H_T$ gives the stated
condition $M\ge(LD_0/\epsilon)^{1/a}H_T^{1/a}$ and
$H_T\ge\epsilon/u_0$.
\end{proof}

% ============================================================
\subsection{Gaussian-Sketch Spectrum}
\label{app:toc-spectrum}

\begin{lemma}[Spectrum of the Gaussian-sketch covariance]
\label{lem:sketch-spectrum}
Under Assumptions~\ref{ass:covariance} and $d\ge2M$, there is an event
$\mathcal G_M^{\mathrm{spec}}$ with probability at least
$1-Ce^{-cM}$ on which, simultaneously for $1\le j\le M$,
\[
    c j^{-a}\le\mu_j(\Sigma_S)\le Cj^{-a}.
\]
The constants are independent of $d$.
\end{lemma}

\begin{proof}
Work in the eigenbasis of $\Sigma$, so
$W=\sum_{j=1}^d\lambda_jS_{:,j}S_{:,j}^\top$. We use the standard
Gaussian singular-value tail bounds: if $G$ is an $m\times n$ standard
Gaussian matrix, then
$s_{\max}(G)\le\sqrt m+\sqrt n+t$ with failure at most
$2e^{-ct^2}$, and, for $m\ge n$,
$s_{\min}(G)\ge\sqrt m-\sqrt n-t$ with the same type of tail bound.
Apply these to $S_{1:k}$ for every $k\le\rho M$, with fixed sufficiently
small $\rho$ and $t=c_0\sqrt M$. A union bound over at most $M$ values of
$k$ gives
\[
    cI_k\preceq S_{1:k}^\top S_{1:k}\preceq CI_k
    \qquad(1\le k\le\rho M)
\]
except with probability $Ce^{-cM}$.

For dyadic blocks $I_\ell=(\ell,2\ell]\cap\{1,\ldots,d\}$, the upper
singular-value bound gives, simultaneously,
\[
    \|S_{:,I_\ell}\|_{\mathrm{op}}^2
    \le C(1+\ell/M).
\]
For $\ell\le M$ the failure probabilities sum to at most
$C\log(M)e^{-cM}$; for $\ell>M$ they are bounded by $Ce^{-c\ell}$,
whose dyadic sum is $Ce^{-c'M}$. Thus this event has a failure bound
uniform in the possibly growing ambient dimension $d$. For every
$k\le\rho M$, summing the dyadic blocks above $k$ yields
\[
\left\|\sum_{j>k}\lambda_jS_{:,j}S_{:,j}^\top\right\|_{\mathrm{op}}
\le C\sum_{\ell\gtrsim k}\ell^{-a}(1+\ell/M)
\le Ck^{-a},
\]
where $a>1$ is used for the blocks beyond $M$. The same Gaussian
singular-value lower-tail bound applied to the $M\times2M$ block
$S_{1:2M}$ gives
$\lambda_{\min}(S_{1:2M}S_{1:2M}^\top)\ge c$ except with
probability $Ce^{-cM}$. Intersect these events and the trace event from
Section~\ref{sec:assumptions}; call the intersection
$\mathcal G_M^{\mathrm{spec}}$.

For $2\le j\le\rho M$, take $k=j-1$. The rank-$k$ head plus the tail
bound and the min--max principle give
$\mu_j(W)\le C(j-1)^{-a}\le C'j^{-a}$. For larger $j$, monotonicity
and the upper bound at $\lfloor\rho M\rfloor$ give
$\mu_j(W)\le CM^{-a}\le C'j^{-a}$; $j=1$ follows from
$\mu_1(W)\le\operatorname{tr}(W)\le C$.

For $j\le\rho M$, Loewner monotonicity gives
\[
W\succeq\lambda_jS_{1:j}S_{1:j}^\top,
\]
whose $j$th largest eigenvalue is at least $cj^{-a}$ by the head-block
singular-value bound. Hence $\mu_j(W)\ge cj^{-a}$. Finally,
\[
W\succeq\lambda_{2M}S_{1:2M}S_{1:2M}^\top\succeq cM^{-a}I_M,
\]
so every $j>\rho M$ has $\mu_j(W)\ge cM^{-a}\ge c'j^{-a}$.
This proves the two-sided spectrum claim.
\end{proof}

\subsection{Gaussian-Sketch Effective Dimension}
\label{app:toc-effective-dimension}

\begin{lemma}[General-schedule Gaussian-sketch effective dimension]
\label{lem:sketch-effective-dimension}
On $\mathcal G_M^{\mathrm{spec}}\cap\mathcal G_M$, for a schedule satisfying
the stable learning-rate condition in Section~\ref{subsec:optimization} and
$H_T\to\infty$, let
$f_T(\lambda)=1-\prod_{t=1}^T(1-\eta_t\lambda)$ and
$k_T=\min\{M,H_T^{1/a}\}$. Then
\[
\boxed{
    \operatorname{tr}[f_T(G_M)^2]\asymp r k_T.
}
\]
The constants are uniform in $M,T,d$ and the admissible schedule (and may
depend on the fixed stability and structural parameters).
\end{lemma}

\begin{proof}
The Fisher sandwich and min--max principle compare each ordered
 eigenvalue of $G_M$ with the corresponding eigenvalue of
$I_r\otimes W$ up to fixed factors. For $0\le\eta_t\lambda\le1/2$,
Lemma~\ref{lem:general-step-filter} gives
\[
 c\min\{1,H_T\lambda\}
 \le 1-\prod_{t=1}^T(1-\eta_t\lambda)
 \le C\min\{1,H_T\lambda\}.
\]
Thus the trace is comparable to
\[
 r\sum_{j=1}^M\min\{1,(H_T j^{-a})^2\},
\]
by Lemma~\ref{lem:sketch-spectrum}. If $x=H_T^{1/a}\le M$, the
first $\lfloor x\rfloor$ terms contribute at most $x$, while the tail
is bounded by $CH_T^2x^{1-2a}=Cx$; the first $c x$ terms each give a
positive constant, yielding a matching lower bound. If $x>M$, all $M$
terms are bounded above by $M$ and, using $\mu_M\gtrsim M^{-a}$, each
term is bounded below by a positive constant. Since $H_T\to\infty$,
integer-rounding and the bounded initial range are absorbed into constants.
This proves the claim.
\end{proof}

% ============================================================

\section{Synthetic-Experiment Details}
\label{app:additional-experiments}

All experiments use three classes, $a=1.5$, $b=1.8$, signal scale $0.8$,
and constant learning rate $\eta_t=0.5$, giving $H_T=0.5T$.
Population fitting uses $32{,}768$ Gaussian
feature draws and evaluation uses an independent test set of $65{,}536$
draws.  To hold the truncated teacher fixed when $M$ varies, the time,
sketch, and collapse sweeps use respectively $d=2048$, $1024$, and $720$;
the rate-tuned empirical sweep uses $d=8M$.  The population-time sweep uses
$T\in\{8,16,32,64,128,256,512\}$ and selects $M$ to keep
$M/H_T^{1/a}\approx6.35$. The converged sketch sweep uses
$M\in\{8,12,18,27,40,60,90,128\}$.  The empirical sweep uses
$N\in\{512,1024,2048,4096,8192,16384\}$, with the joint tuning described
in the main text.  Every population or sketch point averages five independent
sketches; every empirical point additionally averages five independently
drawn training samples per sketch.

For the common-scale experiment,
$M\in\{12,18,27,40,60,90\}$,
$T\in\{16,32,64,128,256,512\}$, and
$N\in\{1024,2048,4096,8192,16384\}$.  We retain combinations satisfying
$M/H_T^{1/a}\ge1.25$, leaving $130$ distinct design points. For each
fixed $(M,T)$ and sketch replicate, all sample sizes share the same sketch
and test set, and the training samples are nested across $N$.  Pointwise
error bars average five sketches and five label-sample replicates.  The
sufficient large-sample ratio in Theorem~\ref{thm:metric-scaling} is not
uniformly small on this computational grid, so this experiment assesses
finite-scale exponent stability rather than numerically certifying the
asymptotic high-probability event.

The literal unit-prefactor scale
$M^{1-b}+H_T^{-q}+k_T/N$ gives slopes $0.763$, $0.722$, and $0.363$.
This under-collapse is expected because the theorem specifies the three
summands only up to constants.  With the optimization weight normalized to
one, leave-one-$N$-out fitting gives approximation weights in
$[1.827,1.846]$ and sample-fluctuation weights in $[4.933,5.156]$ across the five
folds. Their stability, together with held-out slopes $0.958$, $0.912$,
and $0.459$, indicates that the improved collapse is not driven by a single
sample-size slice.

\end{document}